%% file: main.tex
\documentclass[11pt]{article}

\input{./preamble}

\usepackage{algorithm}
\usepackage{algorithmic}

\usepackage{./docstyle}
\usepackage{enumitem}
\usetikzlibrary{calc}
\usetikzlibrary{bayesnet}

\usepackage{wrapfig} 
\usepackage{color}
\usepackage{mathtools}
\definecolor{darkgreen}{RGB}{0,179,0}

\newlength\myindent
\newlength\myindentt
\newcommand\bindent{%
    \begingroup
    \setlength{\itemindent}{\myindent}
    \addtolength{\algorithmicindent}{\myindent}
}
\newcommand\eindent{\endgroup}

\newcommand\bbindent{%
    \begingroup
    \setlength{\itemindent}{\myindentt}
    \addtolength{\algorithmicindent}{\myindentt}
}
\newcommand\eeindent{\endgroup}

\newcommand{\myref}[2]{\hyperref[#2]{#1 \ref*{#2}}}

\begin{document}

\title{Why Calibration Error is Wrong Given Model Uncertainty: Using Posterior Predictive Checks with Deep Learning}

\input{author}

\date{\today}
\maketitle

\input{model_uncertainty_main}

\end{document}

%% file: preamble.tex
\usepackage{multirow}
\usepackage{amsmath,amssymb}
\usepackage{relsize}
\usepackage{graphicx}
\usepackage{listings}
\usepackage{xcolor}
\usepackage{booktabs}
\usepackage{fancyref}
\usepackage[parfill]{parskip}
\usepackage{todonotes}
\usepackage{import}
\usepackage{dsfont}
\usepackage{lineno}
\usepackage{float}
\usepackage{caption}
\usepackage[skip=10pt]{subcaption}
\usepackage{bm}
\usepackage{bbm}
\usepackage{commath}
\usepackage[section]{placeins}
\usepackage{fancyhdr}
\usepackage{natbib}
\usepackage{datetime}
\usepackage[hidelinks]{hyperref}

\numberwithin{equation}{section}

\hypersetup{pdfinfo={
	CreationDate={D:20200407195600},
	ModDate={D:20200407195600}
}}

%% file: author.tex
\author{%
  Achintya Gopal \\
  Bloomberg Quant Research\\
  \texttt{achintyagopal@gmail.com} \\
}

%% file: model_uncertainty_main.tex
\begin{abstract}
Within the last few years, there has been a move towards using statistical models in conjunction with neural networks with the end goal of being able to better answer the question, ``what do our models know?''.  From this trend, classical metrics such as Prediction Interval Coverage Probability (PICP) and new metrics such as calibration error have entered the general repertoire of model evaluation in order to gain better insight into how the uncertainty of our model compares to reality. One important component of uncertainty modeling is model uncertainty (epistemic uncertainty), a measurement of what the model does and does not know. However, current evaluation techniques tends to conflate model uncertainty with aleatoric uncertainty (irreducible error), leading to incorrect conclusions. 
In this paper, using \textit{posterior predictive checks}, we show how calibration error and its variants are almost always incorrect to use given model uncertainty, and further show how this mistake can lead to trust in bad models and mistrust in good models. Though posterior predictive checks has often been used for in-sample evaluation of Bayesian models, we show it still has an important place in the modern deep learning world.
\end{abstract}

\section{Introduction}

Classical machine learning models and many deep learning models are evaluated using aggregate statistics such as RMSE and accuracy. Once a model has been found to perform well on one or more of these metrics, the model might be deployed into the real world. However, metrics such as these have been found to not accurately describe the limitations of what the models know. Especially as machine learning moves towards being more adopted in fields such as medicine \citep{zhou20021medimaging,chen2018voxresnet,wachinger2018deepnat,havaei2017brain} where inaccurate predictions can lead to devastating results and self driving cars \citep{2016arXiv160407316B,huang2020autonomous,badue2021self} where models might need to make decisions in the face of new environments, a measurement of uncertainty becomes crucial. 

In response to this, statistical approaches have been merged with deep learning in methods such as variational inference \citep{PracticalVI2011,pmlr-v37-blundell15,MCDropout,NIPS2015_bc731692,rezende2014stochastic}, expectation propagation \citep{ProbBackProp}, and stochastic gradient MCMC \citep{Welling2011,zhang2020csgmcmc} so that the model gives not only its best estimate in the form of a mean or median, but also gives its uncertainty around the prediction, in the form of both aleatoric uncertainty (the uncertainty arising from hidden variables or measurement errors) and epistemic (model) uncertainty (the uncertainty arising from fitting a model using finite data). 

With this increase in the usage of statistical models, Prediction Interval Coverage Probability (PICP), calibration error \citep{kuleshov2018calibration} and proper scoring rules such as log-likelihood and Brier Score \citep{BrierScore} have also been incorporated into the set of commonly used evaluation metrics.

However, though PICP and calibration error are correct for evaluating the aleatoric uncertainty of univariate distributions, they have been incorrectly used to evaluate model uncertainty \citep{papadopoulos2000,kuleshov2018calibration,Ovadia2019}.
In this paper, we show why it is incorrect to use PICP and calibration error when evaluating model uncertainty and explain scenarios in which it is correct to use; more specifically, we show that, with model uncertainty, CDF (cumulative distribution function) values are not expected to be uniformly distributed. Further, we show how flaws in bad models have been mischaracterized and how good models have been rejected due to its non-zero calibration error being incorrectly attributed to bad performance.

\section{Background}

\subsection{Uncertainties}

Most taxonomies classify uncertainty into three sources: approximation, aleatoric, and epistemic uncertainty \citep{UncertaintyTaxonomy}. Approximation uncertainty quantifies the error from fitting a simple model to complex data. Aleatoric uncertainty (or irreducible error) quantifies the uncertainty of the conditional distribution of the target variable given features. This uncertainty arises from hidden variables or measurement errors and cannot be reduced through collecting more data under the same experimental conditions. Epistemic uncertainty (or model uncertainty) quantifies the uncertainty arising from fitting a model using finite data, i.e. it is inversely proportional to the density of the training examples and can be reduced by collecting data in the low density regions.

These different sources of uncertainty have different techniques for handling them. Using high capacity models such as neural networks removes a large part of the approximation uncertainty. By fitting a full distribution on the target conditional on features $p(y|x, \theta)$, we can model the aleatoric uncertainty from observations. Inaccurate estimates of aleatoric uncertainty can be explained by underfitting (insufficient complexity in the conditional distributions) or overfitting (models with sufficient capacity can memorize the data, leading to the distributions collapsing to deltas). Epistemic uncertainty is the distribution over the parameters $p(\theta)$ such as a Bayesian posterior or ensembles and gives access to what the model knows and does not know. \textit{In this paper, we use $p(\theta)$ to denote the posterior.}

\subsection{Regression}

The goal of regression is to model the conditional probability distribution $p(y|x, \theta)$.  Frequently, this task is reduced to obtaining point estimates of the mean or median by minimizing the mean squared error or mean absolute error respectively. In order to access uncertainties, we must model the full conditional distribution; in addition to this, we can still retrieve point estimates of the mean, median, or any other distributional statistic. 

Building on the statistical viewpoint of modeling, in this paper, whereas most papers view classification models through the lens of discriminative models, we analyze both regression models and classification models from the lens of generative modeling, where both are viewed as conditional distributions. In this way, we care not only about the predicted class of a classification model or the mean/median of a regression model but the complete predicted distribution, from which discriminative models are arguably a special case, e.g. mode of the predicted distribution from a classification model. 

One could argue that most of discriminative modeling is implicitly a generative model, i.e. cross entropy loss for classification is simply the log-likelihood of a multinomial, mean squared error is simply learning the location of a Gaussian, mean absolute error is simply learning the location of a Laplace, etc.

To evaluate the calibration of $p(y|x, \theta)$, the CDF $F(y|x, \theta)$ is used. In many cases such as a Normal and Laplace, this can be trivial to compute.

Though distributional regression often aims to learn $p(y|x, \theta)$, to incorporate model uncertanty, $p(\theta)$ is learned as well. In this case, whereas the CDF without model uncertanty is $F(y | x, \theta)$, the CDF given model uncertainty is:
\begin{equation}\label{eqn:model_unc_cdf}
 F(y | x) = \int d\theta\ p(\theta)\ F(y | x, \theta) 
 \end{equation}

\subsection{Evaluating Uncertainty}

The metrics defined in this section are commonly used in evaluation of uncertainty and were derived using the definition of perfect calibration \citep{GuoCalibration,kuleshov2018calibration}:
\begin{equation}\label{eqn:perf_calib}
 \mathbb{P}(F(Y|X) < p) = p, \quad \forall p \in [0, 1] 
\end{equation}
Though \myref{Equation}{eqn:perf_calib} is correct for a single $y$ and is correct for a set $\{y_i\}_{i=1}^{n}$ where the $y_i$ are conditionally independent given $x_i$, we show in \myref{Section}{sec:eval_w_ppc} that a set $\{y_i\}_{i=1}^{n}$ is not necessarily conditionally independent when model uncertainty is introduced, making calibration error incorrect to use.

\subsubsection{PICP}

A classical metric used to evaluate calibration is Prediction Interval Coverage Probability (PICP): the fraction of true observations falling inside the estimated prediction interval. Given a confidence interval $[\frac{\alpha}{2}, 1-\frac{\alpha}{2}]$, the goal for PICP is for (1 - $\alpha$)\% of the observed CDFs $F(y_i|x_i)$ to be found in this range \citep{papadopoulos2000,SQR}.

\subsubsection{Calibration Error}\label{sec:calib_err_intro}

\citet{GuoCalibration} showed that modern neural network-based classifiers are poorly calibrated. In order to diagnose this issue, Expected Calibration Error (ECE) was introduced:
\begin{equation}
\begin{aligned}
 \text{acc}(B_m) \coloneqq \frac{1}{\abs{B_m}} \sum_{i \in B_m} \mathbb{I}_{y_i = \hat{y}_i} &\qquad
 \text{conf}(B_m) \coloneqq \frac{1}{\abs{B_m}} \sum_{i \in B_m} \hat{p}_i \\
 \text{ECE} \coloneqq \sum_{m=1}^{M} \frac{\abs{B_m}}{N} &\left[ \text{acc}(B_m) - \text{conf}(B_m) \right]
\end{aligned}\label{eqn:ece}
\end{equation}
where $B_{m}$ is the set of indices of samples whose prediction confidence $\hat{p}_i$ (the estimated probability for label $y_i$) falls into the interval $(\frac{m-1}{M}, \frac{m}{M} ]$, $M$ is the number of intervals that are evaluated, $N$ is the number of data points we are evaluating, $\hat{y}_i$ is the predicted label for the $i$-th sample and ${y}_i$ is the true label for the $i$-th sample. 

\citet{kuleshov2018calibration} extended the analysis in \citet{GuoCalibration} to neural network-based regressors, introducing calibration error as a metric to quantitatively measure how well the quantiles are aligned:
\begin{equation}
\begin{aligned}
 \hat{p}_j &\coloneqq {\abs{\{y_n  | F(y_n|x_n) < p_j, n=1,\dots, N \}}}\ /\ {N} \\ 
 \text{cal}(y_1,\dots,y_N) &\coloneqq \sum_{j=1}^{M} (p_j - \hat{p}_j)^2
\end{aligned}\label{eqn:calib_err}
\end{equation}
where $F(\cdot|x_n)$ is the predicted CDF function, $N$ is the number of data points we are evaluating, $M$ is the number of quantiles that are evaluated, and $\{p_j\}_{j=1}^{M}$ is the set of quantiles are evaluated.

\subsection{Posterior Predictive Checks (PPC)}\label{sec:ppc}

Posterior predictive checks can be viewed as a generalization of hypothesis testing to the Bayesian framework. This technique for checking the fit of a Bayesian model boils down to drawing simulated values from the posterior predictive distribution and compare these samples to the observed data. Similar to hypothesis testing, any systematic differences between the simulations and the observed data indicate potential failings of the model. More details on this can be found in \citet{gelman2013bayesian}.

In the language of hypothesis testing, the null hypothesis being tested is: we have a distribution of functions $p(\theta)$ (collection of models) where the data we observe is from \textit{one} of these models ($p(Y | X, \theta)$), but we do not know which model (i.e. the true model is latent). In the notation above, $\theta$ can be viewed as the parameters of our model (e.g. a neural network); $Y$ is the target we are modeling and can be as simple as classes for image classification or as complicated as the images themselves for image generative modeling; $X$ is data we are conditioning on such as the images for image classification or the image classes when doing conditional image generation. $p(\theta)$ is the posterior in the Bayesian framework; $p(\theta)$ can also be a finite mixture of delta distributions such as in the case of ensembling. 

Following the steps of statistical hypothesis testing, a test statistic $T$ is defined over $X$, $Y$ and the model; the distribution of $T$ is then derived from $p(\theta)$ and $p(Y | X, \theta)$. In the case of posterior predictive checks, due to the complexity of $p(\theta)$, the distribution of $T$ is built up empirically (\myref{Algorithm}{alg:ppc}). In the language of posterior predictive checks, an empirical distribution is built up from $y^{rep}$, replicated data that could have been observed according to the model (samples from the posterior predictive distribution).

\myref{Algorithm}{alg:ppc} shows the general algorithm for PPC. For ensembles, the sampling from $p(\theta)$ is equivalent to randomly picking a model from the ensemble; for MC Dropout, the sampling is equivalent to sampling a single mask to apply to all rows. $T$ can then be any statistic such as log-likelihood, ECE, accuracy, Brier score, etc. In essence, any metric that could be used for model evaluation. Importantly, though the test statistic is computed from a sampled $\theta$, $T$ is not a function of $\theta$ as it is latent. More concrete examples are given in \myref{Appendix}{sec:ppc_example}.

Given the distribution of $T$, in hypothesis testing a significance level is picked, or in other words, the values of $T$ are partitioned into two regions: values where the hypothesis will be accepted and values where the hypothesis will be rejected. 
Though not discussed in detail in this paper, the design of the test statistic is very important in order to find regions where the model is failing which goes hand-in-hand with building trust in the model. 

A difference between the usage of posterior predictive checks in \citet{gelman2013bayesian} and this paper is that our evaluation is out of sample. The hypothesis we are testing should hold no matter if the data is out of sample or even out of distribution (covariate shift).

\begin{algorithm}[t]
   \caption{Creating a sample of a test statistic $T$ given model uncertainty $p(\theta)$. The test statistic $T$ is a function of the labels $y$, features $x$, and the model ($p(\theta)$ and $p(y | x, \theta))$. Importantly, $T$ is not a function of the $\theta$ sampled in the first step of the algorithm.}\label{alg:ppc}
\begin{algorithmic}
   \STATE {\bfseries Input:} data $\{x_i\}_{i=1}^{N}$, model distribution $p(\theta)$
   \STATE $\theta \sim p(\theta)$
   \STATE $y_{i} \sim p(y | x_i, \theta)$
   \STATE {\bfseries Output:} T($\{y_{i}\}_{i=1}^{N}$, $\{x_i\}_{i=1}^{N}$, $p(\theta)$, $p(y | x, \theta)$)
\end{algorithmic}
\end{algorithm}

\section{Evaluating Model Uncertainty Using PPCs}\label{sec:eval_w_ppc}

The main assumption used for PICP, ECE, and calibration error is that the model CDFs should be uniformly distributed. 
Though there has been additional work showing issues with ECE and calibration error, the solutions to these problems boils down to improvements to uniformity testing (such as using the Kolmogorov-Smirnov test \citep{gupta2021calibration} or changes to the binning scheme \citep{ding2020revisiting,nixon2019measuring}).

The intuition behind testing for uniformity in the regression setting is that a well calibrated model means that observations should fall in a 90\% confidence interval 90\% of the time \citep{kuleshov2018calibration}; in classification, the intuition is that given a confidence of 0.9, the true observations should be correctly classified 90\% of the time \citep{GuoCalibration}. Though this intuition holds when working with models without model uncertainty, we show in this section how this intuition can be incorrect when used in settings with model uncertainty (\myref{Section}{sec:eval_cdfs}).
Further, in this section, we introduce new metrics to aid in correct evaluation of model uncertainty (\myref{Section}{sec:metrics}) and discuss how calibration error should be foregone for Bayesian models and ensembles (\myref{Section}{sec:ensembles}).

\subsection{Evaluating CDFs}\label{sec:eval_cdfs}

To start, we choose a very simple test statistic: the CDF of a single observation. Using \myref{Algorithm}{alg:ppc}, given a single data point $x$, the generative process is:
\begin{enumerate}
    \item Sample $\theta$ from $p(\theta)$
    \item Sample $y^{ref}$ from $p(y | x, \theta)$
\end{enumerate}

From this, we can compute the CDF $F(y^{ref} | x)$ using \myref{Equation}{eqn:model_unc_cdf}.
Importantly, we integrated out $\theta$ as it is latent.

Given that the above process is sampling from a mixture (indexed by $\theta$) and $F(y^{ref} | x)$ is the CDF of the mixture, the test statistic is uniformly distributed (more detailed derivation in \myref{Appendix}{sec:uniformity_single}.
From here, we can evaluate $F(y | x)$ (the CDF of the observed) to see how far away from a uniform distribution the observed is.
An important distinction between this test statistic and the ones visualized in \citet{kuleshov2018calibration} is that this is the CDF of \textit{one} data point, not a full data set.


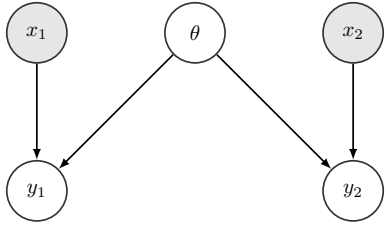
\begin{figure}
\begin{center}
\scalebox{0.8}{
\begin{tikzpicture}[
  every neuron/.style={
    circle,
    minimum size=0.3cm,
    thick
  },
  every data/.style={
    rectangle,
    minimum size=0.4cm,
    thick
  },
]
\tikzstyle{main}=[circle, minimum size = 10mm, thick, draw =black!80, node distance = 16mm]
\tikzstyle{connect}=[-latex, thick]
\tikzstyle{box}=[rectangle, draw=black!100]

  \node[main, fill = black!10] (x1) {$x_1$ };

  \node[main, fill = white!100,right=of x1] (m) {$\theta$ };

  \node[main, fill = black!10,right=of m] (x2) {$x_2$ };

  \node[main, fill = white!100, below=of x1] (y1) {$y_1$ };
  \node[main, fill = white!100, below=of x2] (y2) {$y_2$ };

  \path (x1) edge [connect] (y1)
        (x2) edge [connect] (y2)
        (m) edge [connect] (y1)
        (m) edge [connect] (y2);

\end{tikzpicture}
}
\end{center}
\caption{In this graphical model, we can see that since $\theta$ is latent, $y_1$ and $y_2$ are not conditionally independent given $x_1$ and $x_2$. However, if we were not modeling model uncertainty, $\theta$ would be observed and then $y_1$ and $y_2$ would conditionally independent given $x_1$ and $x_2$.}\label{fig:model_unc_cdf}
\end{figure}

Though the CDF of a single observation is uniformly distributed, we will see that the CDFs (\myref{Equation}{eqn:model_unc_cdf}) of a set of observations might not be. The reason is that when we compute the CDF of $n$ observations, though the CDF of each is marginally uniform, there might be correlation between the CDFs causing a single observation of $n$ CDFs to not be uniformly distributed (\myref{Figure}{fig:model_unc_cdf}). 

More specifically, we are saying that calibration error and PICP assume that a set of CDFs $\{F_i\}_{i=1}^{n}$ should be equal in distribution to a uniform distribution,
or, in other words, $\alpha$\% of the observed CDFs is expected to be found in the range $[0, \alpha]$ as we get more observations of $F_i$. Importantly, though $F_i$ is computed by integrating out $\theta$, the $y_i$ which we observe (according to our hypothesis) is from a single $\theta$.

We can see, using a simple example, that the assumption 
might not hold true in the presence of a latent variable:
\begin{equation}\label{fig:sim_gen}
\begin{aligned}
   \theta &\sim \mathcal{N}(0, 1)
\\ y_i &\sim \mathcal{N}(\theta, 1)
\end{aligned}
\end{equation}
where 
$\theta$ can be interpreted as the parameters of the model. In \myref{Equation}{fig:sim_gen}, $y_i$ is marginally distributed as $\mathcal{N}(0, 2)$. However, \textit{$\theta$ induces correlation} which we show causes the assumption to be violated. Assuming the true data is generated from $\theta = 0.5$ (i.e. the true model has high likelihood in $p(\theta)$), if we use $\alpha=0.5$:
\begin{align}
    \mathbb{P} (F(y_i  | \mu=0.0, \sigma^2=2.0) < 0.5) &= \mathbb{P}(y_i < 0)\label{eqn:label}
    \\ &= F(0 | \mu=0.5, \sigma^2=1.0)
\\        &\neq 0.5 = P(\text{Uniform}(0,1) < 0.5)
\end{align}
The reason we do not use the CDF of $\mathcal{N}(0.5, 1.0)$ in \myref{Equation}{eqn:label} is because we need to use the CDF that we would have access to during evaluation, i.e. the CDF after integrating out $\theta$ (\myref{Equation}{eqn:model_unc_cdf}). Generally, $F(y_i | \mu=0.0, \sigma^2=2.0)$ is never uniformly distributed for any setting of $\theta$. Since $F(y_i |\mu = 0.0, \sigma = 2.0)$ is never uniform, calibration error would be non-zero. Thus, in this simple case, though our model gives high density to the true model (understands its model uncertainty), even with infinite data, \textit{calibration error would lead to mistrust of the correct model.}

Further, say our model is \myref{Equation}{fig:sim_gen} but the true data generating process is $y_i \sim \mathcal{N}(0,2)$. Given infinite observations, the calibration error would be zero; however our model never expects to see data sampled from $\mathcal{N}(0,2)$ since $\mathcal{N}(0,2)$ is not in the hypothesis space. Thus, even though we would have zero calibration error, our model is incorrect and does not actually know what it does not know. In other words, even with infinite data, \textit{calibration error would lead to trust in this incorrect model.}

More details on the connection between this example and a Bayesian model for fitting a normal distribution with known variance can be found in \myref{Appendix}{sec:bayesian_example}.

\subsection{Metrics}\label{sec:metrics}

For PPC, given a test statistic $T$, we can compute the observed value $\hat{T}$ as well as generate samples from the posterior predictive distribution $\mathcal{T}$ of $T$ using \myref{Algorithm}{alg:ppc}. To evaluate $\hat{T}$, we introduce two simple metrics: 
\begin{itemize}
\item \textit{p-value} In this paper, we refer to p-value as the percentage of the distribution $\mathcal{T}$ less than $\hat{T}$. Sometimes, say for normal distributions, p-value is defined to be symmetric; further, for metrics like accuracy, we might define the p-value as the percentage of the posterior greater than the observed. To be less opinionated about the metric, we purely give it as the CDF of the observed value given the posterior predictive distribution. 

Similar to how in hypothesis testing the values of $T$ are partitioned into two regions, the choice of valid and invalid p-values is a decision to be made by the modeler. In this paper, we say a PPC passes if the p-value does not equal zero or one, or in other words, the observed value $\hat{T}$ is not out of distribution.

\item \textit{Sharpness} Similar to \citet{kuleshov2018calibration}, we define sharpness as the width of the posterior predictive distribution $\mathcal{T}$. In this paper, we use the difference between 5th and 95th quantile. The goal is to have thin (confident) posterior predictive distributions while passing posterior predictive checks.

\end{itemize}

Due to the simplicity of these metrics, there are many scenarios in which they will not be the easiest to interpret and will be limited, especially in the case where the posterior predictive distribution is multimodal. Further, the quality of the metrics will be limited by how representative the samples from $\mathcal{T}$ are; though, this can be fixed by increasing the number of samples.
Example pseudocode to compute the above metrics is given in \myref{Appendix}{sec:ppc_example}.

\subsection{Ensembles}\label{sec:ensembles}

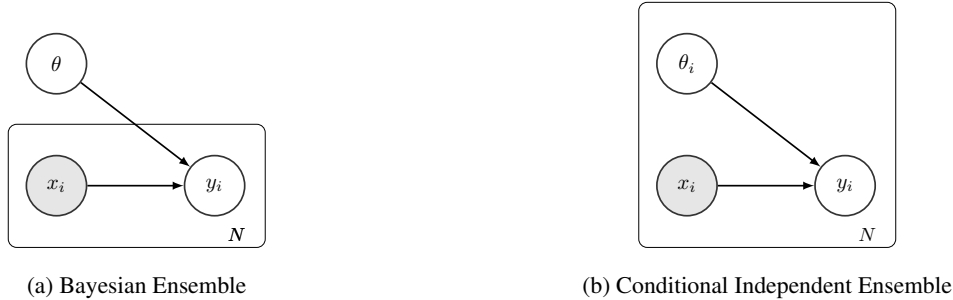
\begin{figure}
\begin{center}
\begin{subfigure}{0.45\linewidth}
\centering
\scalebox{0.8}{
\begin{tikzpicture}[
  every neuron/.style={
    circle,
    minimum size=0.3cm,
    thick
  },
  every data/.style={
    rectangle,
    minimum size=0.4cm,
    thick
  },
]
\tikzstyle{main}=[circle, minimum size = 10mm, thick, draw =black!80, node distance = 16mm]
\tikzstyle{connect}=[-latex, thick]
\tikzstyle{box}=[rectangle, draw=black!100]

  \node[main, fill = black!10] (x1) {$x_i$ };

  \node[main, fill = white!100,above=of x1,yshift=-0.6cm] (m) {$\theta$ };

  \node[main, fill = white!100, right=of x1] (y1) {$y_i$ };

  \path (x1) edge [connect] (y1)
        (m) edge [connect] (y1);
  \plate [inner sep=.3cm,xshift=.02cm,yshift=.2cm] {plate1} {(x1)(y1)} {$N$};
  \plate [inner sep=0.3cm,xshift=.02cm,yshift=.2cm,opacity=0] {plate1} {(x1)(y1)(m)} {$N$};

\end{tikzpicture}
}
\caption{Bayesian Ensemble}\label{fig:bayes_ensem}
\end{subfigure}\hfill
\begin{subfigure}{0.45\linewidth}
\centering
\scalebox{0.8}{
\begin{tikzpicture}[
  every neuron/.style={
    circle,
    minimum size=0.3cm,
    thick
  },
  every data/.style={
    rectangle,
    minimum size=0.4cm,
    thick
  },
]
\tikzstyle{main}=[circle, minimum size = 10mm, thick, draw =black!80, node distance = 16mm]
\tikzstyle{connect}=[-latex, thick]
\tikzstyle{box}=[rectangle, draw=black!100]

  \node[main, fill = black!10] (x1) {$x_i$ };

  \node[main, fill = white!100,above=of x1,yshift=-0.6cm] (m) {$\theta_i$ };

  \node[main, fill = white!100, right=of x1] (y1) {$y_i$ };

  \path (x1) edge [connect] (y1)
        (m) edge [connect] (y1);
  \plate [inner sep=0.3cm,xshift=.02cm,yshift=.2cm] {plate1} {(x1)(y1)(m)} {$N$};

\end{tikzpicture}
}
\caption{Conditional Independent Ensemble}\label{fig:indep_ensem}
\end{subfigure}\hfill
\end{center}
\caption{Comparison of Conditionally Independent Ensemble and Bayesian Ensemble using Graphical Models.}\label{fig:comp_ensem}
\end{figure}

One observation that can be made from the discussion in \myref{Section}{sec:eval_cdfs} is that expecting uniform model CDFs is valid if our model uncertainty is conditionally independent given $x$ (\myref{Figure}{fig:indep_ensem}) or if the distribution over our models is a delta distribution (point estimate). In the case of Bayesian models and Bayesian posteriors, by definition, this is not true and so for the evaluation of Bayesian models, testing uniformity of CDFs must be foregone. However, \textit{is the model uncertainty conditionally independent in classical ensembles?}

Given the training procedure of classical ensembles, there is nothing stopping us from treating the model uncertainty as conditionally independent. However, one could argue that training ensembles is equivalent to variational inference with a mixture of delta distributions as the posterior; the question becomes: \textit{which assumption works better: Bayesian Ensembles (\myref{Figure}{fig:bayes_ensem}) or Conditionally Independent Ensembles (\myref{Figure}{fig:indep_ensem})?}

The validity of a model uncertainty technique can be interpreted in two ways: one in which posterior predictive checks pass and the other being when the assumption aligns with intuition (the first being more scientific). We show results of the first in \myref{Section}{sec:exps} where we can see that the model is better calibrated to understanding its accuracy and calibration assuming \myref{Figure}{fig:bayes_ensem}. In terms of the second argument, intuition also suggests a Bayesian interpretation as we could imagine two images with very little difference (perturb one pixel) and would expect the predictions to be consistent with each other; however, if we take a conditionally independent assumption, our model could give extremely different predictions even given a tiny perturbation. We further discuss the counter-intuitive nature of conditional independent ensembles in \myref{Section}{sec:exp_simulated}.

\section{Experiments}\label{sec:exps}

In this section, we show a few examples of evaluating model uncertainty on simulated data as well as models trained on CIFAR-10 \citep{Krizhevsky09CIFAR10} evaluated on CIFAR-10-C \citep{hendrycks2019robustness}\footnote{CC by 4.0}. 
The goal of this section is to give a few examples to show the intuition behind the results and metrics introduced in \myref{Section}{sec:eval_w_ppc} as well as show a few examples where conclusions based on calibration error are incorrect since calibration error is theoretically wrong to use with model uncertainty. We focus on PPCs on calibration error and accuracy to show that, though calibration error by itself is wrong, it can continue to be used in conjunction with PPCs and that, though accuracy is not inherently an uncertainty metric, PPCs with accuracy can give insight into the model's understanding of what it does not know. Future work entails researching the strengths and weaknesses of different test statistics as PPCs currently are often used only in the context of classical Bayesian models. 

Results on a multitude of tabular datasets can be found in \myref{Appendix}{sec:tabular_exps}.

\subsection{Simulated Data}\label{sec:exp_simulated}

To introduce the intuition behind the incorrectness of aiming for zero calibration error (\myref{Equation}{eqn:calib_err}), we evaluate an ensemble of 50 models and an MC Dropout \citep{MCDropout} model on data shown in \myref{Figure}{fig:simulated_data}, where:
\begin{equation*}
\begin{aligned}
   x &\sim \mathcal{N}(0, 1)
\\ y &\sim \mathcal{N}((x - 1)^2, 0.5) \label{eqn:true_sim_y}
\end{aligned}
\end{equation*}
where data such that $x \in [-1.5,0]$ has been removed. The data was removed so that we can compare performance between data from the same distribution as the training data, as well as data from this region removed during training.

We further compare the performance between assuming conditionally independent ensembling (where a model is sampled at random for every $x$) and Bayesian ensembling to validate that Bayesian ensembling both aligns with intuition as well as performs better empirically. 

We first compare qualitatively that the performance of means from samples of models aligns better with assuming Bayesian ensembling versus conditionally independent ensembles. In \myref{Figure}{fig:simulated_data_sparse}, we can see qualitatively that MC Dropout and Bayesian Ensembles encompass the true quadratic function of the data (\myref{Equation}{eqn:true_sim_y}) whereas Conditionally Independent Ensembles does not since the conditional independence has induced a set of noncontinuous functions. Whereas we do expect there to be noise in samples, we do not expect noisiness in the mean.

\begin{figure}[!t]
\begin{minipage}{\linewidth}
\begin{subfigure}{0.3\linewidth}
\centering
\centerline{\includegraphics[width=\columnwidth]{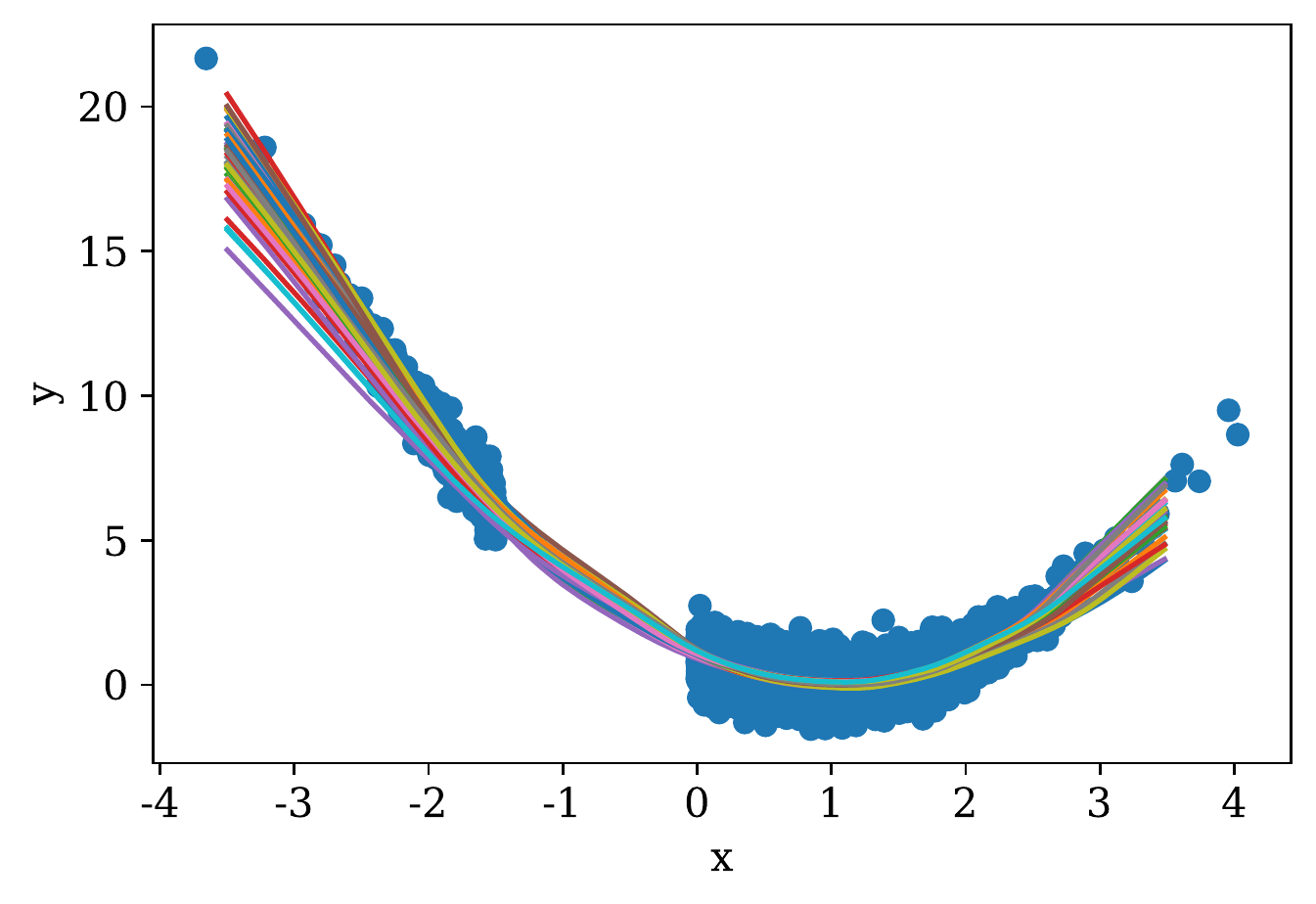}}
\caption{MC Dropout}
\end{subfigure} \hfill
\begin{subfigure}{0.3\linewidth}
\centering
\centerline{\includegraphics[width=\columnwidth]{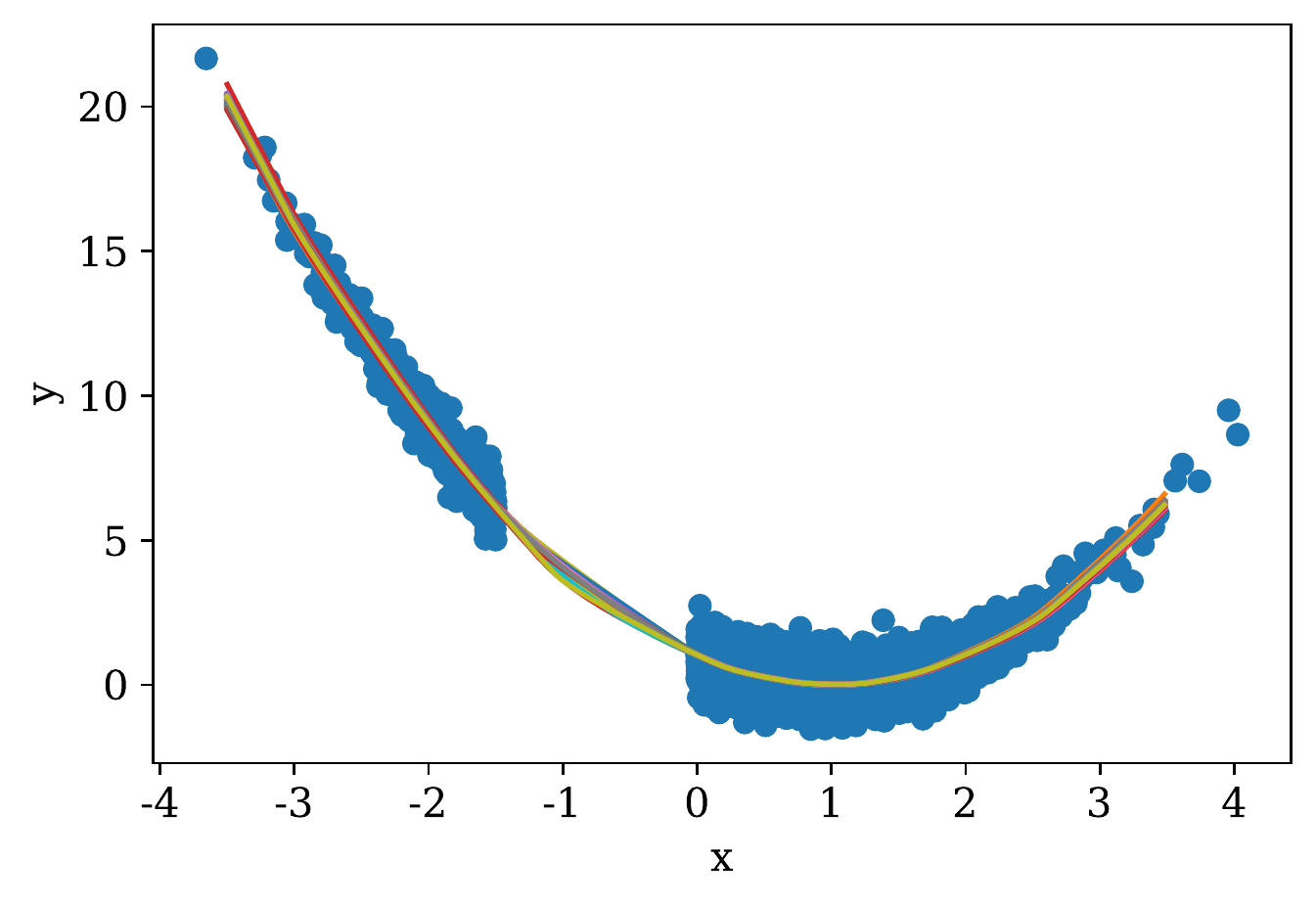}}
\caption{Bayesian Ensemble}
\end{subfigure} \hfill
\begin{subfigure}{0.3\linewidth}
\centering
\centerline{\includegraphics[width=\columnwidth]{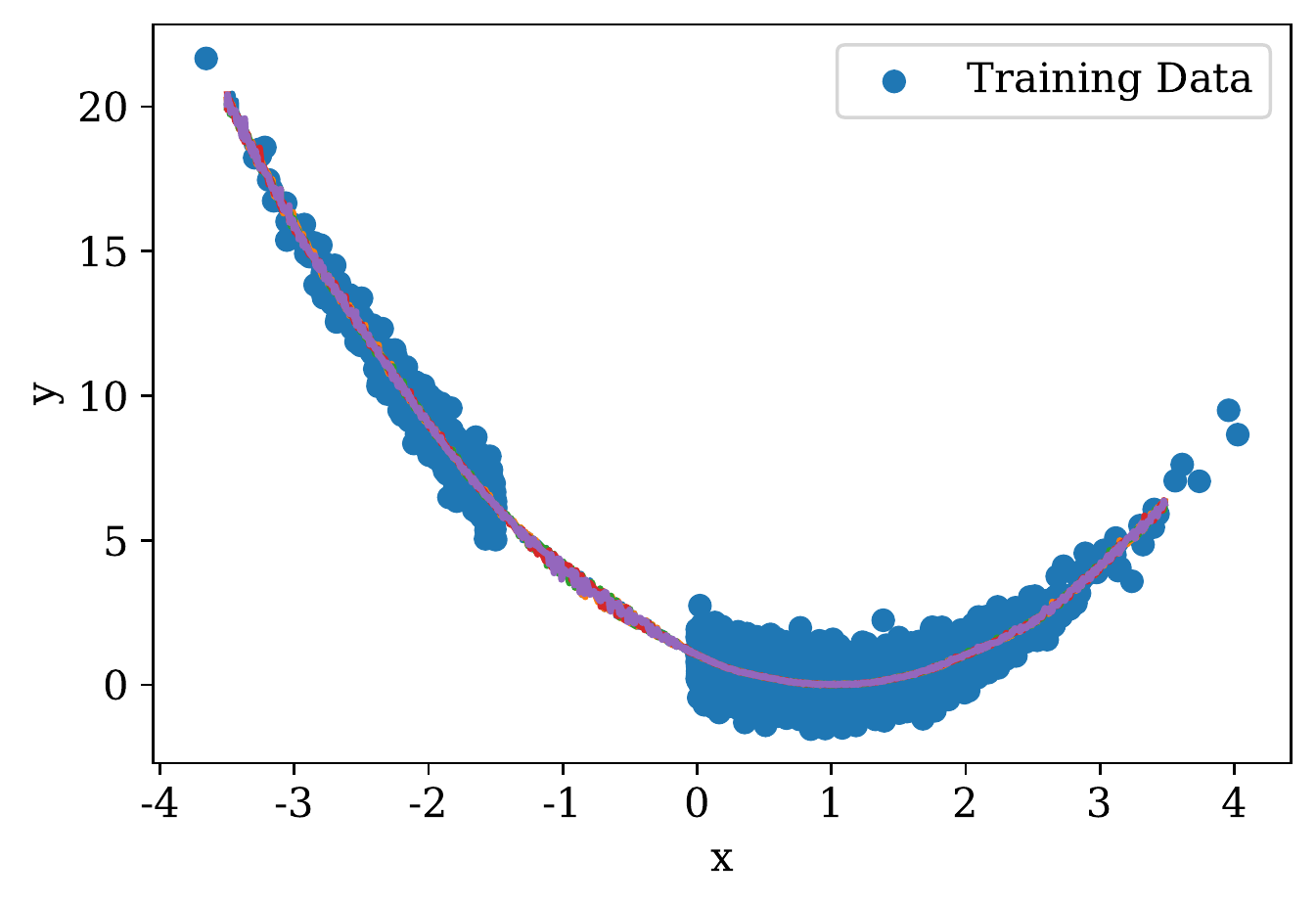}}
\caption{Independent Ensemble}
\end{subfigure}

\caption{Predicted mean from models drawn from the model distribution.}\label{fig:simulated_data}
\end{minipage}

\begin{minipage}{\linewidth}
\begin{subfigure}{0.3\linewidth}
\centering
\centerline{\includegraphics[width=\columnwidth]{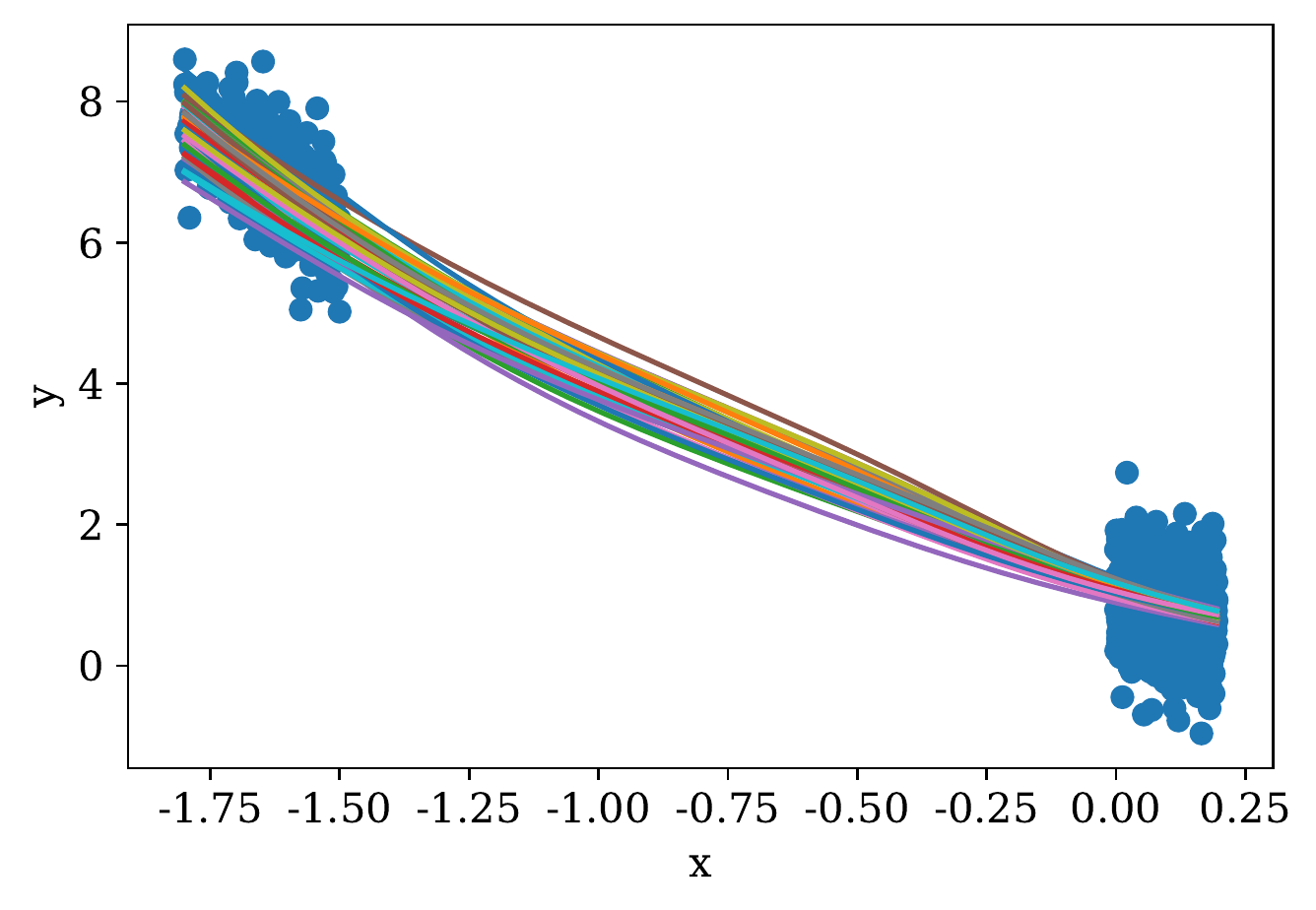}}
\caption{MC Dropout}
\end{subfigure} \hfill
\begin{subfigure}{0.3\linewidth}
\centering
\centerline{\includegraphics[width=\columnwidth]{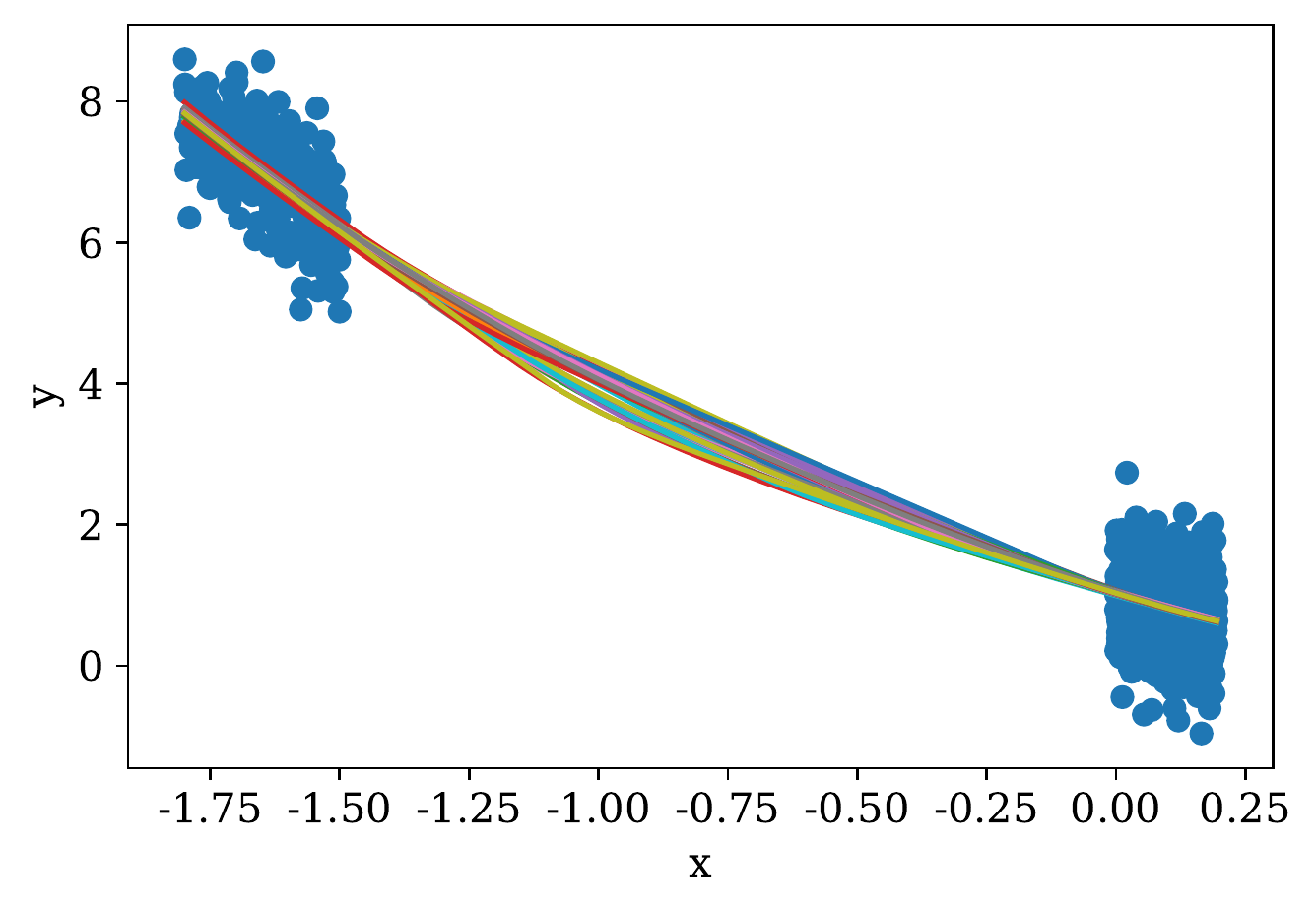}}
\caption{Bayesian Ensemble}
\end{subfigure} \hfill
\begin{subfigure}{0.3\linewidth}
\centering
\centerline{\includegraphics[width=\columnwidth]{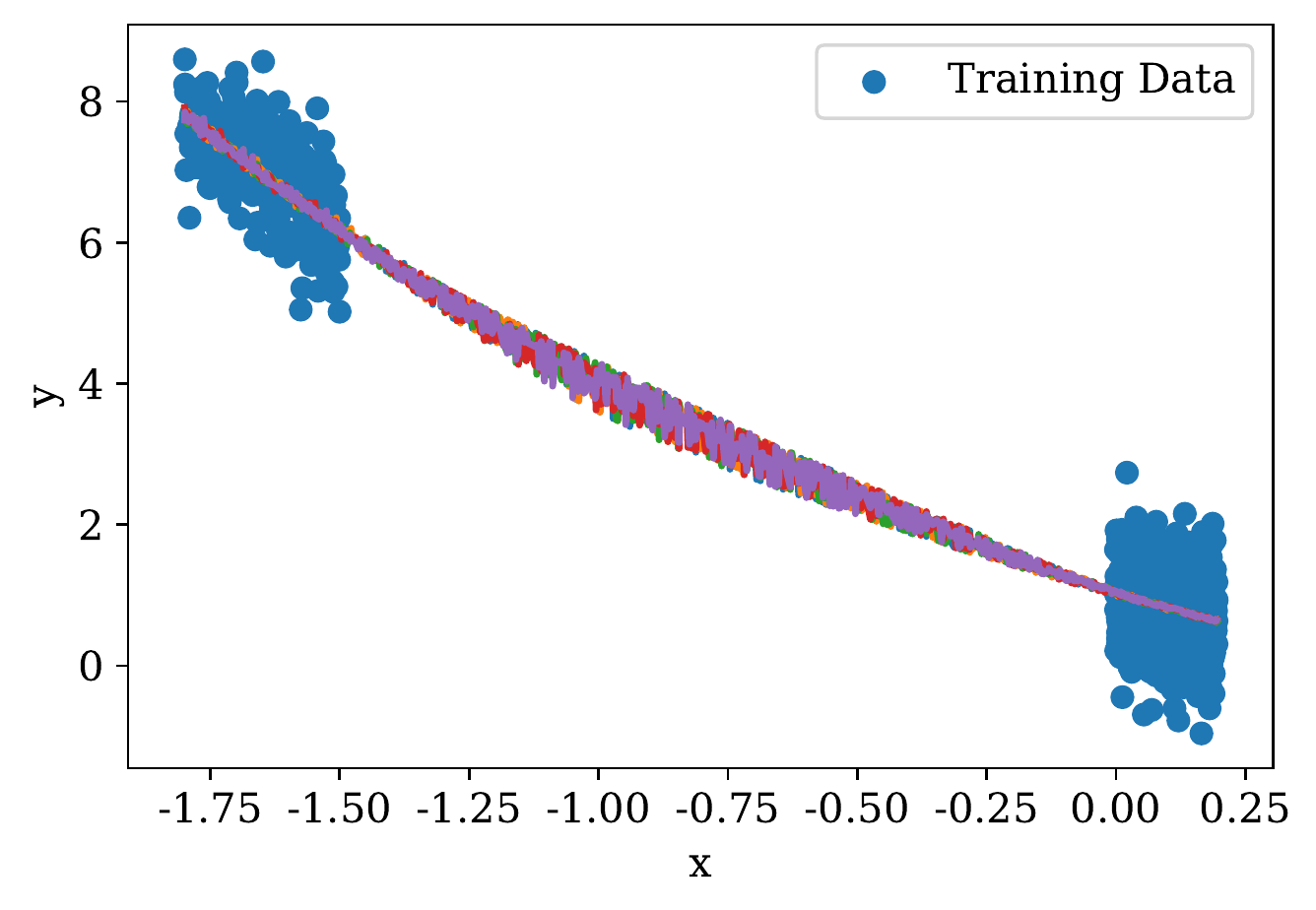}}
\caption{Independent Ensemble}
\label{fig:ind_ens_sim}
\end{subfigure}
\end{minipage}
\caption{Zoomed in version of \myref{Figure}{fig:simulated_data} in the region with no training data.}\label{fig:simulated_data_sparse}
\end{figure}

Quantitatively, we perform a PPC of calibration error (\myref{Equation}{eqn:calib_err}). In \myref{Figure}{fig:sim_ece_is}, we see that though MC Dropout clearly has worse calibration error on i.i.d. test data, it is within what the model expects; thus making MC Dropout not inherently worse than the ensemble. In \myref{Figure}{fig:sim_ece_oos}, we see that for data from the region not observed during training, the (Conditionally) Independent Ensemble fails the check (the observed value is outside the range of the minimum and maximum sampled value).

Even if we were not to know what data the models were trained on, though both Bayesian Ensembles and MC dropout passed the posterior predictive checks since their p-values does not equal zero or one, from \myref{Table}{tbl:sim_data}, since the sharpness of the posterior is smaller for Bayesian Ensembles, we can say that the Bayesian Ensemble is both more confident and understands what it does not know. Further, even though Conditional Independent Ensembles have lower sharpness (i.e. are more confident), since it fails the PPC for out of distribution data (p-value is one), it does not understand what it does not know.

More details on the architectures and on how we created these plots can be found in \myref{Appendix}{sec:appendix_sim}.

\begin{table}[!tb]
\caption{p-value and Sharpness on simulated data, for both in and out of distribution. We say a PPC has passed if the p-value does not equal 0.0 or 1.0. Further, among models which have passed, we prefer models with smaller sharpness as this implies our model has more confidence in its predictions.}\label{tbl:sim_data}
\centering
\begin{tabular}{lcccc}
\toprule
& \multicolumn{2}{c}{In Distribution}
& \multicolumn{2}{c}{Out of Distribution}
\\
 \cmidrule(lr){2-3}
 \cmidrule(lr){4-5} 
& {p-value} & {Sharpness} & {p-value} & {Sharpness} \\
\midrule
Independent Ensemble  & $0.088$  & $0.002 $  & $1.0$ & $0.001$ \\ 
Bayesian Ensemble & $ 0.034 $  & $ 0.008 $   & $ 0.197 $& $0.396$ \\ 
MC Dropout & $ 0.227 $  & $ 0.057 $   & $ 0.186 $ & $1.319$ \\ 
\bottomrule
\end{tabular}
\end{table}
\begin{figure}[!tb]

\begin{subfigure}{0.45\linewidth}
\centering
\centerline{\includegraphics[width=\columnwidth]{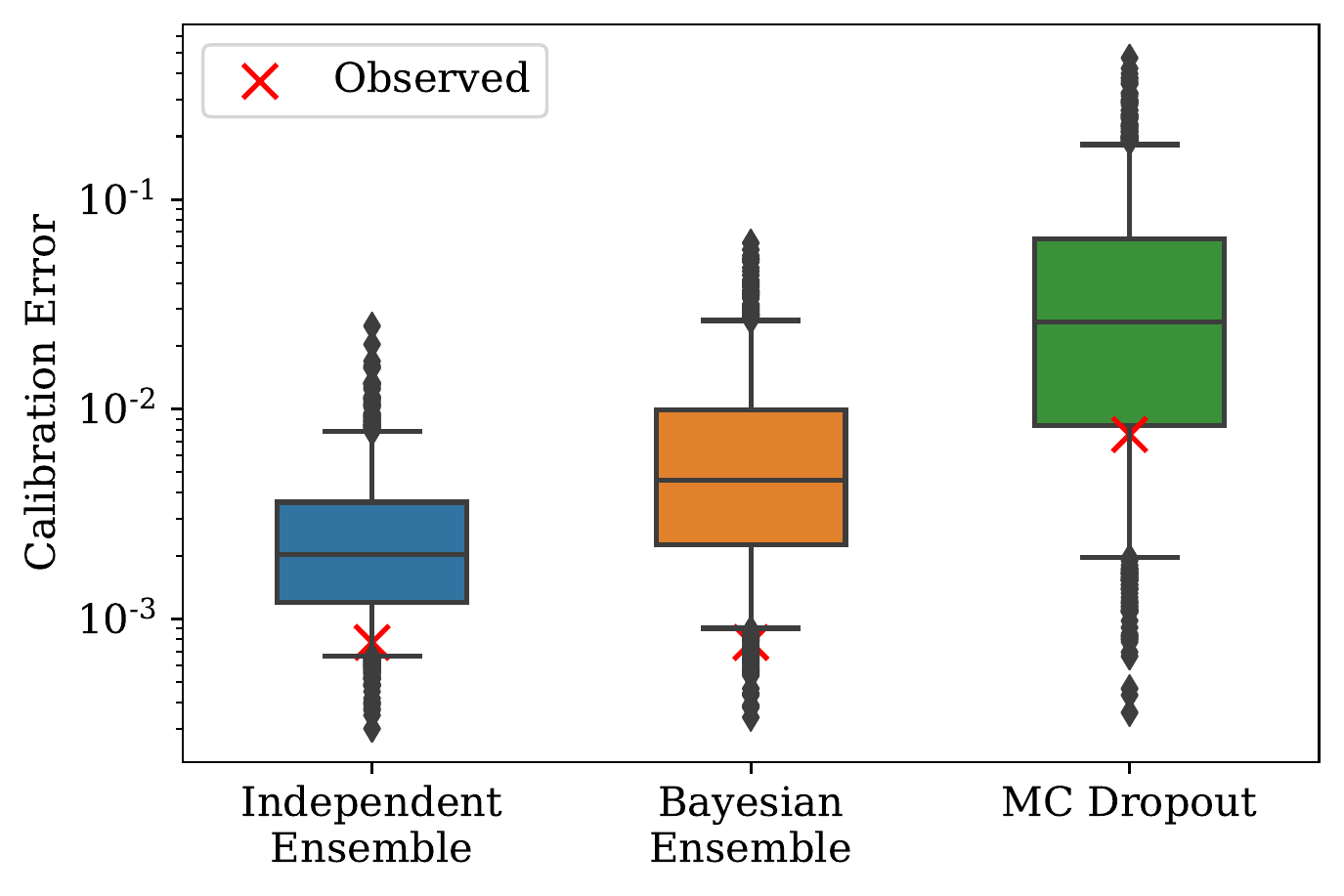}}
\caption{In Distribution}
\label{fig:sim_ece_is}
\end{subfigure} \hfill
\begin{subfigure}{0.45\linewidth}
\centering
\centerline{\includegraphics[width=\columnwidth]{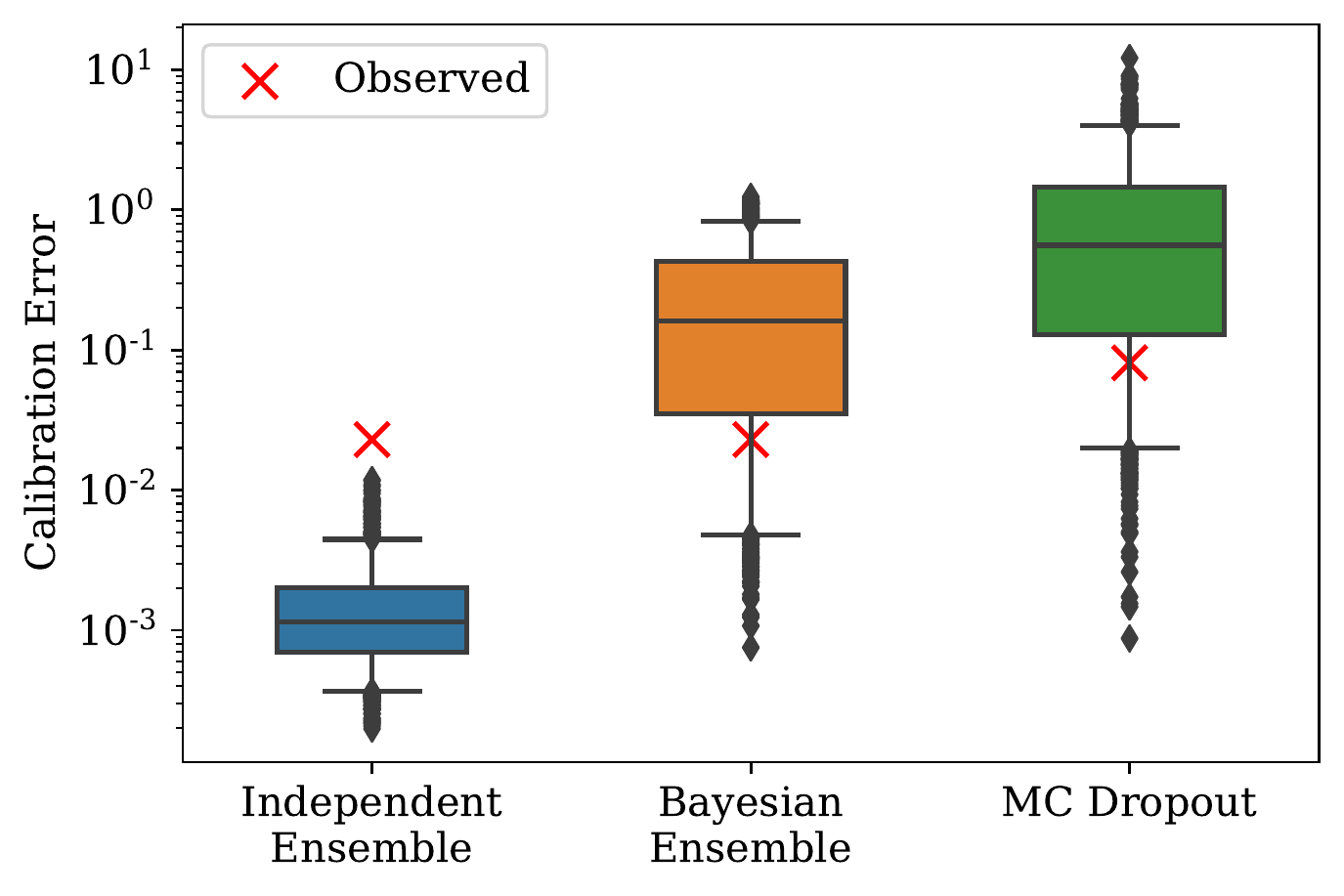}}
\caption{Out of Distribution}
\label{fig:sim_ece_oos}
\end{subfigure}
\caption{Posterior predictive checks for calibration error on i.i.d. data and data from the region with no training data. In the box plot, the box spans the 25th and 75th percentile and the whiskers represent the 5th and 95th percentile of the samples of the calibration error using \myref{Algorithm}{alg:ppc}. The red $X$ is the observed calibration error (on true data) given the model.}\label{fig:sim_ece}
\end{figure}

\subsection{Distribution Shift for Image Classification}\label{sec:exp_images}

In \myref{Section}{sec:exp_simulated}, we found that models can have non-zero calibration errors yet be expected by the model; in this section, we test to see if the same phenomenon is found in image classification models. In \citet{Ovadia2019}, a thorough study was conducted on the performance of multiple models on multiple different datasets with distribution shift, amongst which was CIFAR-10. However, many conclusions were made based on the degradation of ECE (\myref{Equation}{eqn:ece}) on different severities of image corruption; more specifically, if the ECE increased, it was interpreted as the model not knowing what it does not know. In this section, we re-evaluate the ensemble and MC Dropout CIFAR-10 models trained in \citet{Ovadia2019}\footnote{\href{https://github.com/google-research/google-research/tree/master/uq_benchmark_2019}{https://github.com/google-research/google-research/tree/master/uq\_benchmark\_2019}: Apache License, Version 2.0} to test how well the models can account for the degradation in both ECE and accuracy.

Taking inspiration from \citet{Ashukha2020Pitfalls}, we recalibrated the ensembles and MC Dropout models as we found this helped improve performance in terms of PPC. More details on the methodology and how we ensured no data leakage can be found in \myref{Appendix}{sec:appendix_recalib}.

\begin{figure}[!bt]
\begin{subfigure}{0.45\linewidth}
\centering
\centerline{\includegraphics[width=\columnwidth]{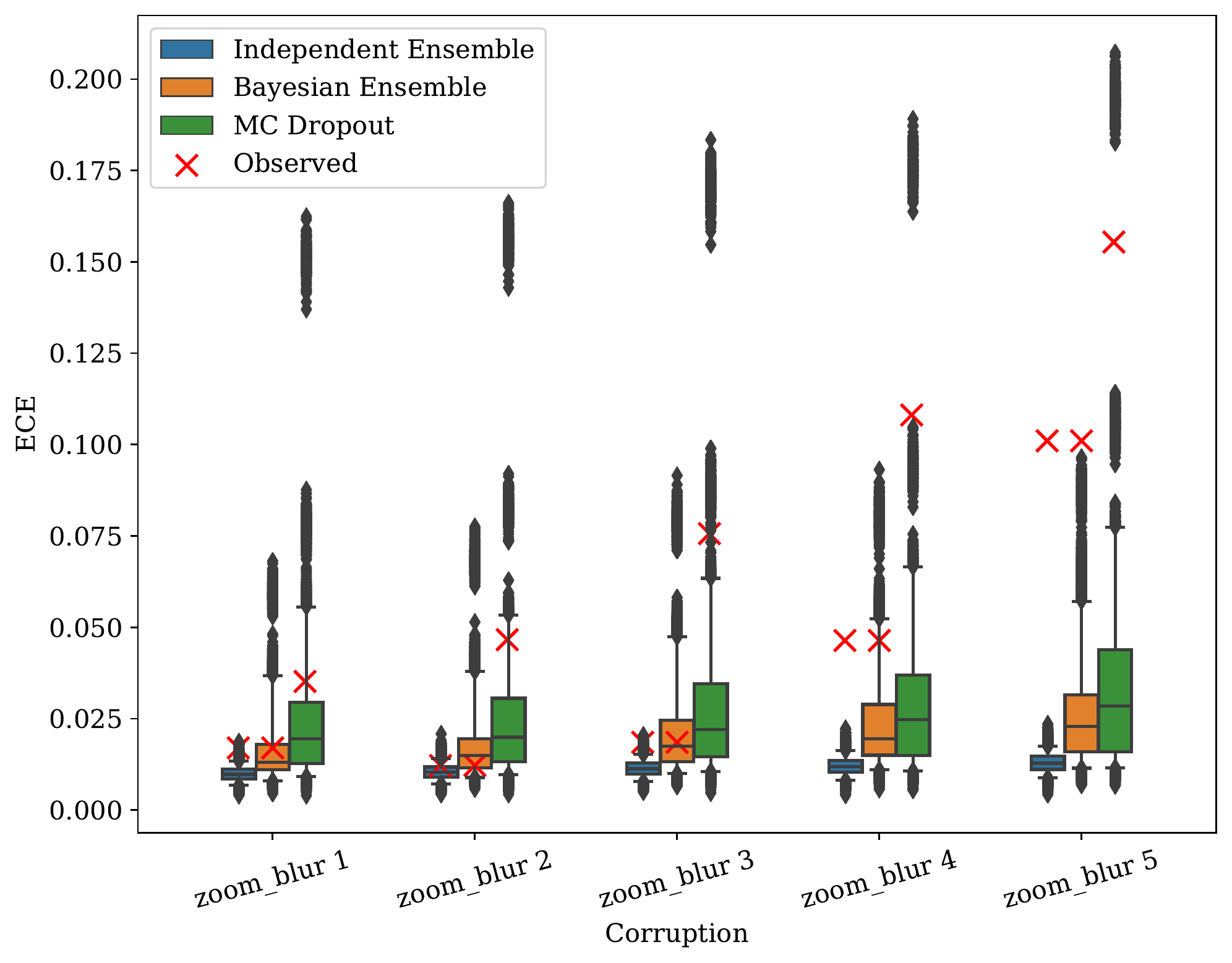}}
\end{subfigure} \hfill
\begin{subfigure}{0.45\linewidth}
\centering
\centerline{\includegraphics[width=\columnwidth]{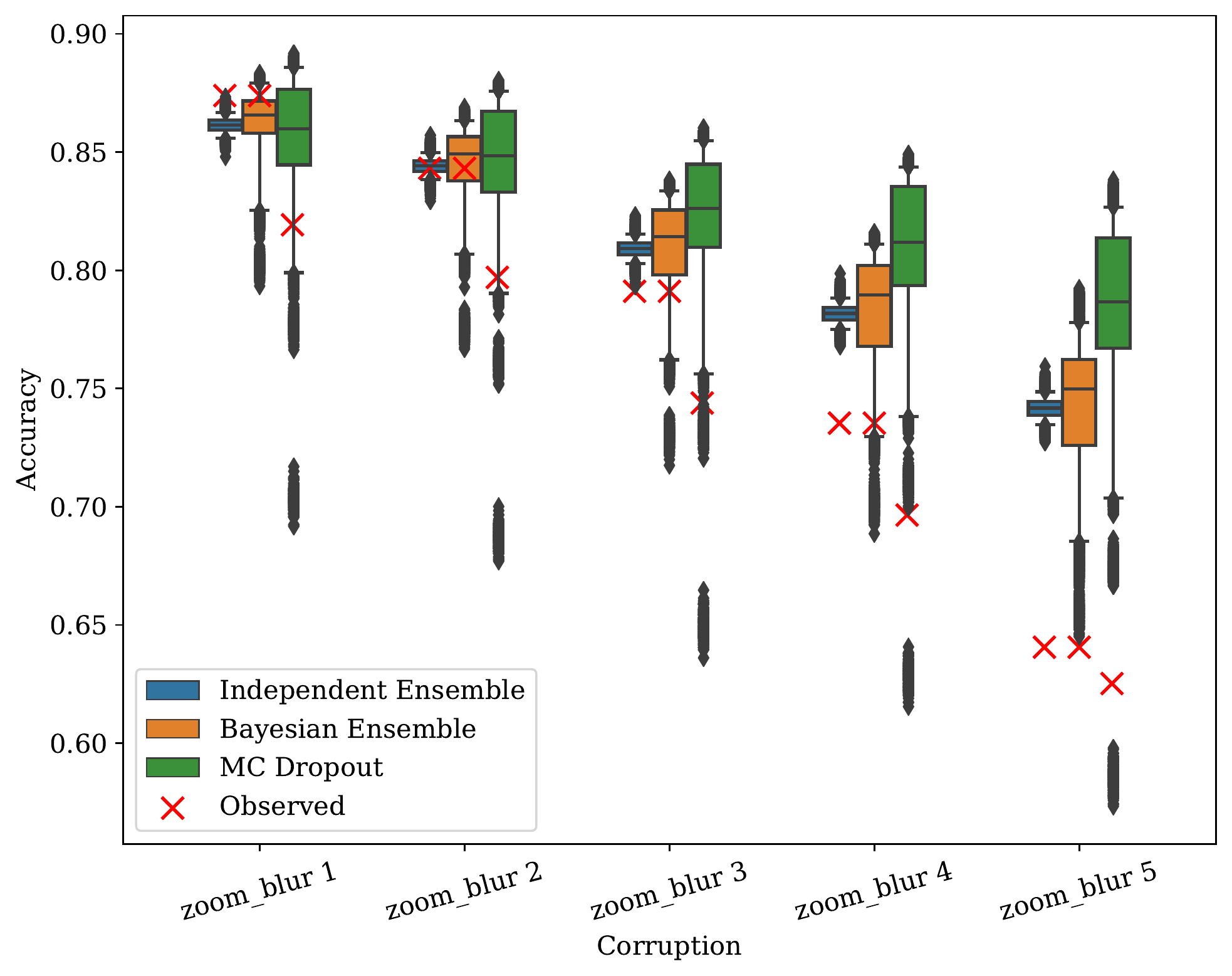}}
\end{subfigure}
\caption{Posterior predictive checks for ECE and accuracy of recalibrated models trained on CIFAR-10, evaluated on Zoom Blur data from CIFAR-10-C. In the box plot, the box spans the 25th and 75th percentile and the whiskers represent the 5th and 95th percentile of the samples of the test statistic using \myref{Algorithm}{alg:ppc}. The red $X$ is the test statistic computed on the true data.}\label{fig:zoom_blur}
\end{figure}

\begin{figure}[!b]
\begin{subfigure}{0.45\linewidth}
\centering
\centerline{\includegraphics[width=\columnwidth]{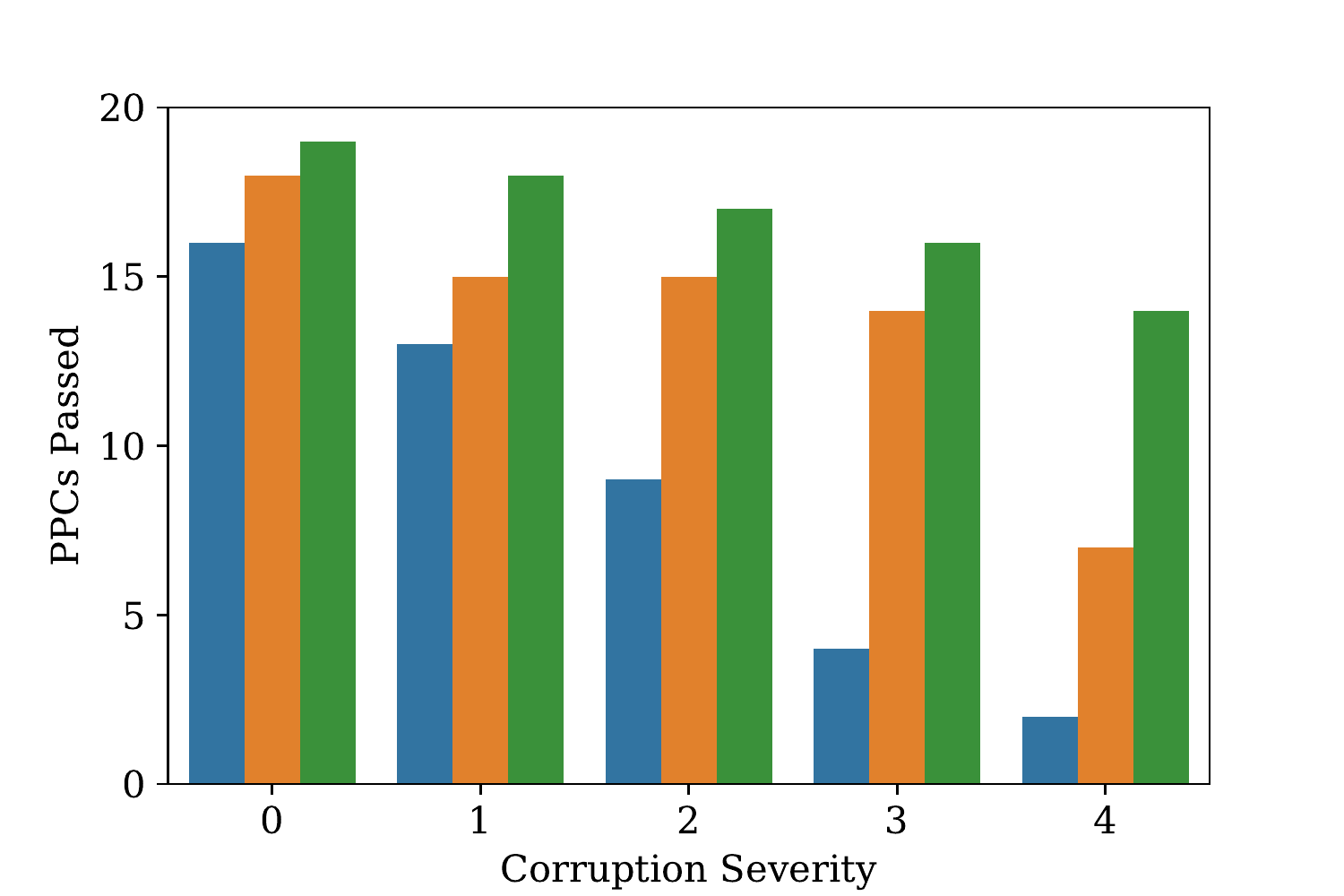}}
\caption{ECE}
\end{subfigure} \hfill
\begin{subfigure}{0.45\linewidth}
\centering
\centerline{\includegraphics[width=\columnwidth]{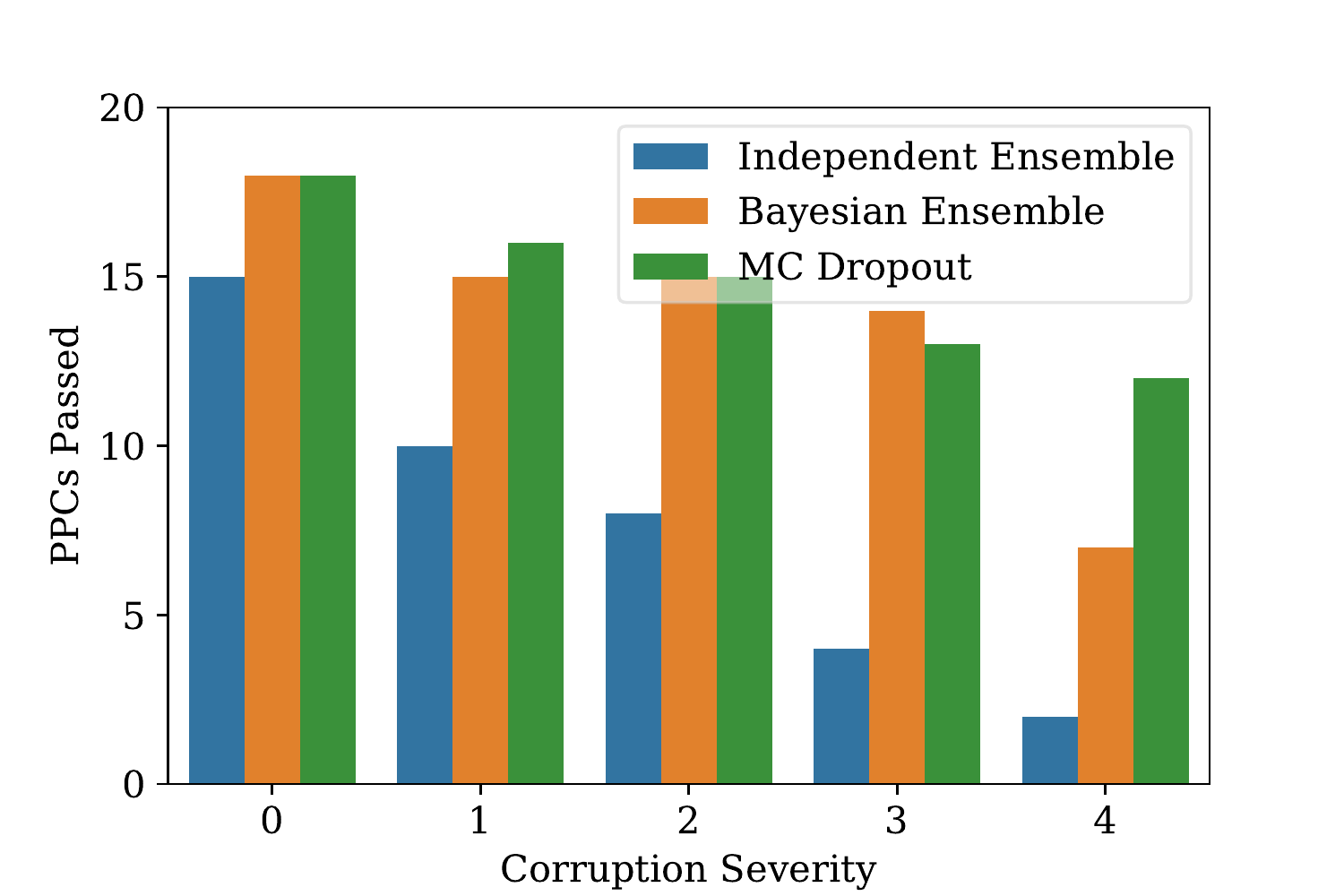}}
\caption{Accuracy}
\end{subfigure}
\caption{For each model, we plot how many PPCs for different corruptions at a specific severity passed. A PPC passes if the observed value falls within the range defined by the samples of the test statistic (p-value is not zero or one).}\label{fig:summary_cifar10c}
\end{figure}

Since CIFAR-10-C has 19 different image corruptions, each with five different severities, we leave all the posterior predictive checks to \myref{Appendix}{sec:cifar_10_c_full}; however, \myref{Figure}{fig:zoom_blur} shows how well the models perform on the zoom blur corruption, across the five severities. Though the example we show here has the models perform reasonably well on all the severities, there are corruptions such as shot noise and gaussian noise where the models are unable to account for the distribution shift.

We summarize the performance of the three models across the different corruption types and severities in \myref{Figure}{fig:summary_cifar10c} where we plot the number of posterior predictive checks passed, where passing means the observed value falls within the samples of the test statistic (empirical p-value is not zero or one). The exact number of tests passed can be found in the \myref{Appendix}{sec:addl_plot} (\myref{Table}{tbl:summary_cifar10c}). We can see that the Bayesian Ensemble interpretation of ensembling leads to more tests passed and that MC Dropout seems to be able to account for the distribution shift.

The most important conclusion from \myref{Figure}{fig:zoom_blur} is that, though ECE gets significantly larger as the severity increases, Bayesian ensembling and MC Dropout are able to account for such shifts since the observed value falls within the range defined by the samples of the test statistic.
Thus saying that if we were to go by ECE alone, we would assume the models do not understand what they do not know. However, via posterior predictive checks, we can see this conclusion is incorrect and these models are better than what would have been expected. Similarly, accuracy degrades but the posterior predictive distribution shifts and grows acccordingly.

\section{Conclusion}

Though we do not compare against other methods where the generative process assumes conditionally independent model uncertainty, the results we show in \myref{Section}{sec:exps} suggests that this assumption does not perform well for ensembles; we leave it to future work to evaluate other models where this assumption has been made.

An important note here is that this method of evaluation is more expensive than the current methods and might have negative environmental impact when evaluating larger models. However, we were able to get all our results in this paper without any GPUs; further, posterior predictive checks are theoretically sound whereas the other evaluation methods are not.

Similar to how an increase in aleatoric uncertainty can lead to an increase in expected model error, an increase in model uncertainty can lead to an increase in expected calibration error.
Though we show in this paper that calibration error is incorrect when evaluating model uncertainty, good calibration error can still be a goal in the same way accuracy is a goal even though higher aleatoric uncertainty would lead to worse expected accuracy.
Further, we show that if we want a model that gives both aleatoric and model uncertainty without giving access to a decomposition of the two, calibration error is not a method to verify the two have been modeled correctly. 

Thus, from our experimental section, we see that if we were to go by ECE alone, we would assume the models such as in \myref{Section}{sec:exp_images} do not understand what they do not know. However, via posterior predictive checks, we can see this conclusion is incorrect and these models are better than expected. Equally importantly, our experimental section shows that incorrect assumptions such as conditionally independent model uncertainty can fail posterior predictive checks though they might have qualitatively good calibration errors, giving false trust in the model.
We see future work from this paper to be a re-evaluation of model uncertainty techniques and checking where incorrect conclusions have been made due to an incorrect handling of model uncertainty.

\newpage
\bibliographystyle{chicago}
\bibliography{./biblio}
\newpage

\input{appendix}

%% file: appendix.tex
\appendix

\section{Tabular Experiments}\label{sec:tabular_exps}

In this section, we present the results on tabular experiments. Similar to the image experiment results in \myref{Section}{sec:exp_images}, in this section, we show for some tabular datasets where calibration error and PICP give incorrect (and even opposite) conclusions as compared to PPCs.

We experiment on eight regression UCI datasets \citep{UCIDatasets}, based on the experiments run by \citet{SQR}. However, we focused on Conditional Gaussian \citep{DeepEnsembles}, Bayesian Ridge Regression \citep{BayesianRidgeRegression}, MC Dropout \citep{MCDropout}, and Simultaneous Quantile Regression (SQR) \citep{SQR}. We chose to not analyze the other models compared in the original paper (e.g. Quality Driven \citep{QualityDrivenLoss2018}) as the others give results for a few quantiles but not all, i.e. we cannot sample, thus precluding us from using PPCs.
However, we additionally add Conditional Residual Flows \citep{Chen2019ResidualFlows} (a specific type of normalizing flow) as a simple generalization of Conditional Gaussian that allows for more complex output distributions. 

For our experiments, we performed a hyperparameter tuning for each model, choosing the model with the best average validation loss across 20 seeds. For each seed, we split the dataset into three sets: 72\% for training, 18\% for validation, and 10\% for test. For the results in \myref{Table}{tbl:calibration_v1} and \myref{Table}{tbl:calibration_v2}, we evaluate the metrics per seed and show the mean and standard deviation of each metric across the 20 seeds.

\begin{table*}
\caption{Results on percentage of data captured from 0.025 to 0.975 quantiles (\textit{PICP}), mean squared error (\textit{MSE}), and calibration error (\textit{Calib}) with 100 buckets. The results are from minimizing the average validation loss across 20 seeds.  The optimal value for MSE is 0; the optimal values for calibration error and PICP, assuming no model uncertainty, is 0 and 0.95, respectively. To allow for comparison to MC Dropout (a model with model uncertainty), for each model, we give a count of how many PPCs across the 20 seeds passed for PICP (\textit{PICP PPC}) and calibration error (\textit{Calib PPC}). A PPC passes if the observed value falls within the range of 0.025 to 0.975 quantiles (95\% confidence interval). The number next to each dataset is the size of the test set. The number in parenthesis for the PPC Counts (for MC Dropout) is the number of tests that would have passed had we treated the model uncertainty as conditionally independent (\myref{Section}{sec:ensembles}).}\label{tbl:calibration_v1}
\small
\centering
\begin{tabular}{lcclcl}
\toprule
    & \multicolumn{5}{c}{bostonHousing (51)}
\\
 \cmidrule(lr){2-6} 
 & {MSE} & {PICP}& {PICP PPC} & {Calib}  & {Calib PPC}  
\\
\midrule
ConditionalFlow  & $0.13 \pm 0.06$  & $0.93 \pm 0.04$  & $14$  & $0.3 \pm 0.3$  & $19$  \\ 
ConditionalGaussian  & $0.14 \pm 0.07$  & $0.93 \pm 0.03$  & $17$  & $0.4 \pm 0.3$  & $19$  \\ 
BayesianRidgeRegression  & $0.28 \pm 0.08$  & $0.95 \pm 0.03$  & $19$  & $0.6 \pm 0.4$  & $19$  \\ 
MC Dropout  & $0.11 \pm 0.05$  & $0.50 \pm 0.07$  & $2 (0)$  & $3.4 \pm 0.8$  & $20 (0)$  \\ 
SQR  & $0.11 \pm 0.05$  & $0.73 \pm 0.08$  & $0$  & $1.0 \pm 0.6$  & $12$  \\ 
\bottomrule
\toprule
     & \multicolumn{5}{c}{concrete (103)}
\\
 \cmidrule(lr){2-6} 
 & {MSE} & {PICP}& {PICP PPC} & {Calib}  & {Calib PPC}  
\\
\midrule
ConditionalFlow  & $0.09 \pm 0.03$  & $0.92 \pm 0.03$  & $13$  & $0.2 \pm 0.2$  & $19$  \\ 
ConditionalGaussian  & $0.09 \pm 0.03$  & $0.91 \pm 0.02$  & $13$  & $0.4 \pm 0.4$  & $16$  \\ 
BayesianRidgeRegression  & $0.40 \pm 0.07$  & $0.95 \pm 0.03$  & $17$  & $0.18 \pm 0.08$  & $20$  \\ 
MC Dropout  & $0.07 \pm 0.03$  & $0.69 \pm 0.05$  & $18 (0)$  & $1.1 \pm 0.6$  & $20 (2)$  \\ 
SQR  & $0.07 \pm 0.03$  & $0.74 \pm 0.04$  & $0$  & $0.7 \pm 0.4$  & $9$  \\ 
\bottomrule
\toprule
     & \multicolumn{5}{c}{energy (77)}
\\
 \cmidrule(lr){2-6} 
 & {MSE} & {PICP}& {PICP PPC} & {Calib}  & {Calib PPC}  
\\
\midrule
ConditionalFlow  & $0.0022 \pm 0.0006$  & $0.96 \pm 0.02$  & $18$  & $0.4 \pm 0.4$  & $18$  \\ 
ConditionalGaussian  & $0.0030 \pm 0.0009$  & $0.97 \pm 0.03$  & $16$  & $0.6 \pm 0.3$  & $15$  \\ 
BayesianRidgeRegression  & $0.09 \pm 0.02$  & $0.89 \pm 0.04$  & $7$  & $0.5 \pm 0.2$  & $16$  \\ 
MC Dropout  & $0.0026 \pm 0.0010$  & $0.97 \pm 0.02$  & $ 15 (18)$  & $0.6 \pm 0.5$  & $1 (13)$  \\ 
SQR  & $0.0028 \pm 0.0012$  & $0.97 \pm 0.03$  & $17$  & $0.9 \pm 0.6$  & $13$  \\ 
\bottomrule
\toprule
     & \multicolumn{5}{c}{kin8nm (820)}
\\
 \cmidrule(lr){2-6} 
 & {MSE} & {PICP}& {PICP PPC} & {Calib}  & {Calib PPC}  
\\
\midrule
ConditionalFlow  & $0.081 \pm 0.013$  & $0.952 \pm 0.006$  & $20$  & $0.11 \pm 0.11$  & $12$  \\ 
ConditionalGaussian  & $0.071 \pm 0.005$  & $0.947 \pm 0.007$  & $19$  & $0.04 \pm 0.04$  & $17$  \\ 
BayesianRidgeRegression  & $0.59 \pm 0.02$  & $0.951 \pm 0.007$  & $20$  & $0.12 \pm 0.09$  & $8$  \\ 
MC Dropout  & $0.069 \pm 0.005$  & $0.69 \pm 0.02$  & $4 (0)$  & $1.3 \pm 0.3$  & $20 (0)$  \\ 
SQR  & $0.072 \pm 0.005$  & $0.892 \pm 0.011$  & $0$  & $0.04 \pm 0.02$  & $18$  \\ 
\bottomrule
\end{tabular}
\end{table*}

\begin{table*}
\caption{Results on percentage of data captured from 0.025 to 0.975 quantiles (\textit{PICP}), mean squared error (\textit{MSE}), and calibration error (\textit{Calib}) with 100 buckets. The results are from minimizing the average validation loss across 20 seeds.  The optimal value for MSE is 0; the optimal values for calibration error and PICP, assuming no model uncertainty, is 0 and 0.95, respectively. To allow for comparison to MC Dropout (a model with model uncertainty), for each model, we give a count of how many PPCs across the 20 seeds passed for PICP (\textit{PICP PPC}) and calibration error (\textit{Calib PPC}). A PPC passes if the observed value falls within the range of 0.025 to 0.975 quantiles (95\% confidence interval). The number next to each dataset is the size of the test set. The number in parenthesis for the PPC Counts (for Dropout) is the number of tests that would have passed had we treated the model uncertainty as conditionally independent (\myref{Section}{sec:ensembles}).}\label{tbl:calibration_v2}
\small
\centering
\begin{tabular}{lcclcl}
\toprule
     & \multicolumn{5}{c}{naval-propulsion-plant (1194)}
\\
 \cmidrule(lr){2-6} 
 & {MSE} & {PICP}& {PICP PPC} & {Calib}  & {Calib PPC}  
\\
\midrule
ConditionalFlow  & $0.009 \pm 0.004$  & $0.98 \pm 0.03$  & $0$  & $1.6 \pm 1.6$  & $0$  \\ 
ConditionalGaussian  & $0.012 \pm 0.006$  & $0.9995 \pm 0.0007$  & $0$  & $2.5 \pm 0.3$  & $0$  \\ 
BayesianRidgeRegression  & $0.055 \pm 0.004$  & $1.0 \pm 0$  & $0$  & $8.033 \pm 0.006$  & $0$  \\ 
MC Dropout  & $0.005 \pm 0.002$  & $0.98 \pm 0.02$  & $1 (0)$  & $1.3 \pm 0.8$  & $14 (0)$  \\ 
SQR  & $0.0052 \pm 0.0012$  & $0.998 \pm 0.002$  & $0$  & $1.7 \pm 0.6$  & $0$  \\ 
\bottomrule
\toprule
     & \multicolumn{5}{c}{power-plant (957)}
\\
 \cmidrule(lr){2-6} 
 & {MSE} & {PICP}& {PICP PPC} & {Calib}  & {Calib PPC}  
\\
\midrule
ConditionalFlow  & $0.048 \pm 0.007$  & $0.952 \pm 0.007$  & $19$  & $0.05 \pm 0.05$  & $14$  \\ 
ConditionalGaussian  & $0.048 \pm 0.006$  & $0.947 \pm 0.010$  & $15$  & $0.06 \pm 0.10$  & $15$  \\ 
BayesianRidgeRegression  & $0.072 \pm 0.007$  & $0.964 \pm 0.007$  & $8$  & $0.03 \pm 0.03$  & $19$  \\ 
MC Dropout  & $0.046 \pm 0.007$  & $0.59 \pm 0.02$  & $20 (0)$  & $2.0 \pm 0.3$  & $20 (0)$  \\ 
SQR  & $0.048 \pm 0.007$  & $0.93 \pm 0.02$  & $10$  & $0.2 \pm 0.2$  & $5$  \\ 
\bottomrule
\toprule
     & \multicolumn{5}{c}{wine-quality-red (160)}
\\
 \cmidrule(lr){2-6} 
 & {MSE} & {PICP}& {PICP PPC} & {Calib}  & {Calib PPC}  
\\
\midrule
ConditionalFlow  & $0.6 \pm 0.2$  & $0.90 \pm 0.02$  & $5$  & $0.24 \pm 0.14$  & $17$  \\ 
ConditionalGaussian  & $0.59 \pm 0.09$  & $0.95 \pm 0.02$  & $17$  & $0.4 \pm 0.2$  & $12$  \\ 
BayesianRidgeRegression  & $0.63 \pm 0.11$  & $0.95 \pm 0.02$  & $18$  & $0.18 \pm 0.13$  & $18$  \\ 
MC Dropout  & $0.57 \pm 0.09$  & $0.36 \pm 0.04$  & $0 (0)$  & $4.3 \pm 0.6$  & $20 (0)$  \\ 
SQR  & $0.59 \pm 0.10$  & $0.81 \pm 0.09$  & $1$  & $1.0 \pm 0.7$  & $3$  \\ 
\bottomrule
\toprule
     & \multicolumn{5}{c}{yacht (31)}
\\
 \cmidrule(lr){2-6} 
 & {MSE} & {PICP}& {PICP PPC} & {Calib}  & {Calib PPC}  
\\
\midrule
ConditionalFlow  & $0.004 \pm 0.003$  & $0.95 \pm 0.04$  & $18$  & $1.4 \pm 1.4$  & $15$  \\ 
ConditionalGaussian  & $0.004 \pm 0.002$  & $0.99 \pm 0.02$  & $20$  & $2.0 \pm 1.4$  & $12$  \\ 
BayesianRidgeRegression  & $0.35 \pm 0.10$  & $0.94 \pm 0.03$  & $19$  & $0.6 \pm 0.4$  & $20$  \\ 
MC Dropout  & $0.004 \pm 0.002$  & $0.96 \pm 0.04$  & $8 (19)$  & $2.8 \pm 2.1$  & $8 (10)$  \\ 
SQR  & $0.004 \pm 0.004$  & $0.96 \pm 0.04$  & $19$  & $1.9 \pm 1.9$  & $12$  \\ 
\bottomrule
\end{tabular}
\end{table*}

Importantly, though PICP and calibration error are both correct to use for most of these models, according to the results of our paper (Section 3.1 and Section 3.2), using PICP and calibration error for MC Dropout is incorrect since MC Dropout is modeling model uncertainty, not simply aleatoric uncertainty. We summarize the results of our experiments below; in the results, we use the phrasing ``what the model expects'' to mean the posterior predictive distribution, e.g. ``the observed value is larger than what the model expects'' means ``the observed value is considered a (right-)tail event with respect to the model posterior predictive distribution.'' 

\begin{itemize}

  \item Since PICP and calibration error are correct to use for models without model uncertainty, ordering by the distance from the optimal value aligns with sorting by the number of PPCs passed for the metric. For example, sorting by calibration error and sorting by the number of PPCs passed for calibration give similar results, \textit{except for MC Dropout (a model with model uncertainty)}

  \item For the datasets concrete, kin8nm, power-plant, and wine-quality-red, the calibration error of MC Dropout was the worst amongst all the other models (with respect to the incorrect goal of aiming for zero calibration error for MC Dropout). However, with respect to the PPCs, the observed values for calibration error are expected by the model, much more than the calibration error obtained by the other models on these four datasets.

  \item Concrete and power-plant are interesting as, according to both PICP and calibration error, dropout is the worst. However, with respect to the PPCs, the observed values for PICP and calibration error are expected by the model, much more than the PICPs and calibration error obtained by the other models. However, for the other two datasets (kin8nm and wine-quality-red) where calibration error is large but expected by the model, the observed PICP is weak with respect to both the classical method of evaluation and the number of PPCs passed.

  \item Energy is interesting with respect to calibration error as, though MC Dropout got a lower calibration error than SQR, it passed less tests suggesting that the \textit{low error observed is smaller than what MC Dropout was expecting} (possibly due to too much model uncertainty, i.e. too underconfident). In other words, calibration error would lead one to believe MC Dropout is better calibrated than SQR on energy whereas the opposite is true.

  \item Yacht is interesting since the results seem strong for MC Dropout with respect to PICP (using the incorrect way to analyze the number). However, once we look at the number of tests passed for PICP, it is less than all the other models; suggesting, with respect to PICP, MC Dropout is worse than SQR though they have extremely similar PICPs. 
\end{itemize}

In conclusion, given these results, for MC Dropout (i.e. a model modeling model uncertainty), we find scenarios in which:
\begin{itemize}
  \item the calibration error can be low but still not expected by the model, i.e. \textit{a bad model that calibration error would say is good}
  \item the calibration error can be high but still within what the model expects, i.e. \textit{a good model that calibration error would say is bad}
\end{itemize}

\subsection{Model Training Details}

\begin{figure*}
\centering
\begin{tikzpicture}[
  every neuron/.style={
    circle,
    minimum size=0.3cm,
    thick
  },
  every data/.style={
    rectangle,
    minimum size=0.4cm,
    thick
  },
]

  \node [align=center,every neuron/.try, data 1/.try, minimum width=0.3cm, draw] (input-x)  at ($ (2.5, 0) $) {$\mathbf{X}$};
  \node [align=center,every data 1/.try, minimum width=1.0cm,draw] (nn)  at ($ (input-x) + (1.5, 0) $) {NN};

  \node [align=center,every neuron/.try, data 1/.try, minimum width=0.3cm, draw] (input-y)  at ($ (input-x) + (1.0, 1.5) $) {$\mathbf{y}$};
  \node [align=center,every data 1/.try, minimum width=1.5cm,draw] (affine)  at ($ (input-y) + (2.0, 0) $) {Affine};

  \node [align=center,every data 1/.try, minimum width=1.5cm,draw] (cquar)  at ($ (affine) + (2.0, 0) $) {Residual};
  \node [align=center,every data 1/.try, minimum width=1.5cm,draw] (actnorm)  at ($ (cquar) + (2.0, 0) $) {Affine};
  \node [align=center,every neuron/.try, data 1/.try, minimum width=0.3cm, draw] (output)  at ($ (actnorm) + (2.0, 0) $) {$\mathbf{z}$};

  \node [align=center, every data 1/.try] (scale-box)  at ($ (affine) - (0.35, 1.5) $) {};
  \node [align=center, every data 1/.try] (shift-box)  at ($ (scale-box) + (1.0, 0) $) {};
  \node [align=center, every data 1/.try] (h-box)  at ($ (cquar) - (0.1, 1.5) $) {};

  \draw [black,solid,->] ($(input-x.east)$) -- ($(nn.west)$);
  \draw [black,solid,->] ($(input-y.east)$) -- ($(affine.west)$);
  \draw [black,solid,->] ($(affine.east)$) -- ($(cquar.west)$);
  \draw [black,solid,->] ($(cquar.east)$) -- ($(actnorm.west)$);
  \draw [black,solid,->] ($(actnorm.east)$) -- ($(output.west)$);

  \draw [black,solid,-] ($(nn.east)$) -- ($(scale-box.west)$);
  \draw [black,solid,-] ($(scale-box.west)$) -- ($(scale-box.east)$);
  \draw [black,solid,-] ($(scale-box.east)$) -- ($(shift-box.west)$);
  \draw [black,solid,-] ($(scale-box.west)$) -- ($(h-box.east)$);

  \draw [black,solid,->] ($(scale-box.west)$) -- ($(affine.south) - (0.5,0)$) node[midway,left] {scale};
  \draw [black,solid,->] ($(shift-box.west)$) -- ($(affine.south) + (0.5,0)$) node[midway,right] {shift};
  \draw [black,solid,->] ($(h-box.east)$) -- ($(cquar.south)$) node[midway,right] {h};

  \draw[blue,thick,dotted] ($(cquar.north west)+(-0.3,0.3)$) rectangle ($(actnorm.south east)+(0.3,-0.3)$);

  \node[text=blue] at ($(actnorm.south east)+(0.6,-0.1)$) {x2};

\end{tikzpicture}
\caption{In this figure, we show our conditional flow model. A neural network takes $X$ and outputs the scale and shift of an affine flow and a condition vector for Residual Flows.  After the first conditional affine flow, there are two Residual Flows with unconditional affine flows placed in between these two and at the end. The condition used in the Residual Flow is the same for both Residual Flows. The Conditional Flow model is equivalent to a Conditional Gaussian if we removed the Residual Flows.}\label{fig:conditional_flow}
\end{figure*}
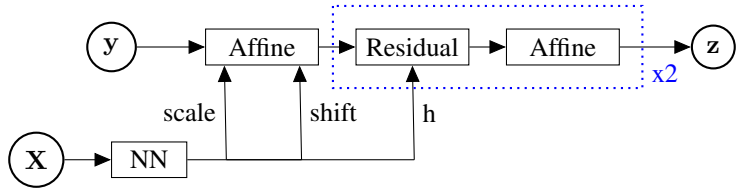

\myref{Figure}{fig:conditional_flow} shows a representation of the Conditional Flow network we used. The Residual Flows are conditioned on a 64-dimensional representation from the condition network. Each Conditional Residual Flow has two hidden layers in the residual connection with a hidden size 64 with ELU \citep{ELU2015} as our activation.
The condition network has a single hidden layer with size 64 and ReLU as our activation. The Residual Flows are conditioned on the 64-dimensional representation from the condition network. For training, we directly used the gradient for the change of variables instead of using the estimators introduced in \citet{Chen2019ResidualFlows} since for one-dimension, the term for the change of variables can be computed in closed form efficiently. For more background on normalizing flows, read \citet{papamakarios2021normalizing}.

For the other neural network approaches, we used networks with two hidden layers with size 128 and used ReLU as our nonlinearity. 
For hyperparameter tuning on all our neural networks, we searched the following grid: 

\begin{itemize}
\item Learning rate: 1e-2, 1e-3, 1e-4
\item Weight decay: 0, 1e-3, 1e-2, 1e-1
\item Dropout Rates: 0.1 0.25, 0.5 0.75
\end{itemize}

The hyperparameters we used were taken from \cite{SQR}. We trained each of our neural networks for 2000 epochs with batch size 128.

For SQR and MC Dropout, we computed the estimated mean and CDF from 1000 samples.
For Conditional Flows, we computed the estimated mean by using the Gauss-Legendre quadrature with 100 points to approximate the integral.

\section{On Model Uncertainty Hypothesis}

A key component of the evaluation methodology of this paper is that it is hypothesized that the true model (the model from which the data we observe is generated) is in our hypothesis space.
However, sometimes in modeling, we do not believe that the true model is in our hypothesis space. In this scenario, the null hypothesis in \myref{Section}{sec:ppc} can be viewed more from the standpoint that evaluation of a model \textit{should} assume that your model contains the true model. However, posterior predictive checks will be limited by statistical error (finite test set size) and so an incorrect model could pass all checks. So implicitly, though the null hypothesis is assumed by the evaluator, the hypothesis can only be tested up to statistical error, and thus one could argue the hypothesis is that the true model is in our hypothesis space \textit{up to statistical error}.

\section{Concrete Examples of Posterior Predictive Checks}\label{sec:ppc_example}

\begin{algorithm}[t]
   \caption{A concrete example of \myref{Algorithm}{alg:ppc} using ECE as the test statistic on a Bayesian ensemble of $M$ models, creating $K$ samples from the posterior predictive distribution. }\label{alg:example_ppc1}
\begin{algorithmic}
   \STATE {\bfseries Input:} data $X=\{x_i\}_{i=1}^{N}$, Bayesian Ensemble $\Theta=[\theta_1,\dots,\theta_M ]$, function $ECE$ that computes the ECE given inputs, labels and a model
   \STATE {for $j \leftarrow 1 $ to $K$}
   \bindent
   \STATE Randomly choose $\theta_m$ from $[\theta_1,\dots,\theta_M]$
   \STATE {for $x_i \in X$}
   \bbindent
   \STATE Sample $y_i$ from $p(y|x_i, \theta_m)$ (create a fake label for $x_i$)
   \eeindent
   \STATE $t_j = \text{ECE}(\{x_i\}_{i=1}^{N}, \{y_i\}_{i=1}^{N}, [\theta_1,\dots,\theta_M])$
   \eindent
   \STATE {\bfseries Output:} $\{t_j\}_{j=1}^{K}$
\end{algorithmic}
\end{algorithm}

\begin{algorithm}[t]
   \caption{A concrete example of computing metrics introduced in \myref{Section}{sec:metrics} for ECE as the test statistic on a Bayesian ensemble of $M$ models, creating $K$ samples from the posterior predictive distribution. }\label{alg:example_metrics}
\begin{algorithmic}
   \STATE {\bfseries Input:} data $X=\{x_i\}_{i=1}^{N}$, labels  $Y=\{y_i\}_{i=1}^{N}$, Bayesian Ensemble $\Theta=[\theta_1,\dots,\theta_M ]$, function $ECE$ that computes the ECE given inputs, labels and a model
   \STATE Get samples $\hat{T} = \{\hat{t}_j\}_{j=1}^{K}$ for test statistic (\myref{Algorithm}{alg:example_ppc1})
   \STATE Compute observed ECE $t = \text{ECE}(\{x_i\}_{i=1}^{N}, \{y_i\}_{i=1}^{N}, [\theta_1,\dots,\theta_M])$
   \STATE p-value $p = \frac{\abs{\hat{t}_i | \hat{t}_i < t,i=1,\dots,K}}{K}$
   \STATE sharpness $s = \text{quantile}(T, 95) - \text{quantile}(T, 5)$
   \STATE {\bfseries Output:} p-value $p$ and sharpness $s$
\end{algorithmic}
\end{algorithm}

\begin{algorithm}[t]
   \caption{A concrete example of \myref{Algorithm}{alg:ppc} using ECE as the test statistic on a conditionally independent ensemble of $M$ models, creating $K$ samples from the posterior predictive distribution. }\label{alg:example_ppc2}
\begin{algorithmic}
   \STATE {\bfseries Input:} data $X=\{x_i\}_{i=1}^{N}$, Conditionally Independent Ensemble $\Theta=[\theta_1,\dots,\theta_M ]$, function $ECE$ that computes the ECE given inputs, labels and a model
   \STATE {for $j \leftarrow 1 $ to $K$}
   \bindent
   \STATE {for $x_i \in X$}
   \bbindent
   \STATE Randomly choose $\theta_m$ from $[\theta_1,\dots,\theta_M]$
   \STATE Sample $y_i$ from $p(y|x_i, \theta_m)$ (create a fake label for $x_i$)
   \eeindent
   \STATE $t_j = \text{ECE}(\{x_i\}_{i=1}^{N}, \{y_i\}_{i=1}^{N}, [\theta_1,\dots,\theta_M])$
   \eindent
   \STATE {\bfseries Output:} $\{t_j\}_{j=1}^{K}$
\end{algorithmic}
\end{algorithm}

In \myref{Section}{sec:ppc}, we introduced a high-level explanation of Posterior Predictive Checks (PPCs) and gave abstract pseudocode in \myref{Algorithm}{alg:ppc}. In this section, we give a few concrete examples of how \myref{Algorithm}{alg:ppc} would be implemented to get samples from the posterior predictive distribution. 

\myref{Algorithm}{alg:example_ppc1} gives example pseudocode of how a test statistic distribution would be created for ECE given an ensemble of classification models. In words, in order to create fake data from the model, a random model $\theta_m$ is first picked from the ensemble $\Theta$; using  $\theta_m$, fake labels $y_i$ are created for all the data points $x_i$. Importantly, a single model $\theta_m$ is used.

In the case of conditionally independent ensembles (\myref{Algorithm}{alg:example_ppc2}), the sampling for $\theta_m$ is moved inside the loop iterating over $\{x_i\}_{i=1}^{N}$, or in the other words, a different model is sampled per $x_i$ (unlike Bayesian ensembles where a single model is used for all the data points).

Further, once given a test statistic distribution, \myref{Algorithm}{alg:example_metrics} shows how the metrics introduced in \myref{Section}{sec:metrics} can be computed.

\section{Uniformity of the CDF of a Single Observation}\label{sec:uniformity_single}

In this section, we show that, for the generative process shown in \myref{Section}{sec:eval_cdfs} for a single observation, the test statistic of the CDF $F(y^{ref} | x)$ using \myref{Equation}{eqn:model_unc_cdf} is uniformly distributed.

Specifically, the generative process is:
\begin{enumerate}
    \item Sample $\theta$ from $p(\theta)$
    \item Sample $y^{ref}$ from $p(y | x, \theta)$
\end{enumerate}
Notationally, we will use $F(y | x, \theta) $ to represent the CDF of $p(y | x, \theta)$.

Using the generative process, we have:
\begin{equation}
p(y | x) = \int d\theta\ p(\theta)\ p(y | x, \theta)
\end{equation}
From this equation for the PDF, the CDF of the generative process is:
\begin{equation}\label{eqn:cdf_thingy}
F(y | x) = \int d\theta\ p(\theta)\ F(y | x, \theta)
\end{equation}

Further, for a CDF $F$, we can show
$$ \mathbb{P}(F(Y) < y) = \mathbb{P}(Y < F^{-1}(y)) = y $$
where the last equation comes from the definition of the CDF (i.e. quantile function).

Using that \myref{Equation}{eqn:cdf_thingy} is equivalent to \myref{Equation}{eqn:model_unc_cdf} and that CDFs are uniformly distributed, we can see that the test statistic $F(y^{ref} | x)$ is uniformly distributed.

\section{Connection between Bayesian Models and Example in \myref{Section}{sec:eval_cdfs}}\label{sec:bayesian_example}

The argument made in \myref{Section}{sec:eval_cdfs} can be similarly derived using a Bayesian approach to fitting a normal distribution with known variance \citep[Section 2.5]{gelman2013bayesian}. More specifically, say we have a known variance $\sigma^2$, a posterior over the mean $\theta \sim \text{Normal}(\mu, \tau^2)$ (where we derive the posterior given some data $y$), and a likelihood model $y \sim \text{Normal}(\theta, \sigma^2)$. In this scenario, the posterior predictive distribution of an observation $\hat{y}$ (this is the distribution being evaluated with calibration error) is:
\begin{align}
  p(\hat{y} | y) &= \int p(\hat{y} | \theta)\ p(\theta | y) d\theta
\\ &\propto \int \exp{\left( -\frac{1}{2\sigma^2}(\hat{y} - \theta)^2 \right)} \exp{\left( -\frac{1}{2\tau^2}(\theta - \mu)^2 \right)} d\theta
\end{align}

Using $\mathbb{E}\left[\hat{y} | \theta\right] = \theta$ and $\text{var}\left[\hat{y} | \theta\right] = \sigma^2$, we find:
\begin{equation}
   \mathbb{E}\left[ \hat{y} | y \right] = \mathbb{E}\left[ \mathbb{E}\left[ \hat{y} |  y, \theta \right] | y \right] =  \mathbb{E}\left[ \theta | y \right] = \mu
\end{equation}
Or, in other words, the mean of the posterior predictive distribution is the mean of the posterior around $\theta$. 

Further, 
\begin{align}
   \text{var}\left[ \hat{y} | y \right] &= \mathbb{E}\left[ \text{var}\left[ \hat{y} | y, \theta \right] | y \right]  + \text{var}\left[ \mathbb{E}\left[ \hat{y} | y, \theta \right] | y \right] 
\\ & = \mathbb{E}\left[ \sigma^2 | y \right]  + \text{var}\left[ \theta | y \right] 
\\ & = \sigma^2 + \tau^2
\end{align}
Or, in other words, the variance of the posterior predictive distribution is the sum of the true (known) variance of the data ($\sigma^2$) and the variance due to the posterior uncertainty around $\theta$ ($\tau^2$). Specifically, the variance of the posterior predictive distribution is always greater than or equal to the known variance (i.e. the variance of the data). \textit{This difference between the true known variance and the variance of the posterior predictive distribution is the crux of the argument of why calibration error is wrong to use given model uncertainty.} Even when our model assumes that the true variance is some value $\sigma^2$, the posterior predictive distribution will not match this assumption due to uncertainty in the posterior (even if the true mean of the data is the $\mu$). This mismatch between the posterior predictive distribution and the true data distribution will result in non-zero calibration error.

Given infinite data, the posterior in this example ($p(\theta)$) would converge to a single solution (a delta distribution); in this scenario, the variance of the posterior predictive distribution would equal the known variance and the calibration error would be zero. This convergence to a single solution might not occur in non-identifiable distributions.

\section{Training and Evaluation Details for Simulated Data}\label{sec:appendix_sim}

For simulated data, we trained networks with three hidden layers, each with a hidden size of 128 and ELU \citep{ELU2015} as the non-linearity. For MC Dropout, we set the dropout rate to 0.1. The networks had two outputs to parameterize a Gaussian. We trained on 10000 data points with a batch size of 128 for 500 epochs, using Adam \citep{Adam2015} with a learning rate 5e-3 and halving the learning rate every 100 epochs.

To evaluate the CDF given the MC Dropout model, we sampled 128 different masks. When sampling a model for the posterior predictive distribution, we ensured that the same mask was applied for all the rows of data. We then applied \myref{Algorithm}{alg:ppc} to sample 1000 values from the posterior predictive distribution.

For the i.i.d. test data, we sampled 1000 data points using the same generative process as used for training; for the out of distribution data, we sampled 10000 data points in the region removed during training.

All models were trained and evaluated on CPUs.

\section{Evaluation Details for CIFAR-10}\label{sec:appendix_img_exp}

In this section, we give more details on the methodology for doing posterior predictive checks on CIFAR-10. All evaluation was done on CPUs.

\subsection{Recalibration}\label{sec:appendix_recalib}

Using the technique from \citet{Ashukha2020Pitfalls}, we recalibrated the prediction, averaged over the model uncertainty, using temperature scaling. However, to increase capacity, we used a different temperature per model. In the case of ensembles, we had 50 trained parameters; in the case of MC Dropout, we sampled 50 predictions and accordingly also had 50 parameters. The loss function we optimized was:
$$\arg\max_{\{\tau_j\}_{j=1}^{50}} \sum_{x_i \in X} \frac{1}{50} \log \sum_{j=1}^{50} \frac{p(y_i | x_i) / \tau_{j}}{\sum_{y_k \in Y} p(y_k | x_i) / \tau_{j}} $$

\begin{figure}[!tb]
\begin{subfigure}{0.45\linewidth}
\centering
\centerline{\includegraphics[width=\columnwidth]{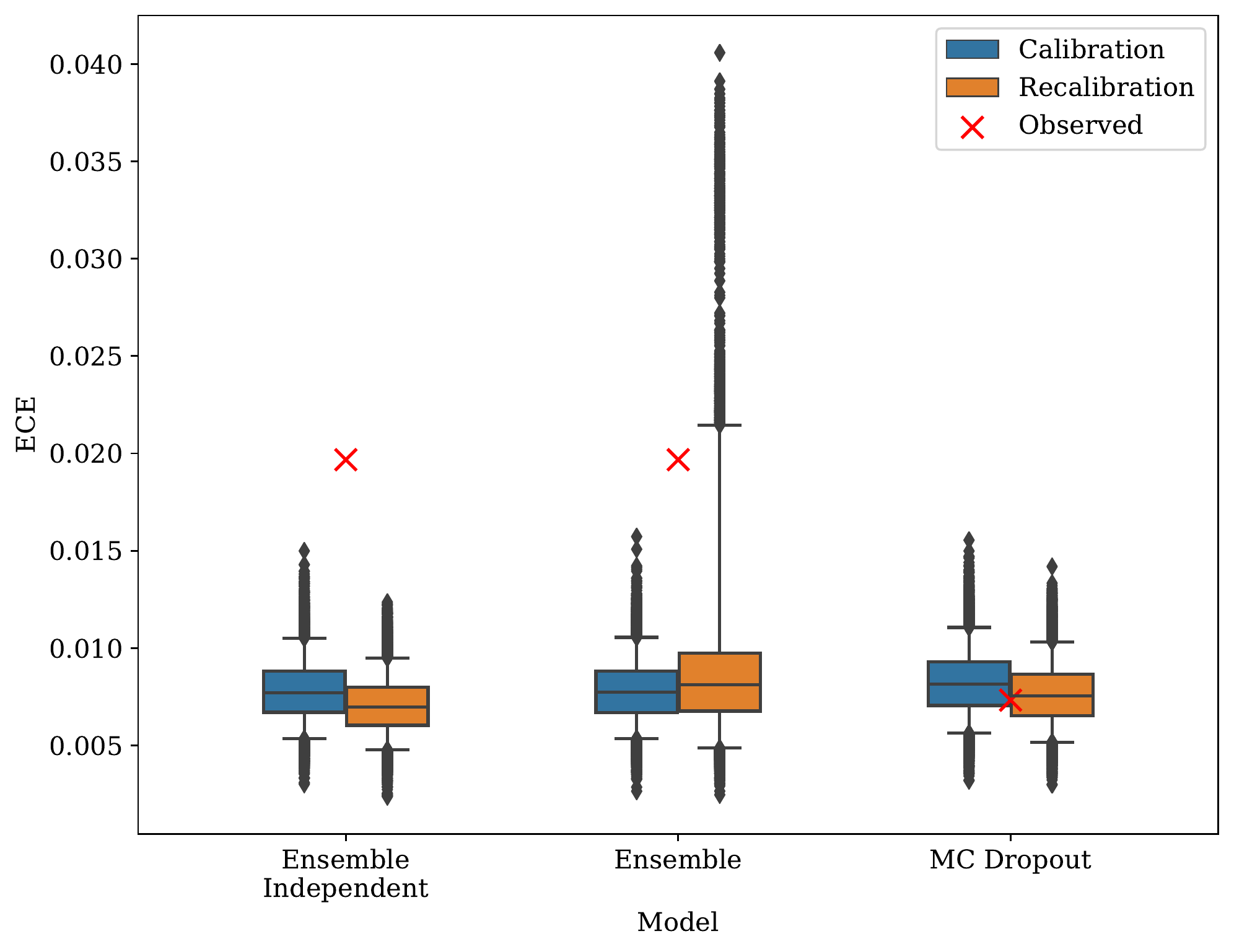}}
\caption{Posterior Check of ECE}
\label{fig:one_hidden_layer}
\end{subfigure} \hfill
\begin{subfigure}{0.45\linewidth}
\centering
\centerline{\includegraphics[width=\columnwidth]{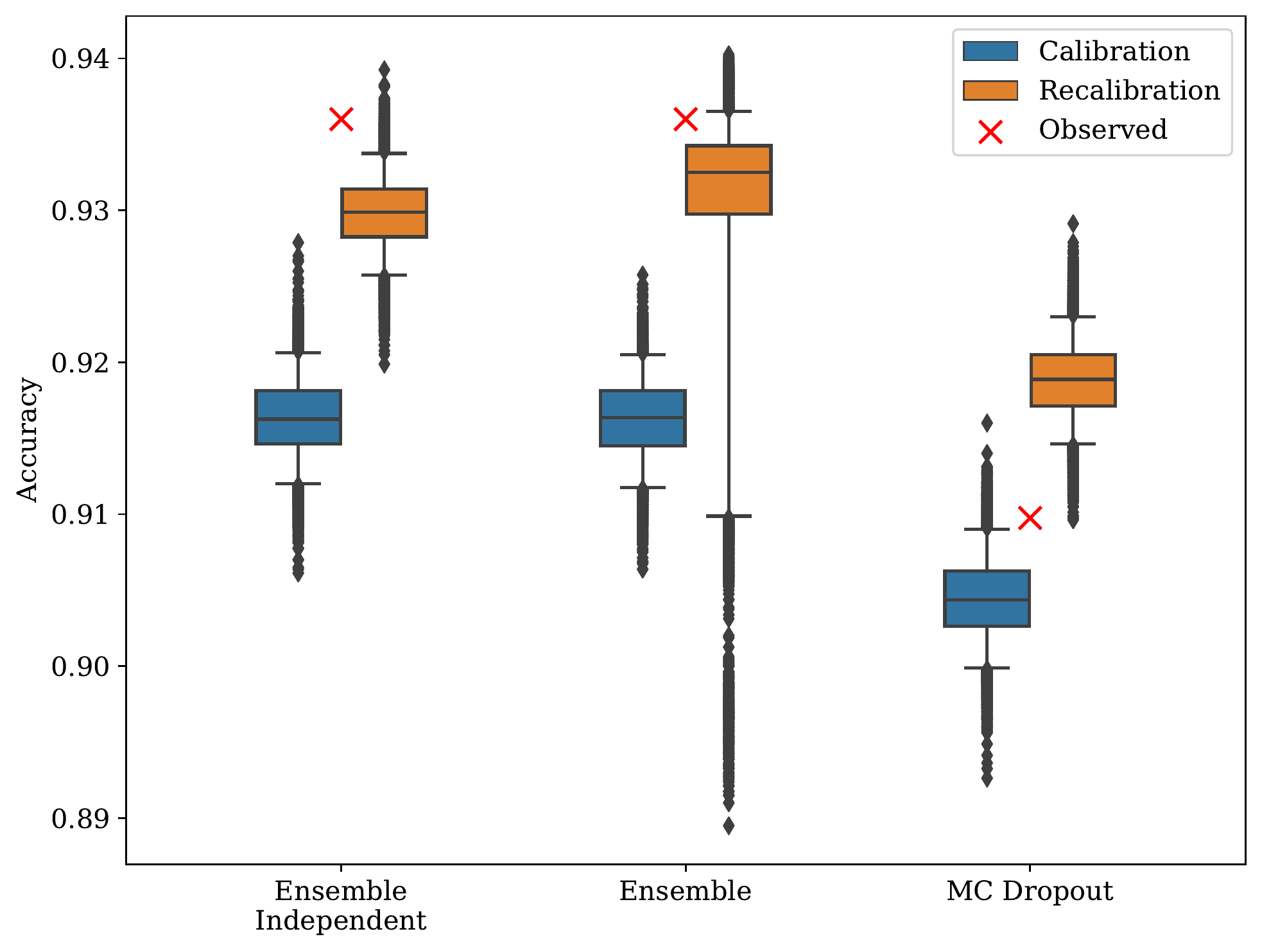}}
\caption{Posterior Check of Accuracy}
\label{fig:one_hidden_layer_felu}
\end{subfigure}
\caption{Posterior predictive checks of models trained and evaluated on CIFAR-10, with and without recalibration.}
\label{fig:cifar_recalib}
\end{figure}

In \myref{Figure}{fig:cifar_recalib}, we compare the performance on the uncorrupted test set with and without recalibration. We see that recalibration is a crucial component for the posterior predictive checks to pass for Bayesian Ensembles and that even with recalibration, conditionally independent ensembles fail the ECE check. An interesting note is that the temperatures learned increased the confidence of few of the models. 

To ensure that we do not evaluate on data that any part of the model has been trained on, we use the (first) 20\% of the test data to recalibrate and evaluate on the remaining 80\%. Further, when evaluating on CIFAR-10-C, we did not evaluate on the corrupted versions of the 20\% we recalibrated on. 

\subsection{Ensembles}

For ensembles, we precomputed the logits predicted on each image per ensemble member. From here, we compute ECE and Accuracy given the labels. This operation is no different than what is currently done to evaluate ensembles. To compute the posterior predictive distribution, we followed in the algorithm in \myref{Algorithm}{alg:example_ppc1}, where sampling from $p(\theta)$ means we randomly sampled a model from the ensemble. For Independent Ensemble, we randomly sampled a different model per image (\myref{Algorithm}{alg:example_ppc2}); we performed this efficiently by precomputing the average probability per image across the models and sampling from this distribution to get the fake label ($y^{ref}$).

\subsection{MC Dropout}

For MC Dropout, we precomputed the logits for 50 samples of masks. From here, the process of doing the PPC is equivalent to Ensembles. Similar to how in \myref{Algorithm}{alg:example_ppc1} we randomly sample a single model from the ensemble, for MC Dropout, we sample \textit{one mask} to apply on all $x_i$.

\section{Additional Plots}\label{sec:addl_plot}

\begin{figure}[!h]
\begin{subfigure}{0.45\linewidth}
\centering
\centerline{\includegraphics[width=\columnwidth]{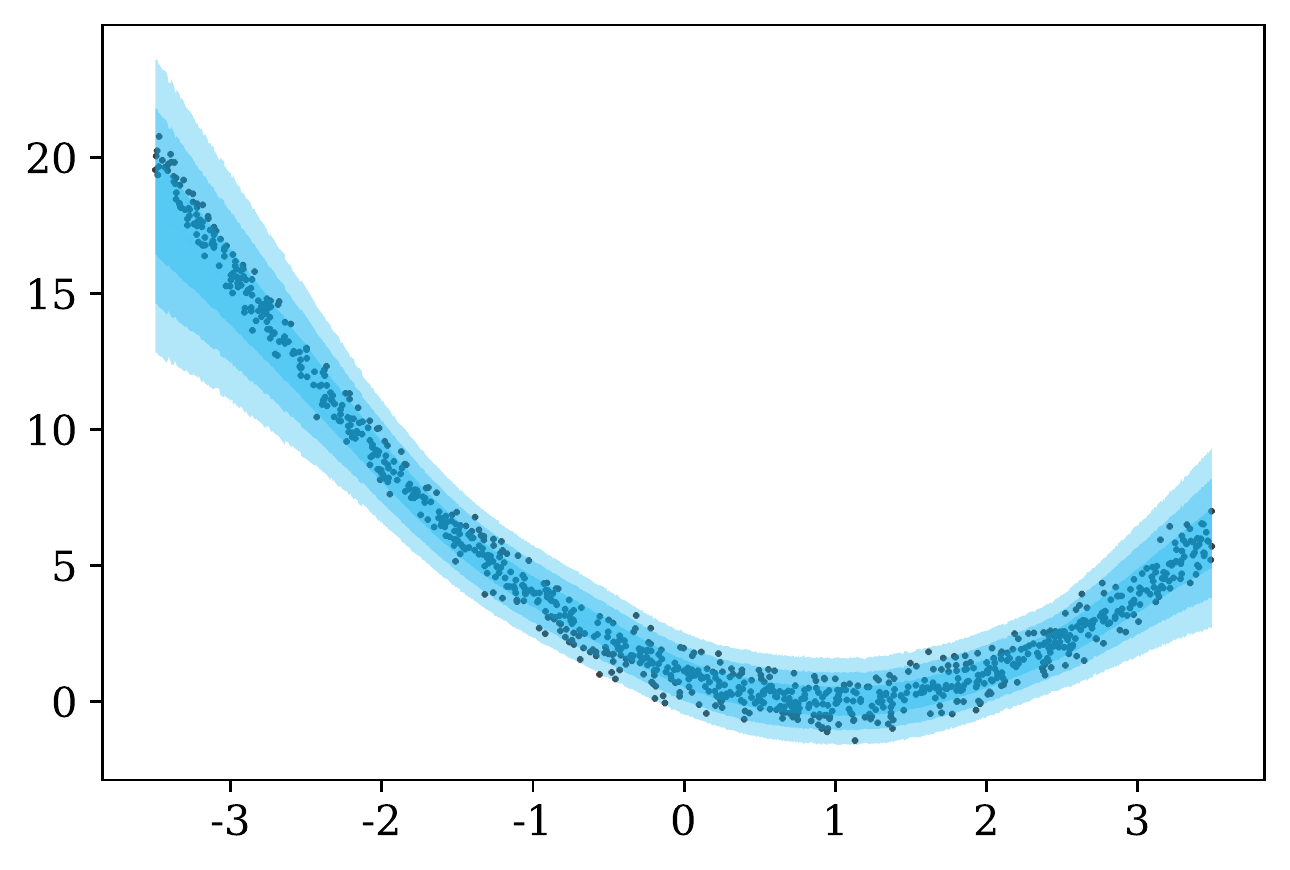}}
\caption{Uncertainty}
\label{fig:var_mc}
\end{subfigure} \hfill
\begin{subfigure}{0.45\linewidth}
\centering
\centerline{\includegraphics[width=\columnwidth]{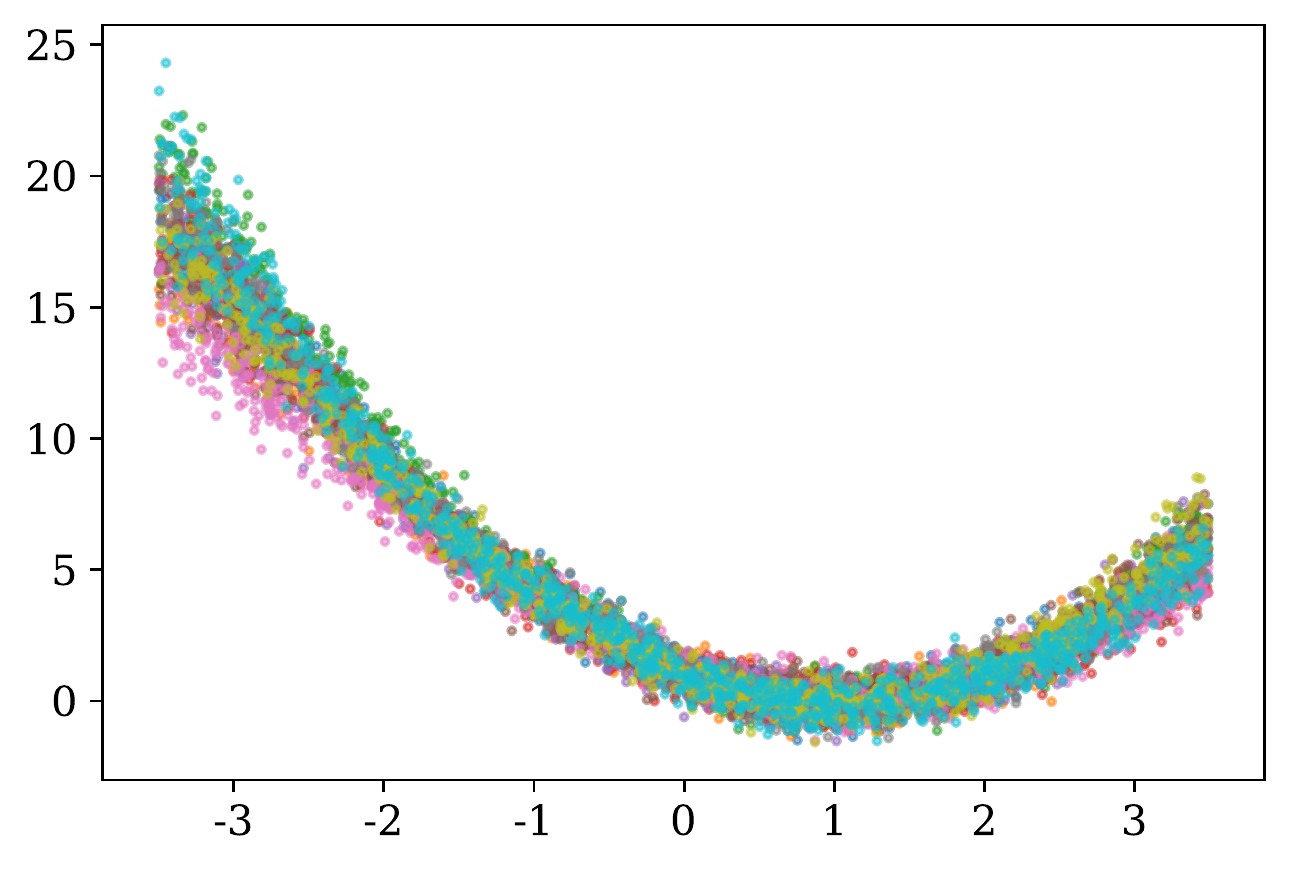}}
\caption{Model Samples}
\label{fig:samples_mc}
\end{subfigure}
\caption{Comparison of a plot showing the sum of aleatoric and epistemic (model) uncertainty versus a plot showing samples drawn from models drawn from the model distribution. \myref{Figure}{fig:var_mc} was generated by using code from \href{https://github.com/aamini/evidential-deep-learning}{https://github.com/aamini/evidential-deep-learning} (Apache License, Version 2.0).}
\end{figure}

Whereas classical evaluation has often created plots similar to \myref{Figure}{fig:var_mc} \citep{amini2020deep}, plots such as \myref{Figure}{fig:samples_mc} show the correlation from the model uncertainty that is obscured by plots such as \myref{Figure}{fig:var_mc}.
\section{Full CIFAR-10-C Results}\label{sec:cifar_10_c_full}

\begin{figure}[!bt]
\begin{subfigure}{0.45\linewidth}
\centering
\centerline{\includegraphics[width=\columnwidth]{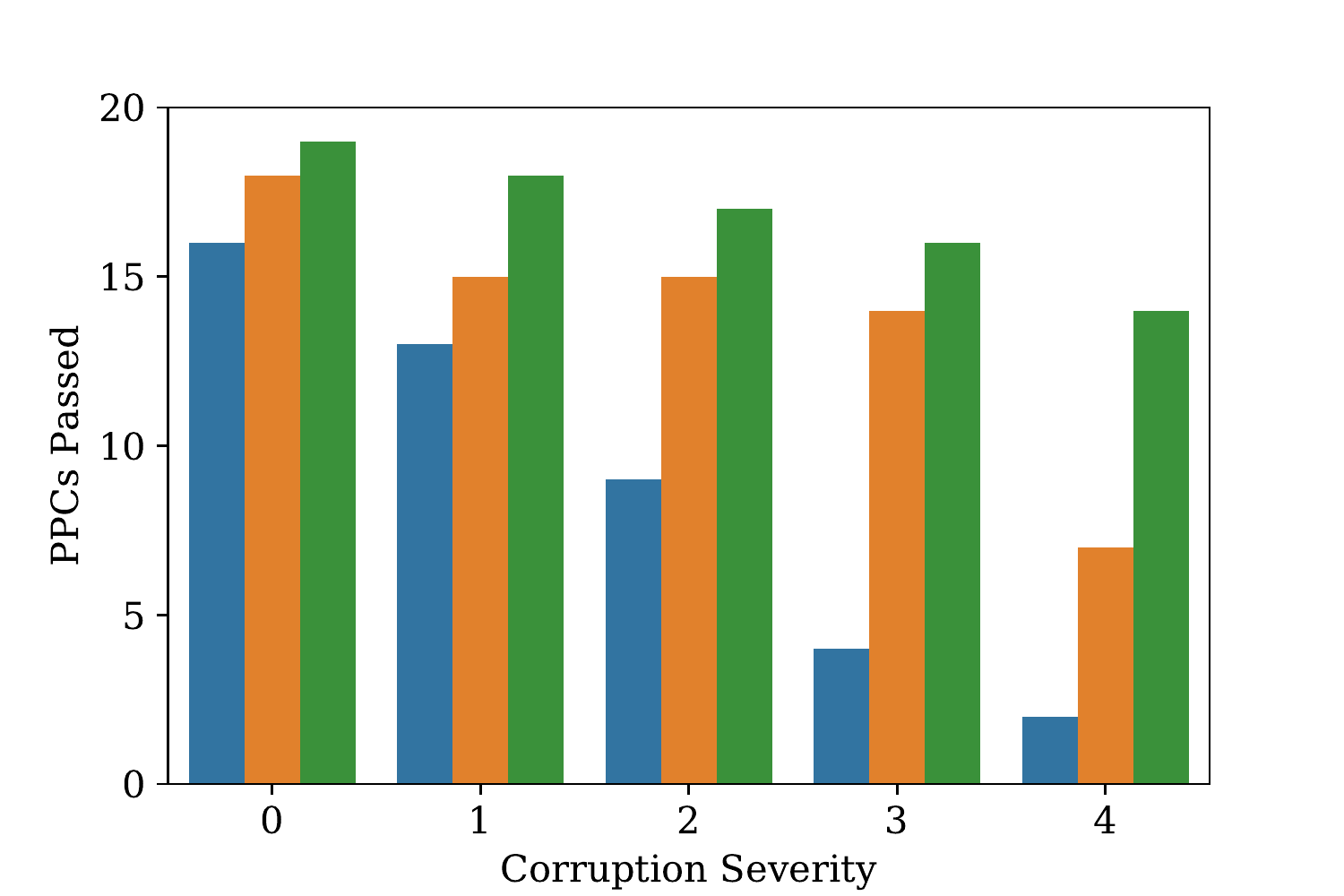}}
\caption{ECE}
\end{subfigure} \hfill
\begin{subfigure}{0.45\linewidth}
\centering
\centerline{\includegraphics[width=\columnwidth]{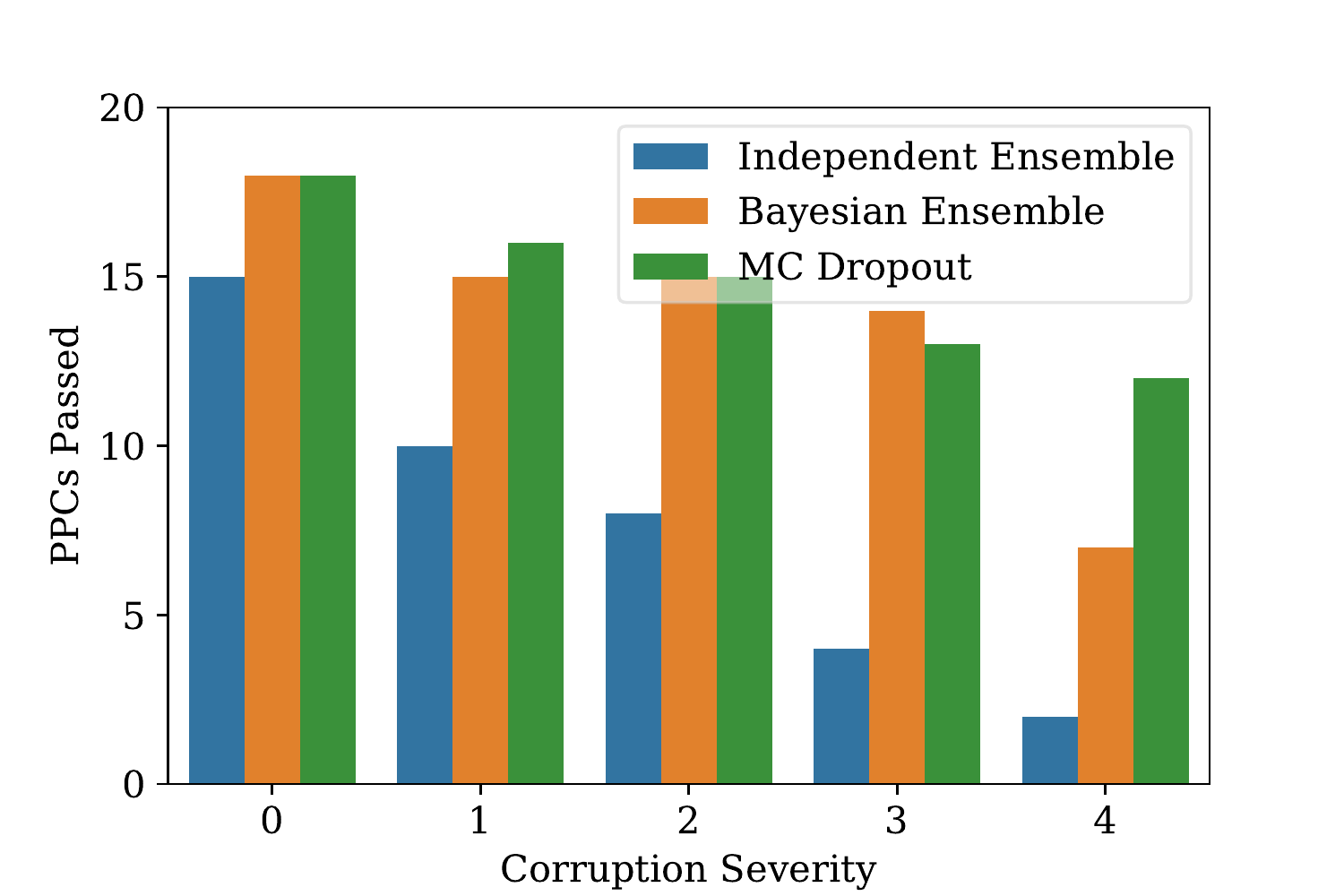}}
\caption{Accuracy}
\end{subfigure}
\caption{For each model, we plot how many PPCs for different corruptions at a specific severity passed. A PPC passes if the observed value falls within the range defined by the samples of the test statistic (p-value is not zero or one).}\label{fig:summary_cifar10c}
\end{figure}

\begin{table*}[!bt]
    \caption{For each model, we give a count of how many PPCs for different corruptions at a specific severity passed. A PPC passes if the observed value falls within the range defined by the samples of the test statistic (p-value is not zero or one)}\label{tbl:summary_cifar10c}

    \begin{subtable}[h]{0.49\textwidth}
    \small
    \centering
    \caption{ECE}
    \begin{tabular}{lccccc}
    \toprule
         & \multicolumn{5}{c}{Corruption Severity}
    \\
     \cmidrule(lr){2-6} 
     & {1} & {2}& {3} & {4}  & {5} 
    \\
    \midrule
    Independent Ensemble & 16&13&9&4&2 \\
    Bayesian Ensemble & 18&15&15&14&7 \\
    MC Dropout & 19&18&17&16&14 \\
    \bottomrule
    \end{tabular}

    \end{subtable} \hfill
    \begin{subtable}[h]{0.49\textwidth}
    \small
    \centering
    \caption{Accuracy}
    \begin{tabular}{lccccc}
    \toprule
         & \multicolumn{5}{c}{Corruption Severity}
    \\
     \cmidrule(lr){2-6} 
     & {1} & {2}& {3} & {4}  & {5} 
    \\
    \midrule
    Independent Ensemble & 15&10&8&4&2 \\ 
    Bayesian Ensemble & 18&15&15&14&7 \\
    MC Dropout & 18&16&15&13&12 \\
    \bottomrule
    \end{tabular}
    \end{subtable} 
\end{table*}

In this section, not only do we show the box plots of the PPCs on all the corruptions in CIFAR-10-C, we also give the p-values of the observed value and the sharpness of the posterior predictive distribution. Importantly, we define the p-value here as the fraction of samples of the test statistics less than the observed value, for both ECE and accuracy.

Though the example we showed in the main paper had the models perform reasonably well on all the severities, there are corruptions such as shot noise and gaussian noise where the models are unable to account for the distribution shift.

We summarize the performance of the three models across the different corruption types and severities in \myref{Figure}{fig:summary_cifar10c} (\myref{Table}{tbl:summary_cifar10c}) where we give the number of posterior predictive checks passed, where passing means the observed value falls within the samples of the test statistic (empirical p-value is not zero or one). We can see that the Bayesian Ensemble interpretation of ensembling leads to more tests passed and that MC Dropout seems to be able to account for the distribution shift.

\input{large_table}

%% file: large_table.tex
\begin{figure}
\begin{minipage}{\textwidth}
\begin{subtable}[h]{0.45\textwidth}
\small
\begin{tabular}{lrrrrr}
\toprule
corruption &    0 &    1 &    2 &    3 &    4 \\
model       &      &      &      &      &      \\
\midrule
Dropout     & 0.11 & 0.02 & 0.00 & 0.00 & 0.00 \\
Ensemble    & 0.20 & 0.00 & 0.00 & 0.00 & 0.00 \\
Independent & 0.00 & 0.00 & 0.00 & 0.00 & 0.00 \\
\bottomrule
\end{tabular}

\caption{p-value for Accuracy}
\end{subtable}\hfill
\begin{subtable}[h]{0.45\textwidth}
\small
\begin{tabular}{lrrrrr}
\toprule
corruption &    0 &    1 &    2 &    3 &    4 \\
model       &      &      &      &      &      \\
\midrule
Dropout     & 0.40 & 0.97 & 1.00 & 1.00 & 1.00 \\
Ensemble    & 0.60 & 1.00 & 1.00 & 1.00 & 1.00 \\
Independent & 0.96 & 1.00 & 1.00 & 1.00 & 1.00 \\
\bottomrule
\end{tabular}

\caption{p-value for ECE}
\end{subtable}

\begin{subtable}[h]{\textwidth}
\small
\centering
\begin{tabular}{lrrrrr}
\toprule
corruption &        0 &        1 &        2 &        3 &        4 \\
model       &          &          &          &          &          \\
\midrule
Dropout     & $2.38\times 10^{-2}$ & $3.05\times 10^{-2}$ & $3.46\times 10^{-2}$ & $2.77\times 10^{-2}$ & $2.50\times 10^{-2}$ \\
Ensemble    & $1.24\times 10^{-2}$ & $1.99\times 10^{-2}$ & $4.02\times 10^{-2}$ & $4.93\times 10^{-2}$ & $7.10\times 10^{-2}$ \\
Independent & $4.75\times 10^{-3}$ & $5.37\times 10^{-3}$ & $6.00\times 10^{-3}$ & $6.00\times 10^{-3}$ & $6.25\times 10^{-3}$ \\
\bottomrule
\end{tabular}

\caption{Sharpness for Accuracy}
\end{subtable}

\begin{subtable}[h]{\linewidth}
\small
\centering
\begin{tabular}{lrrrrr}
\toprule
corruption &        0 &        1 &        2 &        3 &        4 \\
model       &          &          &          &          &          \\
\midrule
Dropout     & $1.20\times 10^{-2}$ & $1.64\times 10^{-2}$ & $1.82\times 10^{-2}$ & $1.70\times 10^{-2}$ & $1.57\times 10^{-2}$ \\
Ensemble    & $5.03\times 10^{-3}$ & $8.64\times 10^{-3}$ & $1.90\times 10^{-2}$ & $2.04\times 10^{-2}$ & $3.71\times 10^{-2}$ \\
Independent & $2.74\times 10^{-3}$ & $3.22\times 10^{-3}$ & $3.63\times 10^{-3}$ & $3.69\times 10^{-3}$ & $3.76\times 10^{-3}$ \\
\bottomrule
\end{tabular}

\caption{Sharpness for ECE}
\end{subtable}

\begin{subfigure}{0.45\linewidth}
\centering
\centerline{\includegraphics[width=\columnwidth]{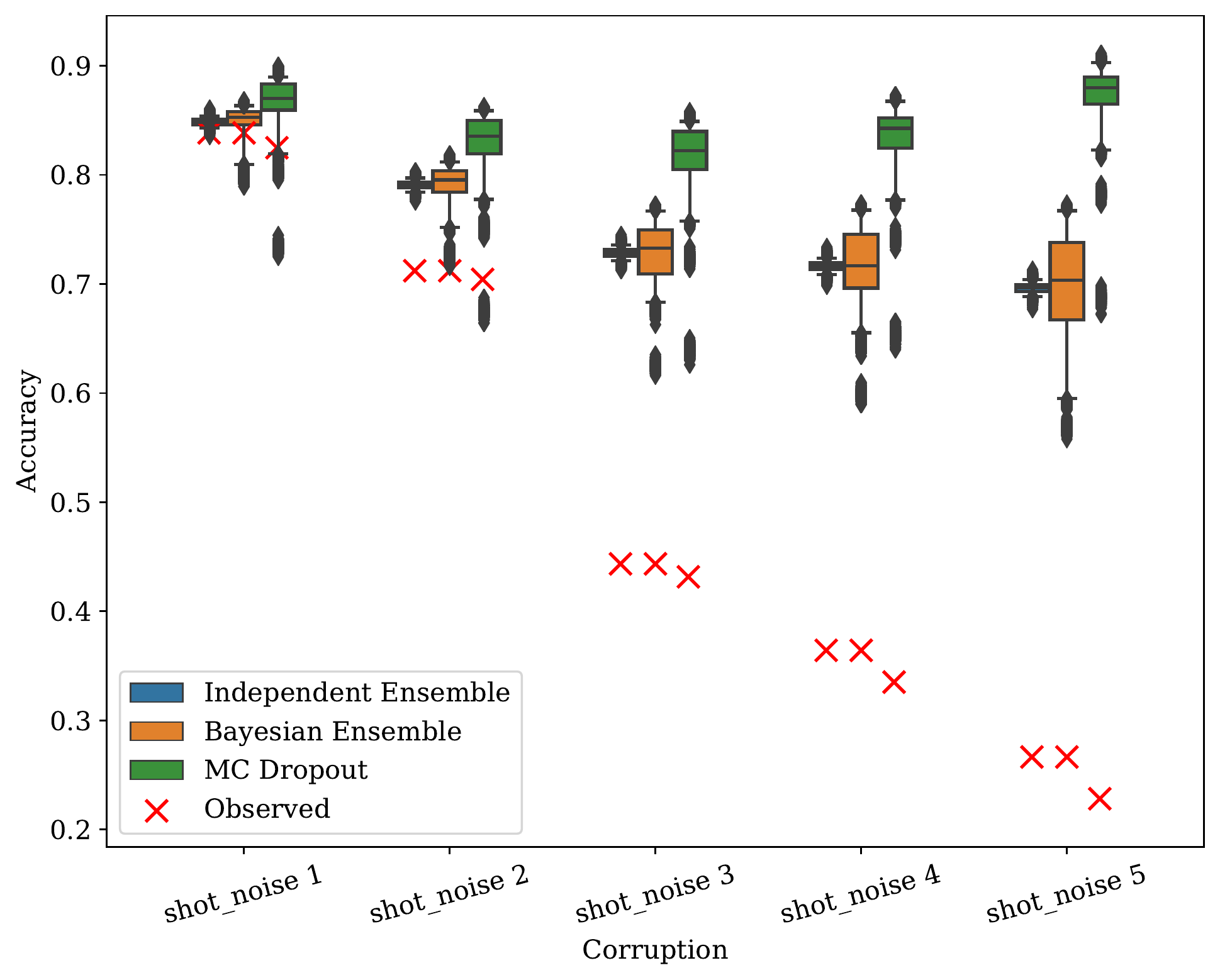}}
\caption{Posterior Check of Accuracy}
\end{subfigure} \hfill
\begin{subfigure}{0.45\linewidth}
\centering
\centerline{\includegraphics[width=\columnwidth]{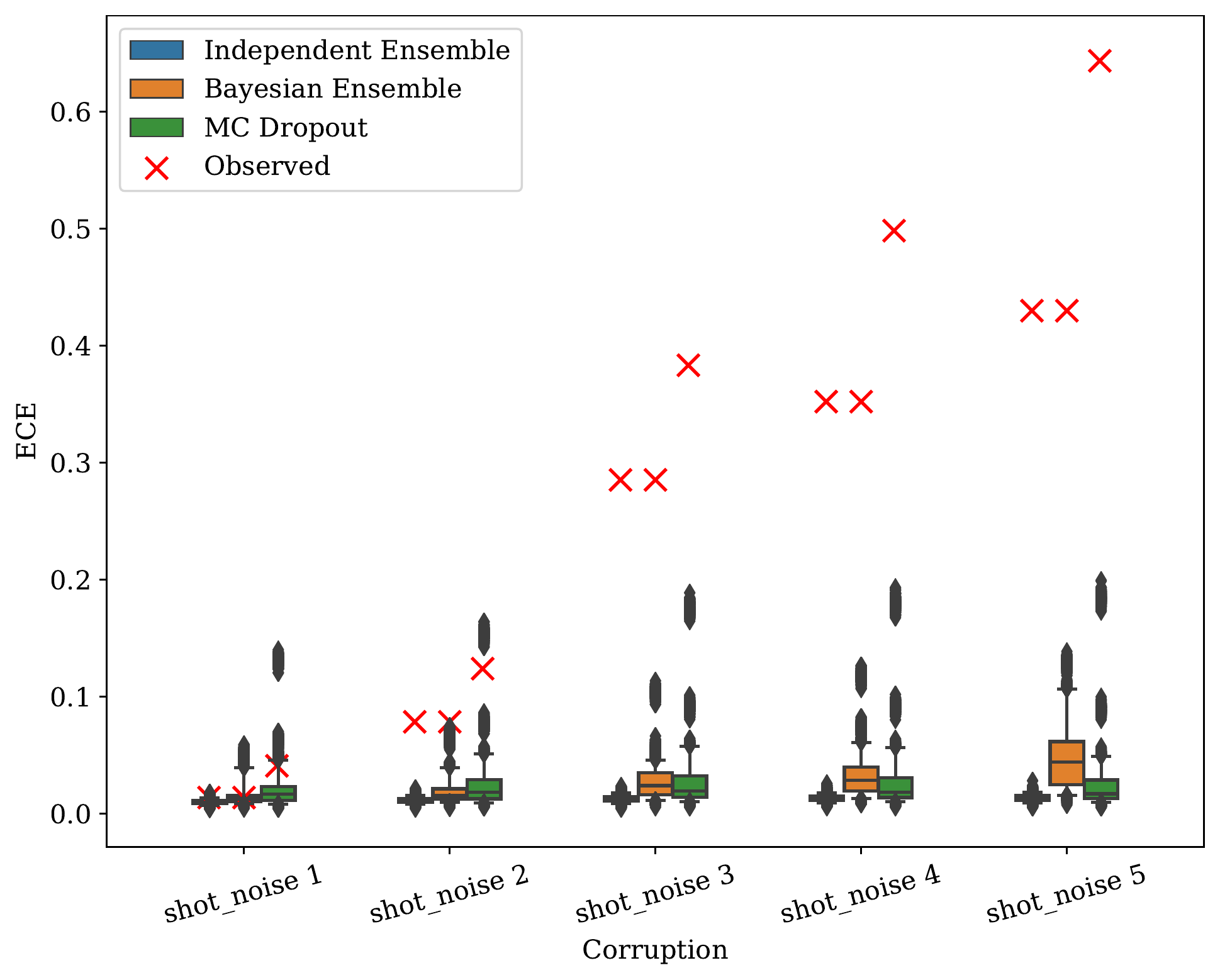}}
\caption{Posterior Check of ECE}
\end{subfigure}
\caption{Posterior predictive checks of recalibrated models trained on CIFAR-10, evaluated on shot noise data from CIFAR-10-C.}\label{fig:shot_noise}
\end{minipage}
\end{figure}
    
\begin{figure}
\begin{minipage}{\textwidth}
\begin{subtable}[h]{0.45\textwidth}
\small
\begin{tabular}{lrrrrr}
\toprule
corruption &    0 &    1 &    2 &    3 &    4 \\
model       &      &      &      &      &      \\
\midrule
Dropout     & 0.12 & 0.02 & 0.02 & 0.00 & 0.00 \\
Ensemble    & 0.14 & 0.04 & 0.04 & 0.02 & 0.00 \\
Independent & 0.00 & 0.00 & 0.00 & 0.00 & 0.00 \\
\bottomrule
\end{tabular}

\caption{p-value for Accuracy}
\end{subtable}\hfill
\begin{subtable}[h]{0.45\textwidth}
\small
\begin{tabular}{lrrrrr}
\toprule
corruption &    0 &    1 &    2 &    3 &    4 \\
model       &      &      &      &      &      \\
\midrule
Dropout     & 0.73 & 0.94 & 0.96 & 0.98 & 0.98 \\
Ensemble    & 0.92 & 0.96 & 0.96 & 0.98 & 1.00 \\
Independent & 1.00 & 1.00 & 1.00 & 1.00 & 1.00 \\
\bottomrule
\end{tabular}

\caption{p-value for ECE}
\end{subtable}
\begin{subtable}[h]{\linewidth}
\small
\centering
\begin{tabular}{lrrrrr}
\toprule
corruption &        0 &        1 &        2 &        3 &        4 \\
model       &          &          &          &          &          \\
\midrule
Dropout     & $2.50\times 10^{-2}$ & $3.53\times 10^{-2}$ & $4.40\times 10^{-2}$ & $4.40\times 10^{-2}$ & $3.09\times 10^{-2}$ \\
Ensemble    & $1.61\times 10^{-2}$ & $1.90\times 10^{-2}$ & $3.02\times 10^{-2}$ & $5.76\times 10^{-2}$ & $6.10\times 10^{-2}$ \\
Independent & $4.75\times 10^{-3}$ & $5.50\times 10^{-3}$ & $6.25\times 10^{-3}$ & $6.88\times 10^{-3}$ & $7.13\times 10^{-3}$ \\
\bottomrule
\end{tabular}

\caption{Sharpness for Accuracy}
\end{subtable}

\begin{subtable}[h]{\linewidth}
\small
\centering
\begin{tabular}{lrrrrr}
\toprule
corruption &        0 &        1 &        2 &        3 &        4 \\
model       &          &          &          &          &          \\
\midrule
Dropout     & $1.43\times 10^{-2}$ & $1.88\times 10^{-2}$ & $2.38\times 10^{-2}$ & $2.67\times 10^{-2}$ & $2.45\times 10^{-2}$ \\
Ensemble    & $5.26\times 10^{-3}$ & $8.85\times 10^{-3}$ & $1.60\times 10^{-2}$ & $2.54\times 10^{-2}$ & $3.21\times 10^{-2}$ \\
Independent & $2.83\times 10^{-3}$ & $3.29\times 10^{-3}$ & $3.67\times 10^{-3}$ & $4.17\times 10^{-3}$ & $4.36\times 10^{-3}$ \\
\bottomrule
\end{tabular}

\caption{Sharpness for ECE}
\end{subtable}

\begin{subfigure}{0.45\linewidth}
\centering
\centerline{\includegraphics[width=\columnwidth]{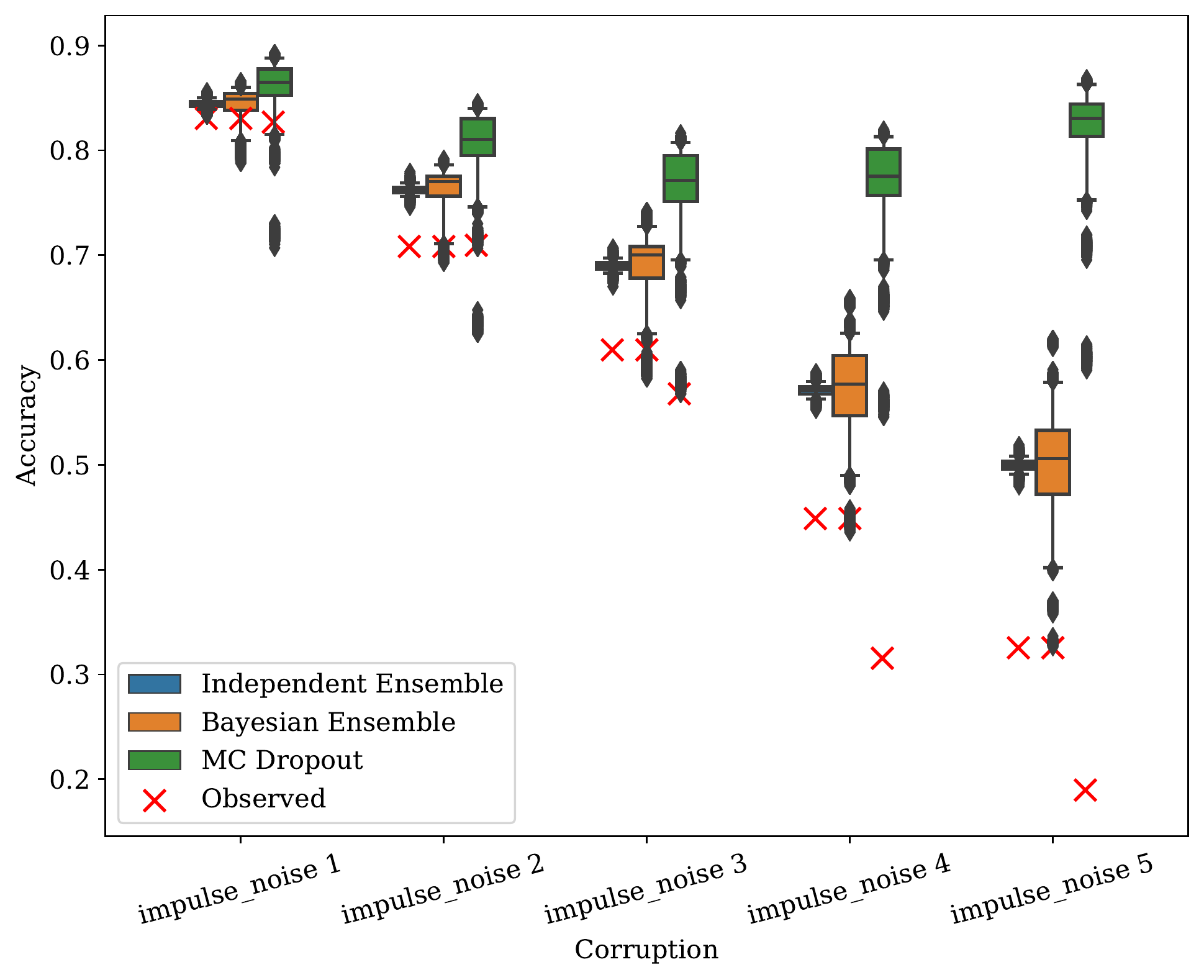}}
\caption{Posterior Check of Accuracy}
\end{subfigure} \hfill
\begin{subfigure}{0.45\linewidth}
\centering
\centerline{\includegraphics[width=\columnwidth]{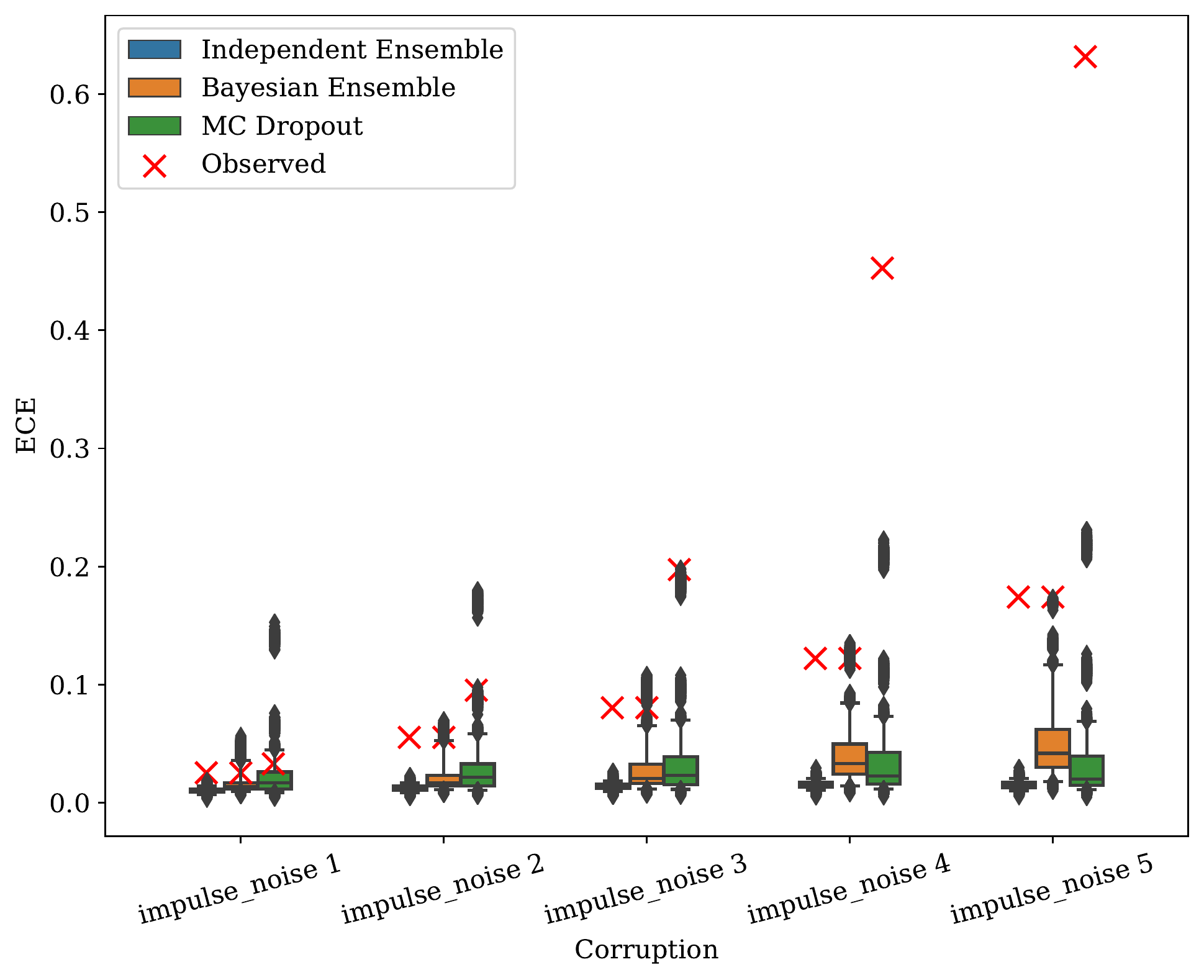}}
\caption{Posterior Check of ECE}
\end{subfigure}
\caption{Posterior predictive checks of recalibrated models trained on CIFAR-10, evaluated on impulse noise data from CIFAR-10-C.}\label{fig:impulse_noise}
\end{minipage}
\end{figure}
    
\begin{figure}
\begin{minipage}{\textwidth}
\begin{subtable}[h]{0.45\textwidth}
\small
\begin{tabular}{lrrrrr}
\toprule
corruption &    0 &    1 &    2 &    3 &    4 \\
model       &      &      &      &      &      \\
\midrule
Dropout     & 0.24 & 0.10 & 0.09 & 0.08 & 0.04 \\
Ensemble    & 0.28 & 0.17 & 0.16 & 0.12 & 0.09 \\
Independent & 0.59 & 0.00 & 0.00 & 0.00 & 0.00 \\
\bottomrule
\end{tabular}

\caption{p-value for Accuracy}
\end{subtable}\hfill
\begin{subtable}[h]{0.45\textwidth}
\small
\begin{tabular}{lrrrrr}
\toprule
corruption &    0 &    1 &    2 &    3 &    4 \\
model       &      &      &      &      &      \\
\midrule
Dropout     & 0.46 & 0.49 & 0.48 & 0.57 & 0.86 \\
Ensemble    & 0.81 & 0.81 & 0.76 & 0.88 & 0.91 \\
Independent & 0.99 & 1.00 & 1.00 & 1.00 & 1.00 \\
\bottomrule
\end{tabular}

\caption{p-value for ECE}
\end{subtable}
\begin{subtable}[h]{\linewidth}
\small
\centering
\begin{tabular}{lrrrrr}
\toprule
corruption &        0 &        1 &        2 &        3 &        4 \\
model       &          &          &          &          &          \\
\midrule
Dropout     & $2.60\times 10^{-2}$ & $3.01\times 10^{-2}$ & $3.10\times 10^{-2}$ & $3.45\times 10^{-2}$ & $3.63\times 10^{-2}$ \\
Ensemble    & $7.75\times 10^{-3}$ & $1.00\times 10^{-2}$ & $1.14\times 10^{-2}$ & $1.44\times 10^{-2}$ & $1.54\times 10^{-2}$ \\
Independent & $4.38\times 10^{-3}$ & $4.75\times 10^{-3}$ & $5.00\times 10^{-3}$ & $5.13\times 10^{-3}$ & $5.50\times 10^{-3}$ \\
\bottomrule
\end{tabular}

\caption{Sharpness for Accuracy}
\end{subtable}

\begin{subtable}[h]{\linewidth}
\small
\centering
\begin{tabular}{lrrrrr}
\toprule
corruption &        0 &        1 &        2 &        3 &        4 \\
model       &          &          &          &          &          \\
\midrule
Dropout     & $1.44\times 10^{-2}$ & $1.40\times 10^{-2}$ & $1.65\times 10^{-2}$ & $1.74\times 10^{-2}$ & $1.78\times 10^{-2}$ \\
Ensemble    & $3.99\times 10^{-3}$ & $5.00\times 10^{-3}$ & $7.13\times 10^{-3}$ & $6.66\times 10^{-3}$ & $5.62\times 10^{-3}$ \\
Independent & $2.65\times 10^{-3}$ & $2.95\times 10^{-3}$ & $2.98\times 10^{-3}$ & $3.17\times 10^{-3}$ & $3.26\times 10^{-3}$ \\
\bottomrule
\end{tabular}

\caption{Sharpness for ECE}
\end{subtable}

\begin{subfigure}{0.45\linewidth}
\centering
\centerline{\includegraphics[width=\columnwidth]{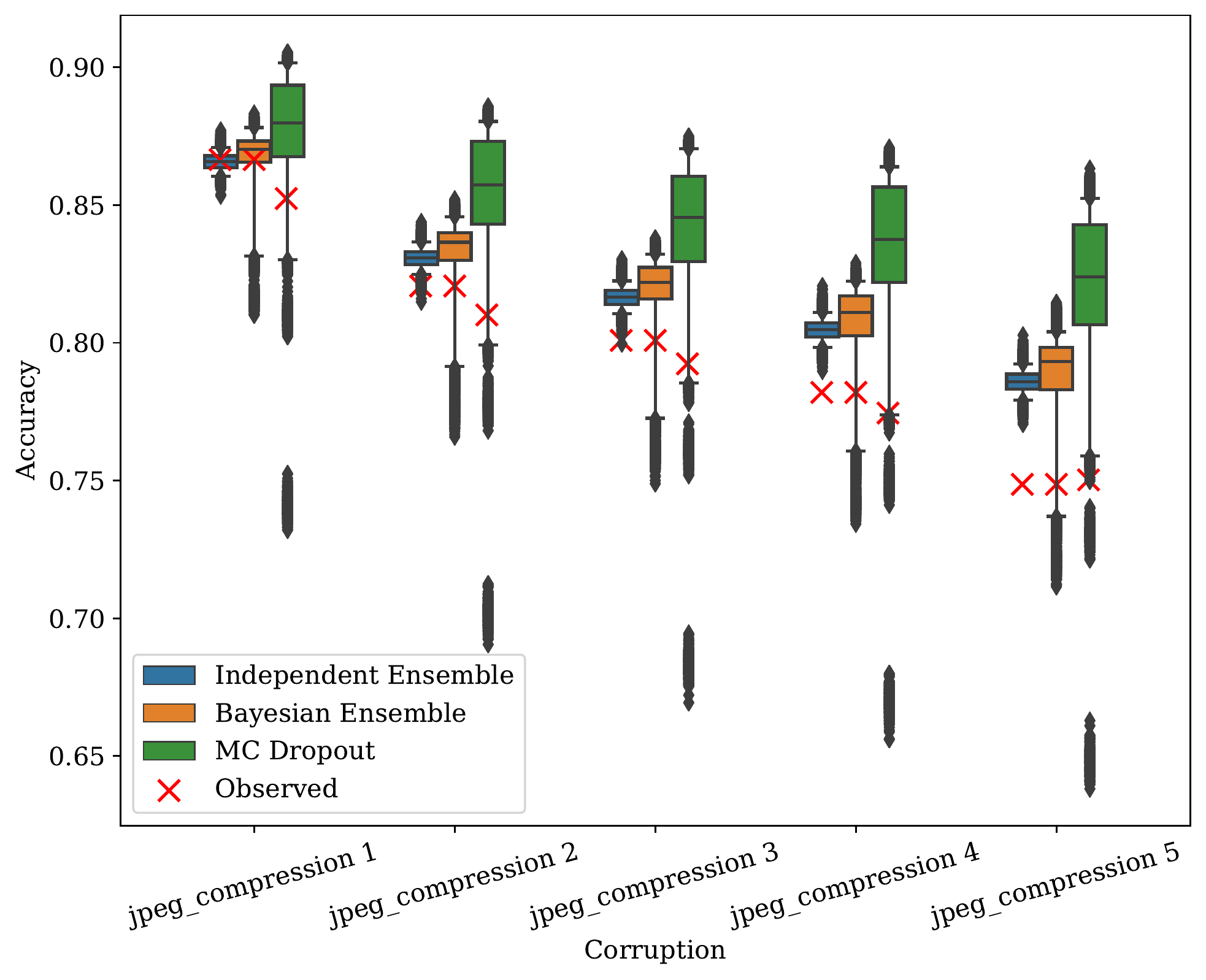}}
\caption{Posterior Check of Accuracy}
\end{subfigure} \hfill
\begin{subfigure}{0.45\linewidth}
\centering
\centerline{\includegraphics[width=\columnwidth]{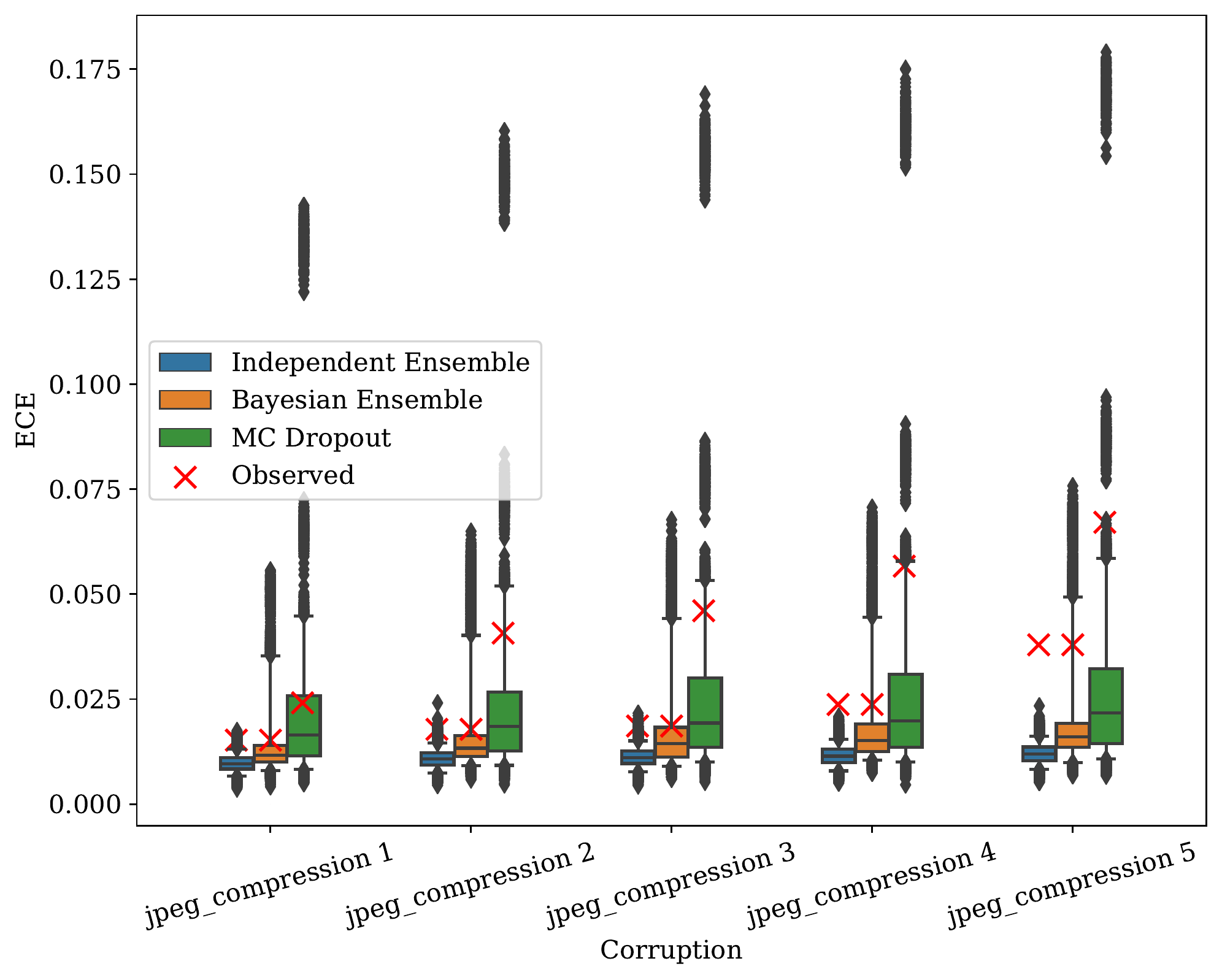}}
\caption{Posterior Check of ECE}
\end{subfigure}
\caption{Posterior predictive checks of recalibrated models trained on CIFAR-10, evaluated on jpeg compression data from CIFAR-10-C.}\label{fig:jpeg_compression}
\end{minipage}
\end{figure}
    
\begin{figure}
\begin{minipage}{\textwidth}
\begin{subtable}[h]{0.45\textwidth}
\small
\begin{tabular}{lrrrrr}
\toprule
corruption &    0 &    1 &    2 &    3 &    4 \\
model       &      &      &      &      &      \\
\midrule
Dropout     & 0.71 & 0.38 & 0.14 & 0.04 & 0.02 \\
Ensemble    & 0.82 & 0.31 & 0.16 & 0.07 & 0.00 \\
Independent & 1.00 & 0.38 & 0.00 & 0.00 & 0.00 \\
\bottomrule
\end{tabular}

\caption{p-value for Accuracy}
\end{subtable}\hfill
\begin{subtable}[h]{0.45\textwidth}
\small
\begin{tabular}{lrrrrr}
\toprule
corruption &    0 &    1 &    2 &    3 &    4 \\
model       &      &      &      &      &      \\
\midrule
Dropout     & 0.44 & 0.18 & 0.41 & 0.90 & 0.96 \\
Ensemble    & 0.71 & 0.29 & 0.55 & 0.92 & 1.00 \\
Independent & 1.00 & 0.80 & 1.00 & 1.00 & 1.00 \\
\bottomrule
\end{tabular}

\caption{p-value for ECE}
\end{subtable}
\begin{subtable}[h]{\linewidth}
\small
\centering
\begin{tabular}{lrrrrr}
\toprule
corruption &        0 &        1 &        2 &        3 &        4 \\
model       &          &          &          &          &          \\
\midrule
Dropout     & $3.19\times 10^{-2}$ & $3.41\times 10^{-2}$ & $3.51\times 10^{-2}$ & $4.17\times 10^{-2}$ & $4.66\times 10^{-2}$ \\
Ensemble    & $1.36\times 10^{-2}$ & $1.88\times 10^{-2}$ & $2.75\times 10^{-2}$ & $3.44\times 10^{-2}$ & $3.64\times 10^{-2}$ \\
Independent & $4.38\times 10^{-3}$ & $4.62\times 10^{-3}$ & $5.13\times 10^{-3}$ & $5.37\times 10^{-3}$ & $5.75\times 10^{-3}$ \\
\bottomrule
\end{tabular}

\caption{Sharpness for Accuracy}
\end{subtable}

\begin{subtable}[h]{\linewidth}
\small
\centering
\begin{tabular}{lrrrrr}
\toprule
corruption &        0 &        1 &        2 &        3 &        4 \\
model       &          &          &          &          &          \\
\midrule
Dropout     & $1.67\times 10^{-2}$ & $1.73\times 10^{-2}$ & $1.99\times 10^{-2}$ & $2.21\times 10^{-2}$ & $2.79\times 10^{-2}$ \\
Ensemble    & $7.01\times 10^{-3}$ & $7.97\times 10^{-3}$ & $1.14\times 10^{-2}$ & $1.39\times 10^{-2}$ & $1.56\times 10^{-2}$ \\
Independent & $2.63\times 10^{-3}$ & $2.81\times 10^{-3}$ & $3.05\times 10^{-3}$ & $3.26\times 10^{-3}$ & $3.58\times 10^{-3}$ \\
\bottomrule
\end{tabular}

\caption{Sharpness for ECE}
\end{subtable}

\begin{subfigure}{0.45\linewidth}
\centering
\centerline{\includegraphics[width=\columnwidth]{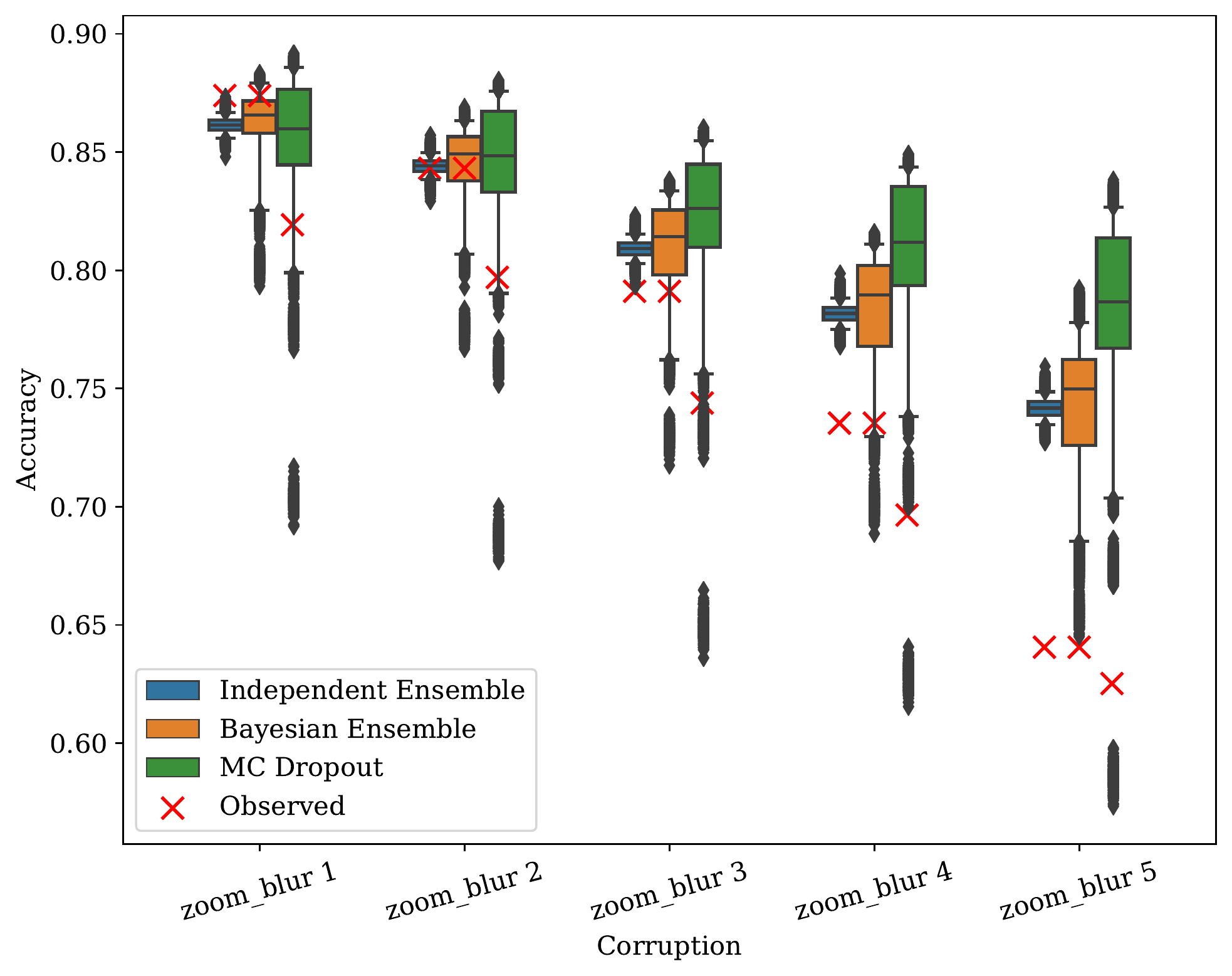}}
\caption{Posterior Check of Accuracy}
\end{subfigure} \hfill
\begin{subfigure}{0.45\linewidth}
\centering
\centerline{\includegraphics[width=\columnwidth]{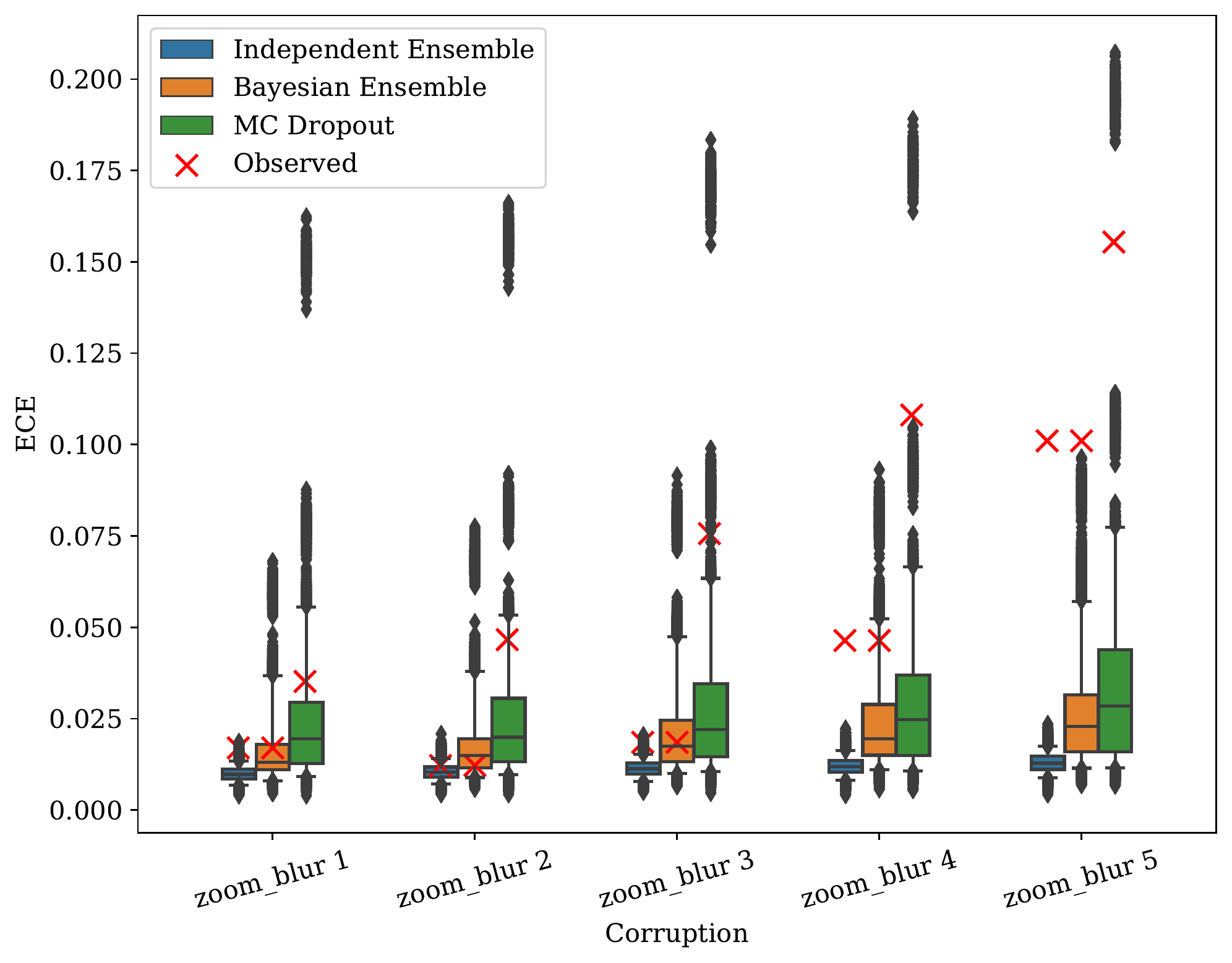}}
\caption{Posterior Check of ECE}
\end{subfigure}
\caption{Posterior predictive checks of recalibrated models trained on CIFAR-10, evaluated on zoom blur data from CIFAR-10-C.}
\end{minipage}
\end{figure}
    
\begin{figure}
\begin{minipage}{\textwidth}
\begin{subtable}[h]{0.45\textwidth}
\small
\begin{tabular}{lrrrrr}
\toprule
corruption &    0 &    1 &    2 &    3 &    4 \\
model       &      &      &      &      &      \\
\midrule
Dropout     & 0.87 & 0.90 & 0.51 & 0.13 & 0.02 \\
Ensemble    & 0.87 & 0.90 & 0.35 & 0.09 & 0.00 \\
Independent & 0.99 & 1.00 & 0.49 & 0.00 & 0.00 \\
\bottomrule
\end{tabular}

\caption{p-value for Accuracy}
\end{subtable}\hfill
\begin{subtable}[h]{0.45\textwidth}
\small
\begin{tabular}{lrrrrr}
\toprule
corruption &    0 &    1 &    2 &    3 &    4 \\
model       &      &      &      &      &      \\
\midrule
Dropout     & 0.20 & 0.21 & 0.04 & 0.65 & 0.98 \\
Ensemble    & 0.55 & 0.53 & 0.06 & 0.90 & 1.00 \\
Independent & 0.86 & 0.88 & 0.42 & 1.00 & 1.00 \\
\bottomrule
\end{tabular}

\caption{p-value for ECE}
\end{subtable}
\begin{subtable}[h]{\linewidth}
\small
\centering
\begin{tabular}{lrrrrr}
\toprule
corruption &        0 &        1 &        2 &        3 &        4 \\
model       &          &          &          &          &          \\
\midrule
Dropout     & $1.86\times 10^{-2}$ & $2.37\times 10^{-2}$ & $3.05\times 10^{-2}$ & $4.03\times 10^{-2}$ & $5.00\times 10^{-2}$ \\
Ensemble    & $5.37\times 10^{-3}$ & $7.38\times 10^{-3}$ & $1.70\times 10^{-2}$ & $2.38\times 10^{-2}$ & $5.27\times 10^{-2}$ \\
Independent & $3.25\times 10^{-3}$ & $3.63\times 10^{-3}$ & $4.25\times 10^{-3}$ & $5.13\times 10^{-3}$ & $6.00\times 10^{-3}$ \\
\bottomrule
\end{tabular}

\caption{Sharpness for Accuracy}
\end{subtable}

\begin{subtable}[h]{\linewidth}
\small
\centering
\begin{tabular}{lrrrrr}
\toprule
corruption &        0 &        1 &        2 &        3 &        4 \\
model       &          &          &          &          &          \\
\midrule
Dropout     & $9.52\times 10^{-3}$ & $1.12\times 10^{-2}$ & $1.50\times 10^{-2}$ & $2.15\times 10^{-2}$ & $3.24\times 10^{-2}$ \\
Ensemble    & $3.00\times 10^{-3}$ & $3.42\times 10^{-3}$ & $6.93\times 10^{-3}$ & $1.07\times 10^{-2}$ & $2.00\times 10^{-2}$ \\
Independent & $1.95\times 10^{-3}$ & $2.19\times 10^{-3}$ & $2.58\times 10^{-3}$ & $2.94\times 10^{-3}$ & $3.55\times 10^{-3}$ \\
\bottomrule
\end{tabular}

\caption{Sharpness for ECE}
\end{subtable}

\begin{subfigure}{0.45\linewidth}
\centering
\centerline{\includegraphics[width=\columnwidth]{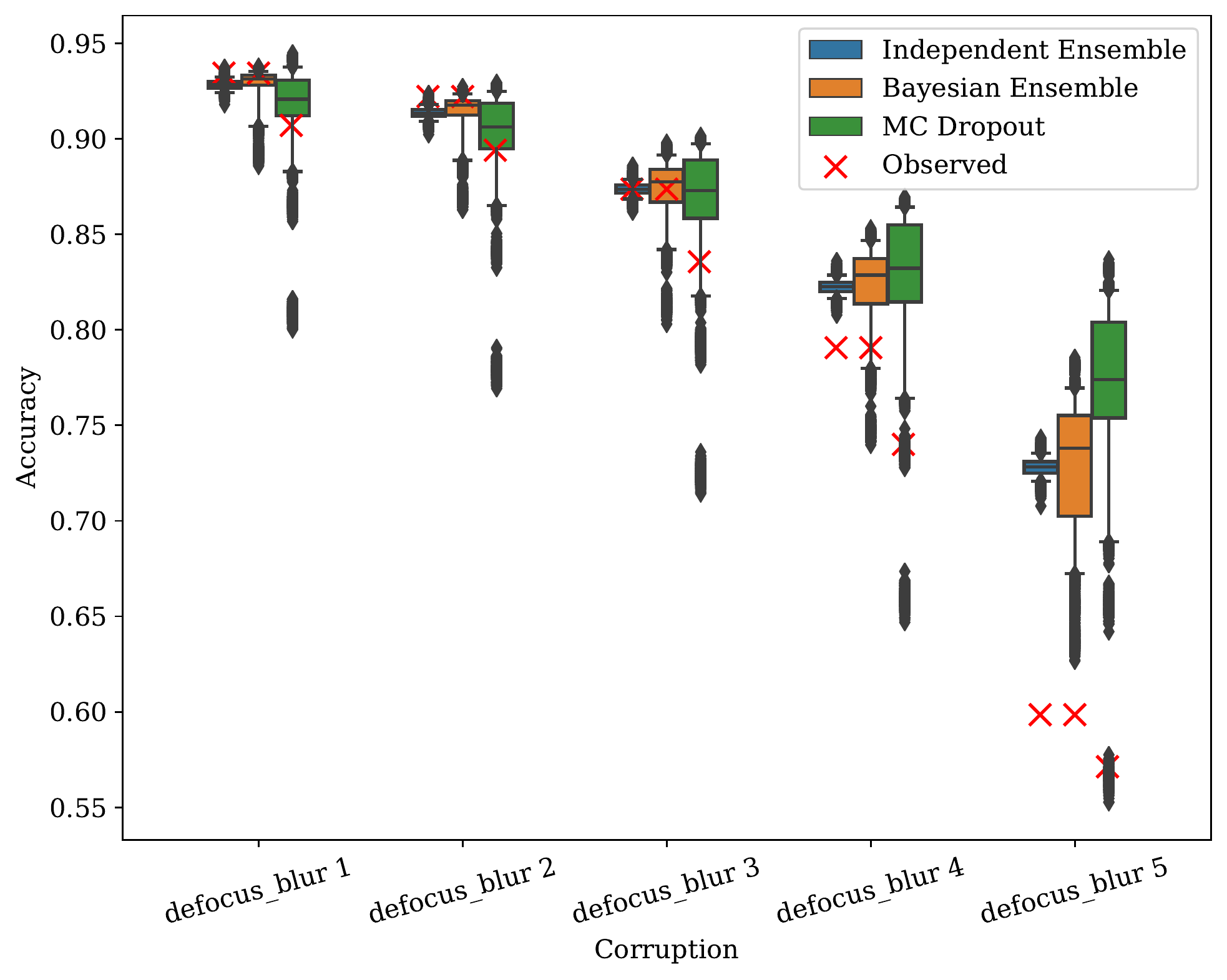}}
\caption{Posterior Check of Accuracy}
\end{subfigure} \hfill
\begin{subfigure}{0.45\linewidth}
\centering
\centerline{\includegraphics[width=\columnwidth]{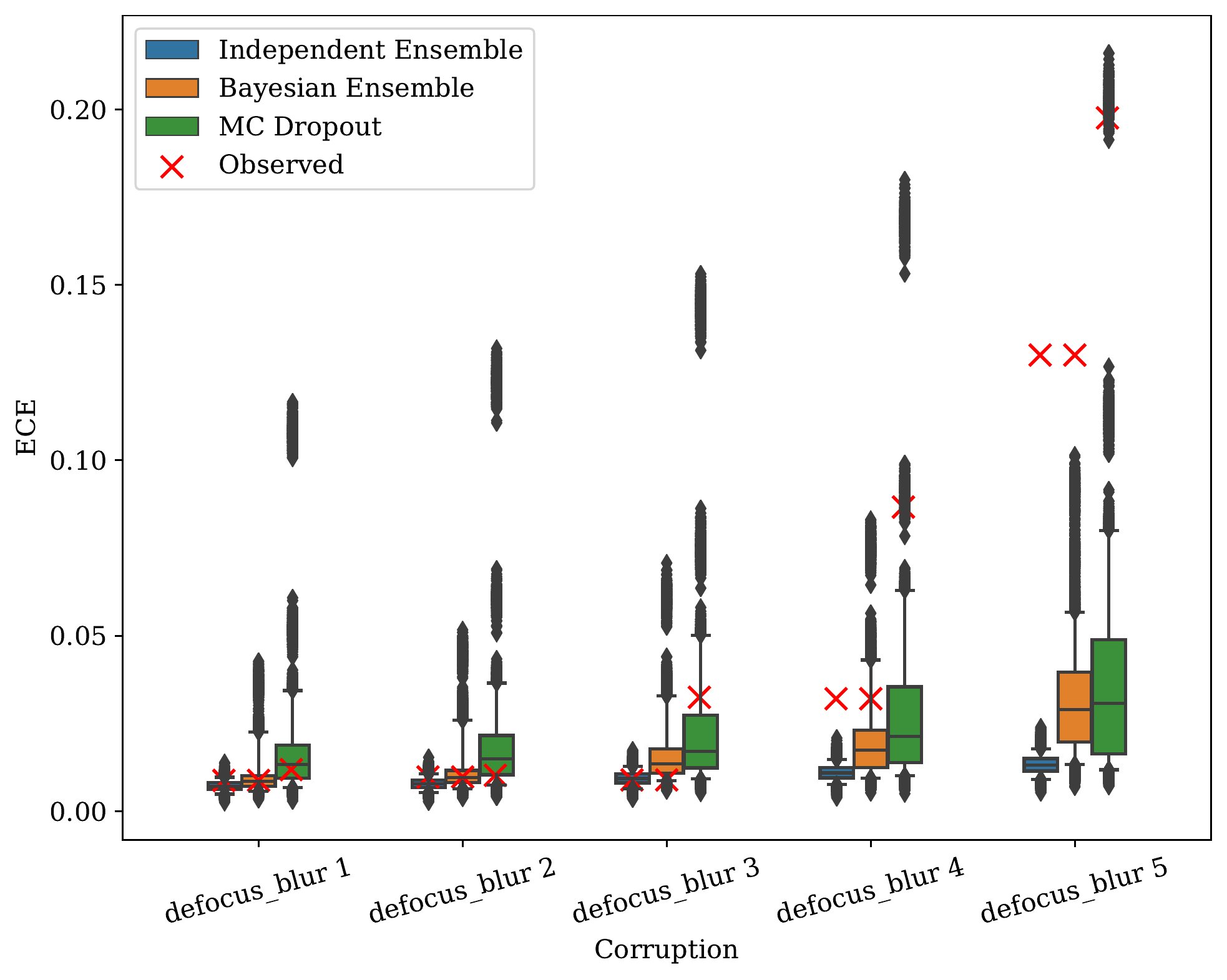}}
\caption{Posterior Check of ECE}
\end{subfigure}
\caption{Posterior predictive checks of recalibrated models trained on CIFAR-10, evaluated on defocus blur data from CIFAR-10-C.}\label{fig:defocus_blur}
\end{minipage}
\end{figure}
    
\begin{figure}
\begin{minipage}{\textwidth}
\begin{subtable}[h]{0.45\textwidth}
\small
\begin{tabular}{lrrrrr}
\toprule
corruption &    0 &    1 &    2 &    3 &    4 \\
model       &      &      &      &      &      \\
\midrule
Dropout     & 0.65 & 0.18 & 0.08 & 0.00 & 0.00 \\
Ensemble    & 0.68 & 0.25 & 0.10 & 0.00 & 0.00 \\
Independent & 0.97 & 0.15 & 0.00 & 0.00 & 0.00 \\
\bottomrule
\end{tabular}

\caption{p-value for Accuracy}
\end{subtable}\hfill
\begin{subtable}[h]{0.45\textwidth}
\small
\begin{tabular}{lrrrrr}
\toprule
corruption &    0 &    1 &    2 &    3 &    4 \\
model       &      &      &      &      &      \\
\midrule
Dropout     & 0.37 & 0.22 & 0.88 & 0.99 & 1.00 \\
Ensemble    & 0.75 & 0.34 & 0.92 & 1.00 & 1.00 \\
Independent & 0.98 & 0.79 & 1.00 & 1.00 & 1.00 \\
\bottomrule
\end{tabular}

\caption{p-value for ECE}
\end{subtable}
\begin{subtable}[h]{\linewidth}
\small
\centering
\begin{tabular}{lrrrrr}
\toprule
corruption &        0 &        1 &        2 &        3 &        4 \\
model       &          &          &          &          &          \\
\midrule
Dropout     & $2.12\times 10^{-2}$ & $2.40\times 10^{-2}$ & $2.41\times 10^{-2}$ & $3.49\times 10^{-2}$ & $4.34\times 10^{-2}$ \\
Ensemble    & $5.25\times 10^{-3}$ & $1.20\times 10^{-2}$ & $1.40\times 10^{-2}$ & $1.80\times 10^{-2}$ & $4.90\times 10^{-2}$ \\
Independent & $3.75\times 10^{-3}$ & $4.25\times 10^{-3}$ & $4.62\times 10^{-3}$ & $5.37\times 10^{-3}$ & $6.13\times 10^{-3}$ \\
\bottomrule
\end{tabular}

\caption{Sharpness for Accuracy}
\end{subtable}

\begin{subtable}[h]{\linewidth}
\small
\centering
\begin{tabular}{lrrrrr}
\toprule
corruption &        0 &        1 &        2 &        3 &        4 \\
model       &          &          &          &          &          \\
\midrule
Dropout     & $1.08\times 10^{-2}$ & $1.17\times 10^{-2}$ & $1.28\times 10^{-2}$ & $1.86\times 10^{-2}$ & $2.33\times 10^{-2}$ \\
Ensemble    & $3.53\times 10^{-3}$ & $4.55\times 10^{-3}$ & $6.29\times 10^{-3}$ & $9.35\times 10^{-3}$ & $1.60\times 10^{-2}$ \\
Independent & $2.20\times 10^{-3}$ & $2.53\times 10^{-3}$ & $2.81\times 10^{-3}$ & $3.27\times 10^{-3}$ & $3.51\times 10^{-3}$ \\
\bottomrule
\end{tabular}

\caption{Sharpness for ECE}
\end{subtable}

\begin{subfigure}{0.45\linewidth}
\centering
\centerline{\includegraphics[width=\columnwidth]{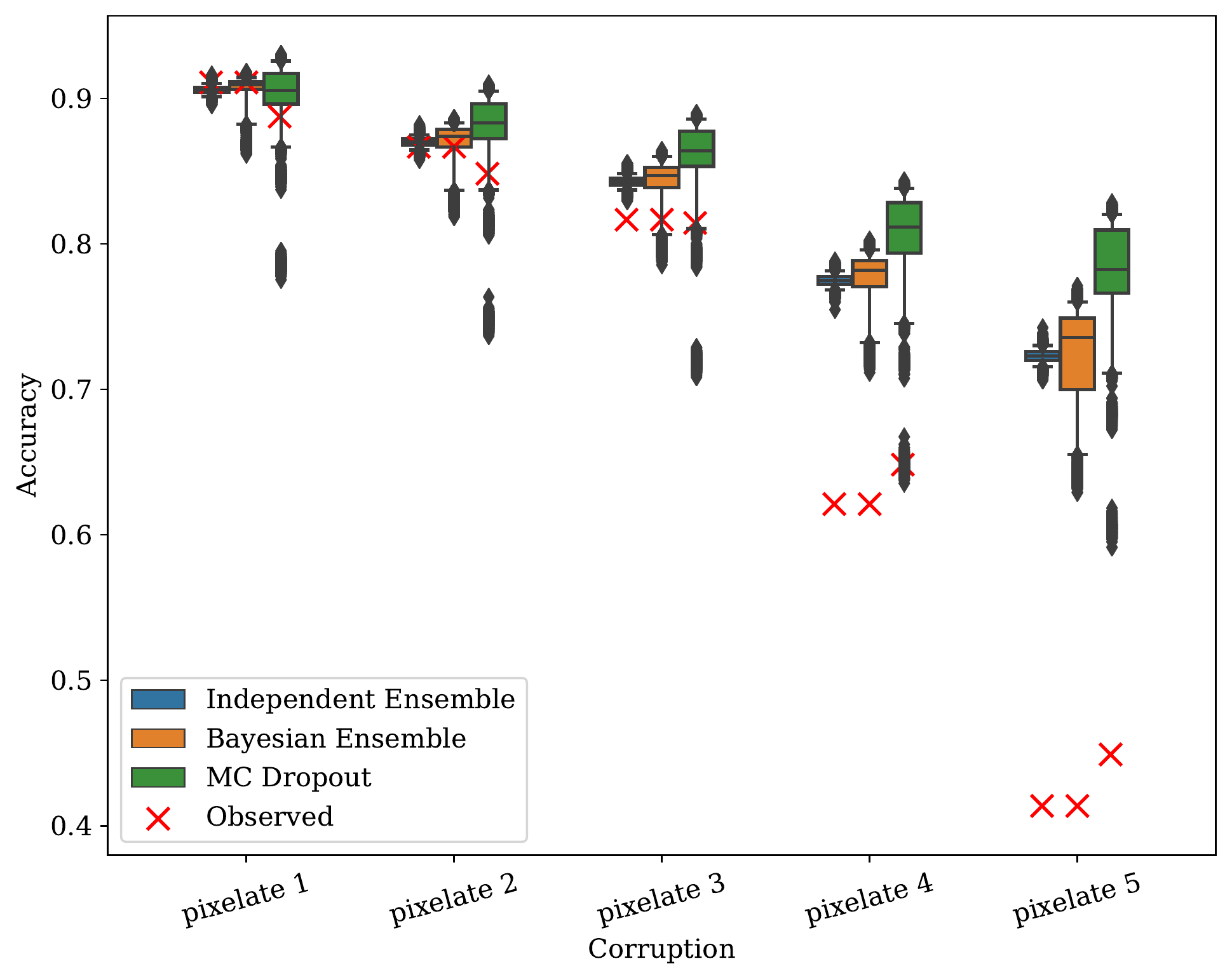}}
\caption{Posterior Check of Accuracy}
\end{subfigure} \hfill
\begin{subfigure}{0.45\linewidth}
\centering
\centerline{\includegraphics[width=\columnwidth]{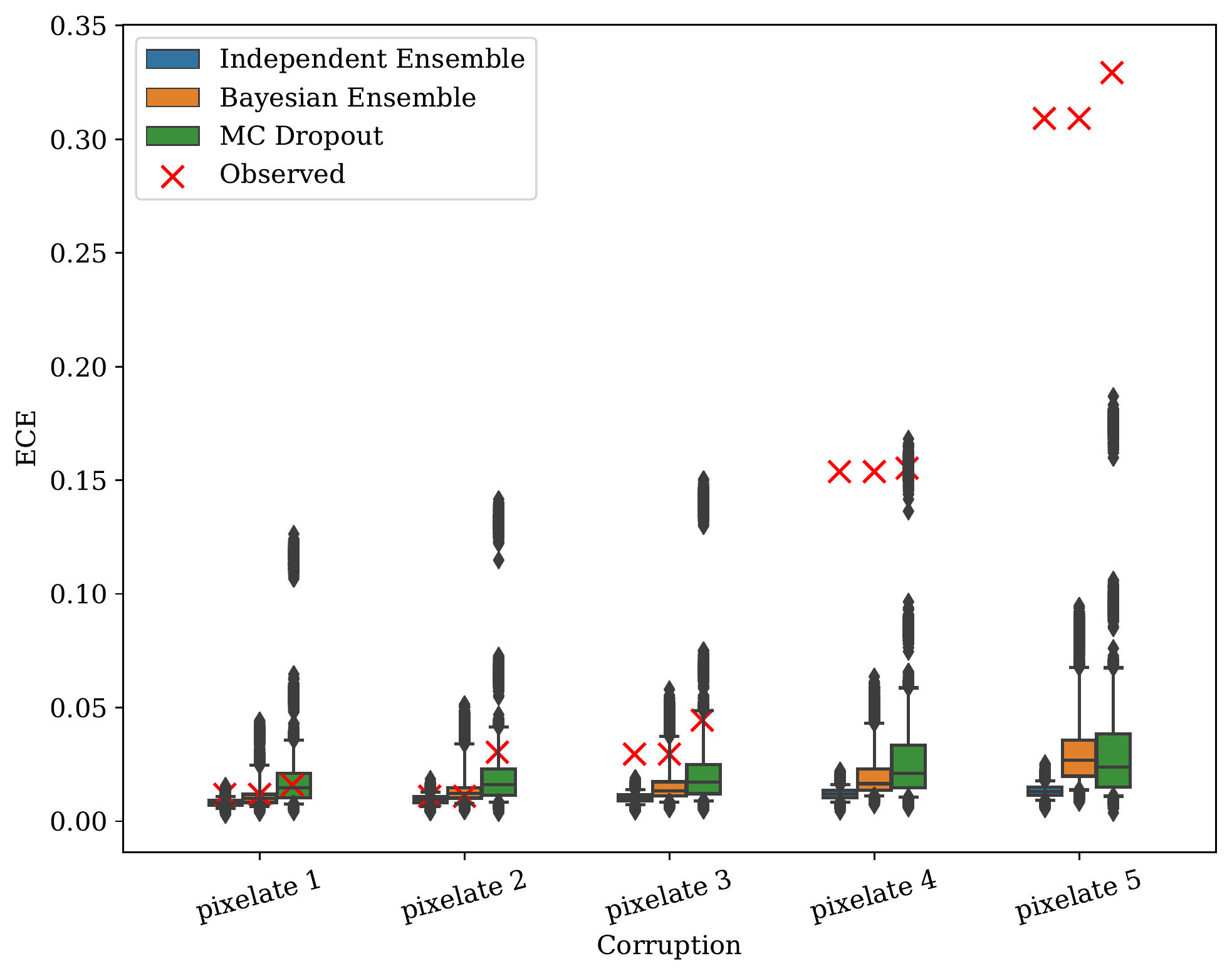}}
\caption{Posterior Check of ECE}
\end{subfigure}
\caption{Posterior predictive checks of recalibrated models trained on CIFAR-10, evaluated on pixelate data from CIFAR-10-C.}\label{fig:pixelate}
\end{minipage}
\end{figure}
    
\begin{figure}
\begin{minipage}{\textwidth}
\begin{subtable}[h]{0.45\textwidth}
\small
\begin{tabular}{lrrrrr}
\toprule
corruption &    0 &    1 &    2 &    3 &    4 \\
model       &      &      &      &      &      \\
\midrule
Dropout     & 0.71 & 0.59 & 0.95 & 0.86 & 0.39 \\
Ensemble    & 0.47 & 0.24 & 0.99 & 0.89 & 0.36 \\
Independent & 0.85 & 0.29 & 1.00 & 1.00 & 0.67 \\
\bottomrule
\end{tabular}

\caption{p-value for Accuracy}
\end{subtable}\hfill
\begin{subtable}[h]{0.45\textwidth}
\small
\begin{tabular}{lrrrrr}
\toprule
corruption &    0 &    1 &    2 &    3 &    4 \\
model       &      &      &      &      &      \\
\midrule
Dropout     & 0.25 & 0.01 & 0.50 & 0.41 & 0.16 \\
Ensemble    & 0.50 & 0.07 & 0.88 & 0.80 & 0.25 \\
Independent & 0.90 & 0.20 & 1.00 & 1.00 & 0.66 \\
\bottomrule
\end{tabular}

\caption{p-value for ECE}
\end{subtable}
\begin{subtable}[h]{\linewidth}
\small
\centering
\begin{tabular}{lrrrrr}
\toprule
corruption &        0 &        1 &        2 &        3 &        4 \\
model       &          &          &          &          &          \\
\midrule
Dropout     & $2.17\times 10^{-2}$ & $2.49\times 10^{-2}$ & $2.05\times 10^{-2}$ & $2.55\times 10^{-2}$ & $2.96\times 10^{-2}$ \\
Ensemble    & $6.12\times 10^{-3}$ & $7.75\times 10^{-3}$ & $5.75\times 10^{-3}$ & $9.66\times 10^{-3}$ & $1.46\times 10^{-2}$ \\
Independent & $3.63\times 10^{-3}$ & $3.87\times 10^{-3}$ & $3.37\times 10^{-3}$ & $4.00\times 10^{-3}$ & $4.50\times 10^{-3}$ \\
\bottomrule
\end{tabular}

\caption{Sharpness for Accuracy}
\end{subtable}

\begin{subtable}[h]{\linewidth}
\small
\centering
\begin{tabular}{lrrrrr}
\toprule
corruption &        0 &        1 &        2 &        3 &        4 \\
model       &          &          &          &          &          \\
\midrule
Dropout     & $1.26\times 10^{-2}$ & $1.31\times 10^{-2}$ & $9.52\times 10^{-3}$ & $1.09\times 10^{-2}$ & $1.51\times 10^{-2}$ \\
Ensemble    & $3.42\times 10^{-3}$ & $3.57\times 10^{-3}$ & $2.93\times 10^{-3}$ & $4.37\times 10^{-3}$ & $5.23\times 10^{-3}$ \\
Independent & $2.21\times 10^{-3}$ & $2.39\times 10^{-3}$ & $2.04\times 10^{-3}$ & $2.33\times 10^{-3}$ & $2.71\times 10^{-3}$ \\
\bottomrule
\end{tabular}

\caption{Sharpness for ECE}
\end{subtable}

\begin{subfigure}{0.45\linewidth}
\centering
\centerline{\includegraphics[width=\columnwidth]{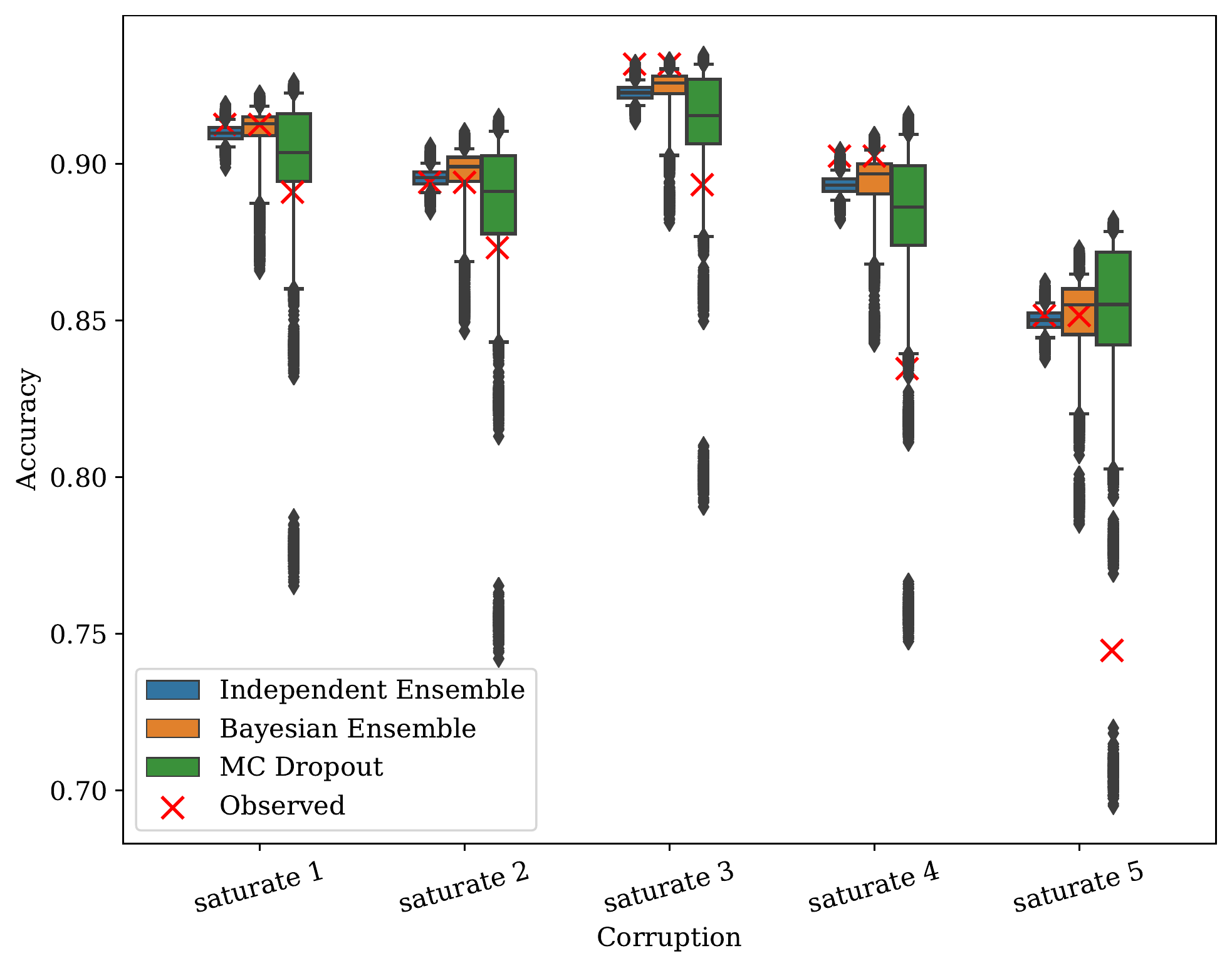}}
\caption{Posterior Check of Accuracy}
\end{subfigure} \hfill
\begin{subfigure}{0.45\linewidth}
\centering
\centerline{\includegraphics[width=\columnwidth]{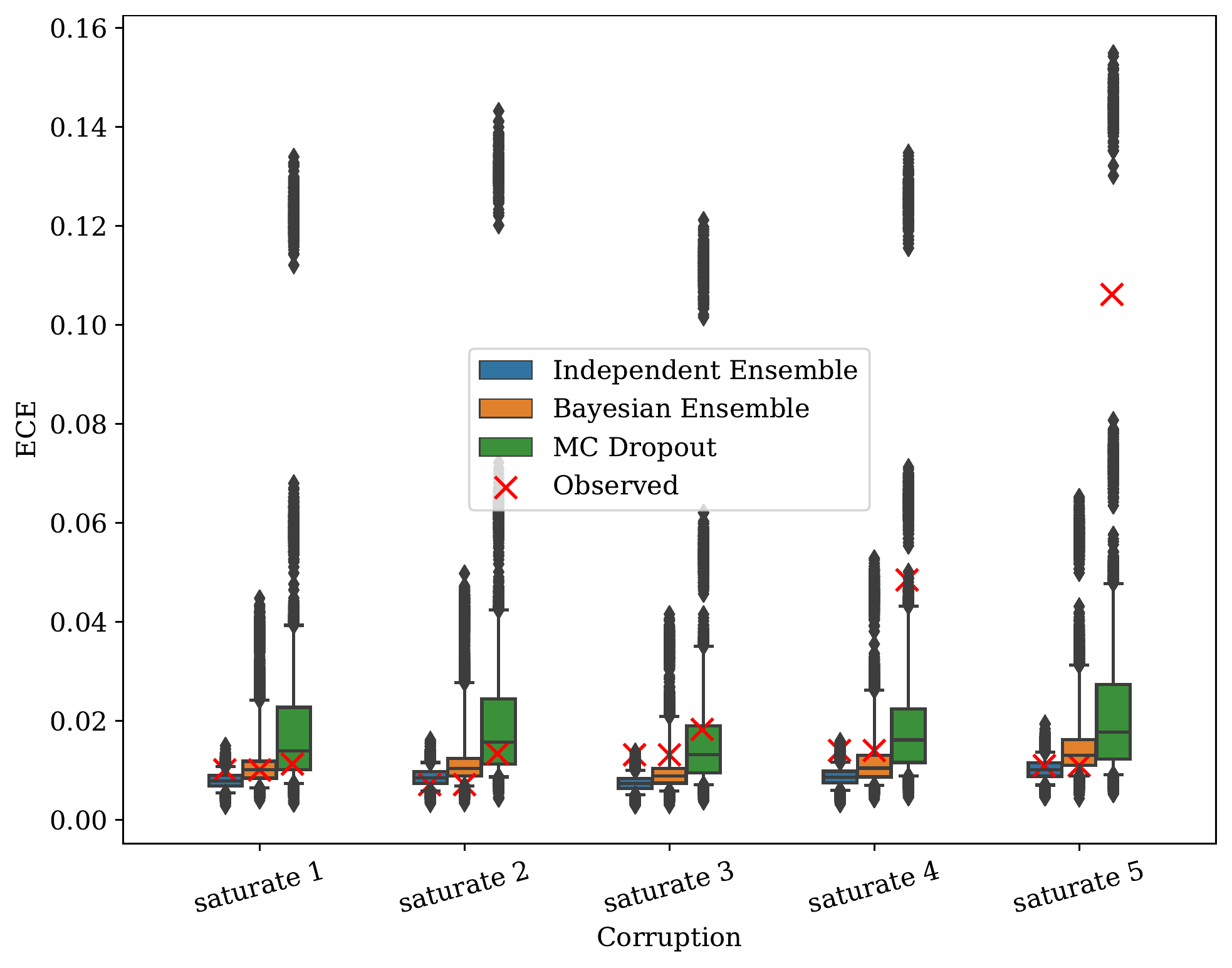}}
\caption{Posterior Check of ECE}
\end{subfigure}
\caption{Posterior predictive checks of recalibrated models trained on CIFAR-10, evaluated on saturate data from CIFAR-10-C.}\label{fig:saturate}
\end{minipage}
\end{figure}
    
\begin{figure}
\begin{minipage}{\textwidth}
\begin{subtable}[h]{0.45\textwidth}
\small
\begin{tabular}{lrrrrr}
\toprule
corruption &    0 &    1 &    2 &    3 &    4 \\
model       &      &      &      &      &      \\
\midrule
Dropout     & 0.66 & 0.32 & 0.22 & 0.10 & 0.02 \\
Ensemble    & 0.63 & 0.25 & 0.22 & 0.09 & 0.00 \\
Independent & 0.95 & 0.26 & 0.06 & 0.00 & 0.00 \\
\bottomrule
\end{tabular}

\caption{p-value for Accuracy}
\end{subtable}\hfill
\begin{subtable}[h]{0.45\textwidth}
\small
\begin{tabular}{lrrrrr}
\toprule
corruption &    0 &    1 &    2 &    3 &    4 \\
model       &      &      &      &      &      \\
\midrule
Dropout     & 0.12 & 0.49 & 0.46 & 0.85 & 0.96 \\
Ensemble    & 0.31 & 0.82 & 0.78 & 0.94 & 1.00 \\
Independent & 0.69 & 1.00 & 1.00 & 1.00 & 1.00 \\
\bottomrule
\end{tabular}

\caption{p-value for ECE}
\end{subtable}
\begin{subtable}[h]{\linewidth}
\small
\centering
\begin{tabular}{lrrrrr}
\toprule
corruption &        0 &        1 &        2 &        3 &        4 \\
model       &          &          &          &          &          \\
\midrule
Dropout     & $2.12\times 10^{-2}$ & $2.55\times 10^{-2}$ & $3.05\times 10^{-2}$ & $2.58\times 10^{-2}$ & $3.08\times 10^{-2}$ \\
Ensemble    & $7.13\times 10^{-3}$ & $1.02\times 10^{-2}$ & $1.21\times 10^{-2}$ & $1.02\times 10^{-2}$ & $1.75\times 10^{-2}$ \\
Independent & $3.75\times 10^{-3}$ & $4.38\times 10^{-3}$ & $4.62\times 10^{-3}$ & $4.50\times 10^{-3}$ & $5.00\times 10^{-3}$ \\
\bottomrule
\end{tabular}

\caption{Sharpness for Accuracy}
\end{subtable}

\begin{subtable}[h]{\linewidth}
\small
\centering
\begin{tabular}{lrrrrr}
\toprule
corruption &        0 &        1 &        2 &        3 &        4 \\
model       &          &          &          &          &          \\
\midrule
Dropout     & $1.08\times 10^{-2}$ & $1.34\times 10^{-2}$ & $1.53\times 10^{-2}$ & $1.29\times 10^{-2}$ & $1.58\times 10^{-2}$ \\
Ensemble    & $3.30\times 10^{-3}$ & $4.93\times 10^{-3}$ & $5.41\times 10^{-3}$ & $4.91\times 10^{-3}$ & $7.11\times 10^{-3}$ \\
Independent & $2.24\times 10^{-3}$ & $2.63\times 10^{-3}$ & $2.87\times 10^{-3}$ & $2.66\times 10^{-3}$ & $3.04\times 10^{-3}$ \\
\bottomrule
\end{tabular}

\caption{Sharpness for ECE}
\end{subtable}

\begin{subfigure}{0.45\linewidth}
\centering
\centerline{\includegraphics[width=\columnwidth]{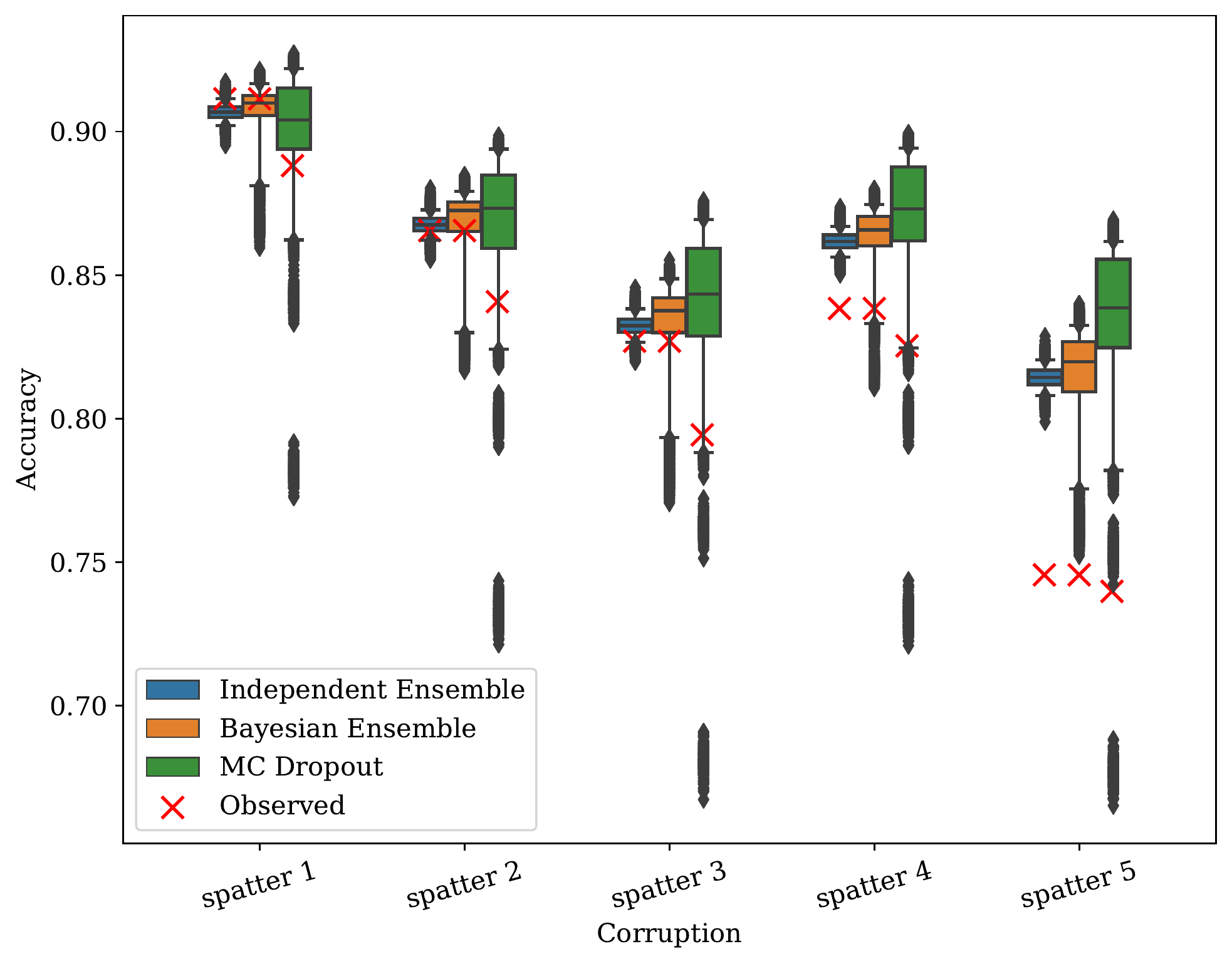}}
\caption{Posterior Check of Accuracy}
\end{subfigure} \hfill
\begin{subfigure}{0.45\linewidth}
\centering
\centerline{\includegraphics[width=\columnwidth]{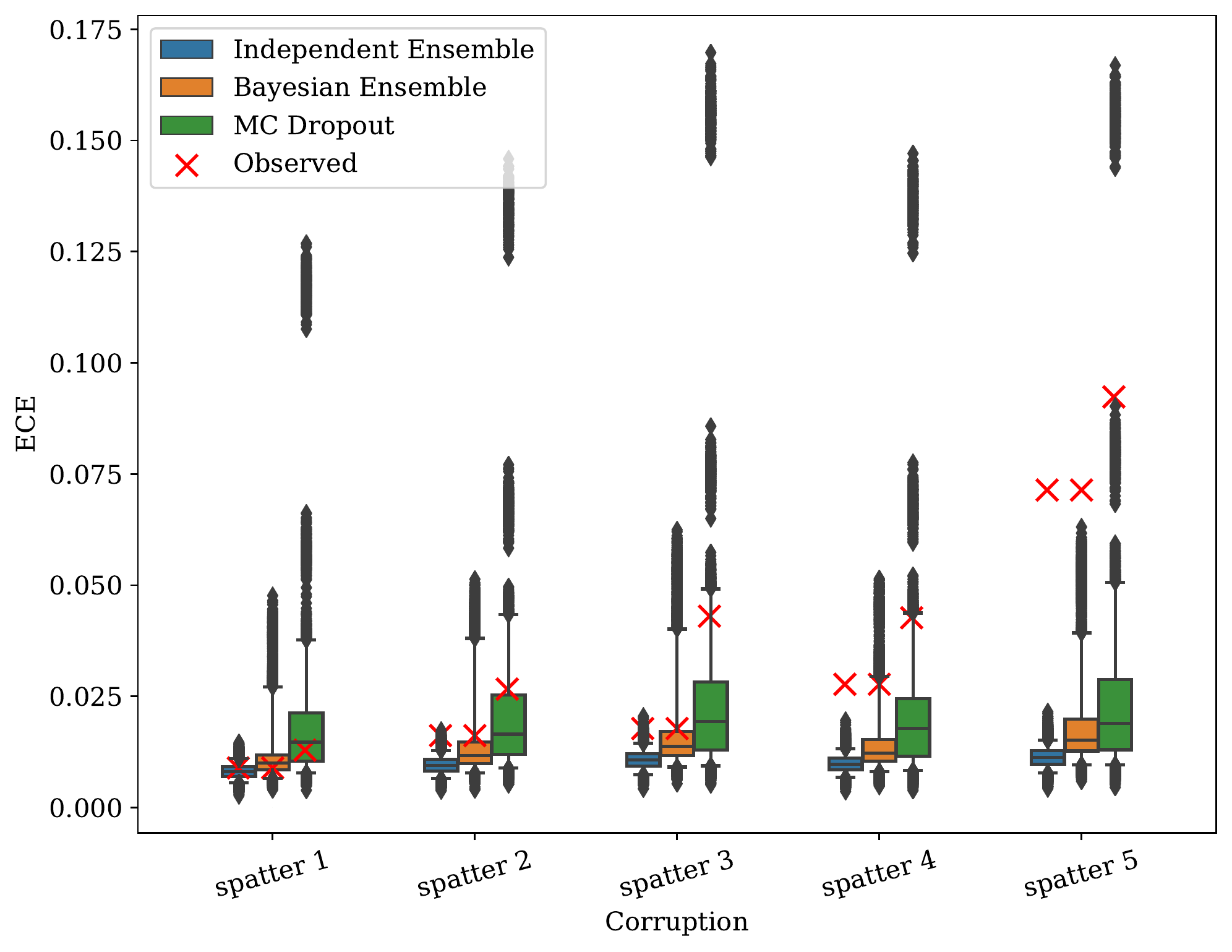}}
\caption{Posterior Check of ECE}
\end{subfigure}
\caption{Posterior predictive checks of recalibrated models trained on CIFAR-10, evaluated on spatter data from CIFAR-10-C.}\label{fig:spatter}
\end{minipage}
\end{figure}
    
\begin{figure}
\begin{minipage}{\textwidth}
\begin{subtable}[h]{0.45\textwidth}
\small
\begin{tabular}{lrrrrr}
\toprule
corruption &    0 &    1 &    2 &    3 &    4 \\
model       &      &      &      &      &      \\
\midrule
Dropout     & 0.92 & 0.80 & 0.65 & 0.37 & 0.02 \\
Ensemble    & 0.69 & 0.42 & 0.20 & 0.16 & 0.00 \\
Independent & 0.98 & 0.94 & 0.04 & 0.00 & 0.00 \\
\bottomrule
\end{tabular}

\caption{p-value for Accuracy}
\end{subtable}\hfill
\begin{subtable}[h]{0.45\textwidth}
\small
\begin{tabular}{lrrrrr}
\toprule
corruption &    0 &    1 &    2 &    3 &    4 \\
model       &      &      &      &      &      \\
\midrule
Dropout     & 0.36 & 0.42 & 0.06 & 0.49 & 0.98 \\
Ensemble    & 0.79 & 0.82 & 0.17 & 0.83 & 1.00 \\
Independent & 1.00 & 1.00 & 0.64 & 1.00 & 1.00 \\
\bottomrule
\end{tabular}

\caption{p-value for ECE}
\end{subtable}
\begin{subtable}[h]{\linewidth}
\small
\centering
\begin{tabular}{lrrrrr}
\toprule
corruption &        0 &        1 &        2 &        3 &        4 \\
model       &          &          &          &          &          \\
\midrule
Dropout     & $2.12\times 10^{-2}$ & $3.16\times 10^{-2}$ & $4.10\times 10^{-2}$ & $6.53\times 10^{-2}$ & $1.40\times 10^{-1}$ \\
Ensemble    & $5.50\times 10^{-3}$ & $9.75\times 10^{-3}$ & $1.19\times 10^{-2}$ & $2.34\times 10^{-2}$ & $8.96\times 10^{-2}$ \\
Independent & $3.25\times 10^{-3}$ & $4.00\times 10^{-3}$ & $4.38\times 10^{-3}$ & $5.25\times 10^{-3}$ & $6.88\times 10^{-3}$ \\
\bottomrule
\end{tabular}

\caption{Sharpness for Accuracy}
\end{subtable}

\begin{subtable}[h]{\linewidth}
\small
\centering
\begin{tabular}{lrrrrr}
\toprule
corruption &        0 &        1 &        2 &        3 &        4 \\
model       &          &          &          &          &          \\
\midrule
Dropout     & $1.06\times 10^{-2}$ & $1.64\times 10^{-2}$ & $2.23\times 10^{-2}$ & $4.02\times 10^{-2}$ & $8.96\times 10^{-2}$ \\
Ensemble    & $3.58\times 10^{-3}$ & $3.86\times 10^{-3}$ & $6.72\times 10^{-3}$ & $1.50\times 10^{-2}$ & $6.71\times 10^{-2}$ \\
Independent & $1.97\times 10^{-3}$ & $2.41\times 10^{-3}$ & $2.70\times 10^{-3}$ & $3.22\times 10^{-3}$ & $4.22\times 10^{-3}$ \\
\bottomrule
\end{tabular}

\caption{Sharpness for ECE}
\end{subtable}

\begin{subfigure}{0.45\linewidth}
\centering
\centerline{\includegraphics[width=\columnwidth]{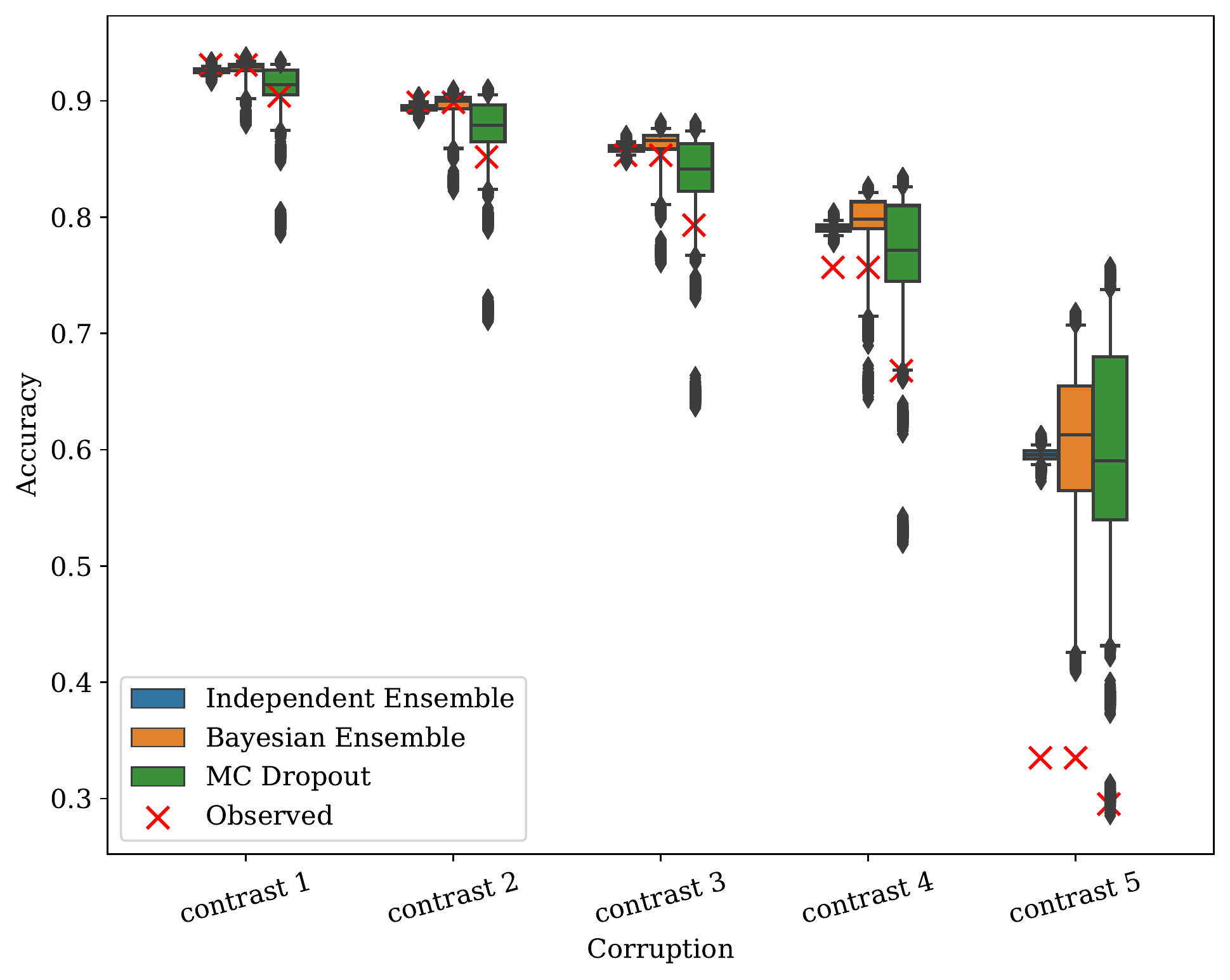}}
\caption{Posterior Check of Accuracy}
\end{subfigure} \hfill
\begin{subfigure}{0.45\linewidth}
\centering
\centerline{\includegraphics[width=\columnwidth]{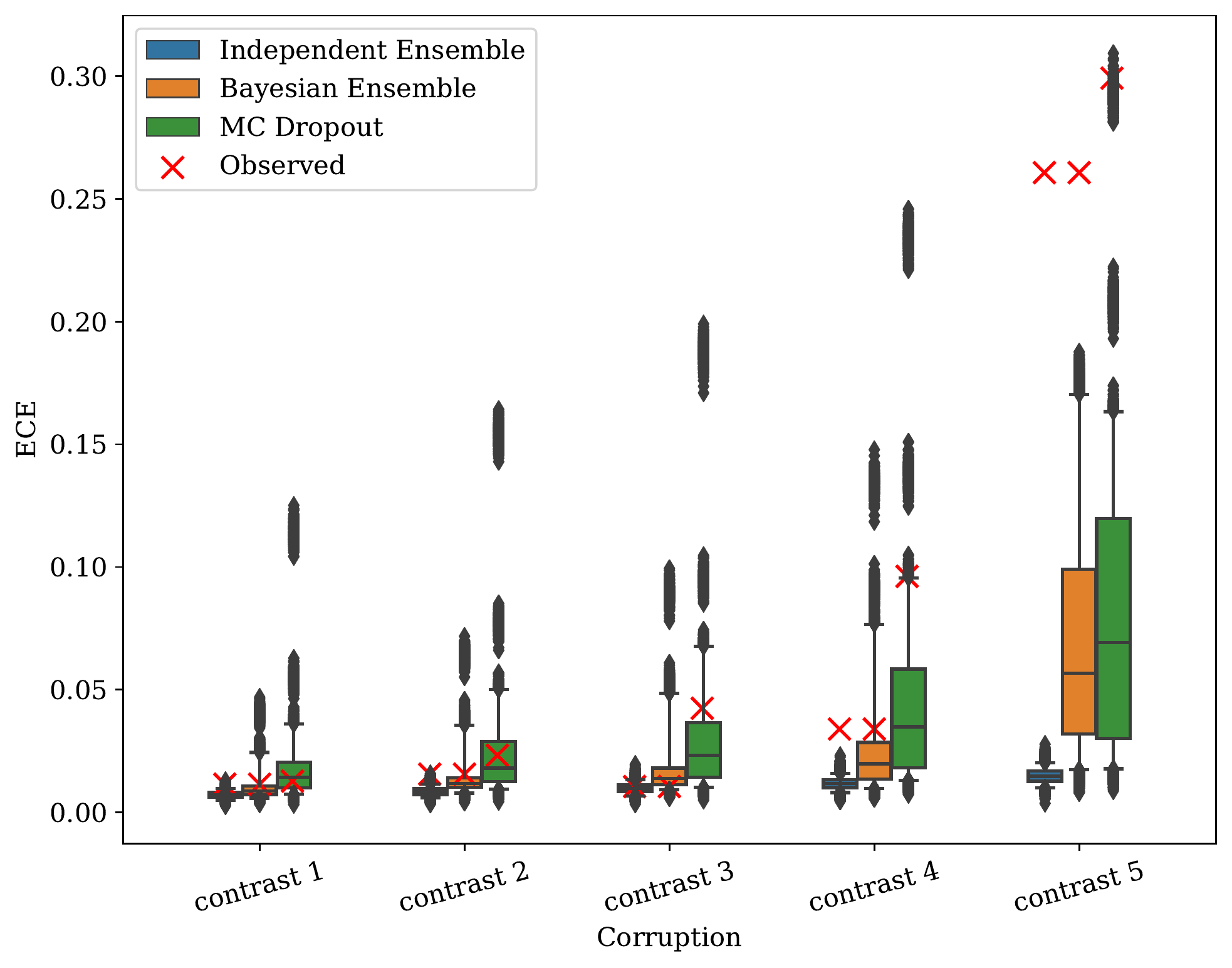}}
\caption{Posterior Check of ECE}
\end{subfigure}
\caption{Posterior predictive checks of recalibrated models trained on CIFAR-10, evaluated on contrast data from CIFAR-10-C.}\label{fig:contrast}
\end{minipage}
\end{figure}
    
\begin{figure}
\begin{minipage}{\textwidth}
\begin{subtable}[h]{0.45\textwidth}
\small
\begin{tabular}{lrrrrr}
\toprule
corruption &    0 &    1 &    2 &    3 &    4 \\
model       &      &      &      &      &      \\
\midrule
Dropout     & 0.10 & 0.02 & 0.00 & 0.00 & 0.00 \\
Ensemble    & 0.20 & 0.00 & 0.00 & 0.00 & 0.00 \\
Independent & 0.00 & 0.00 & 0.00 & 0.00 & 0.00 \\
\bottomrule
\end{tabular}

\caption{p-value for Accuracy}
\end{subtable}\hfill
\begin{subtable}[h]{0.45\textwidth}
\small
\begin{tabular}{lrrrrr}
\toprule
corruption &    0 &    1 &    2 &    3 &    4 \\
model       &      &      &      &      &      \\
\midrule
Dropout     & 0.38 & 0.98 & 1.00 & 1.00 & 1.00 \\
Ensemble    & 0.54 & 1.00 & 1.00 & 1.00 & 1.00 \\
Independent & 0.91 & 1.00 & 1.00 & 1.00 & 1.00 \\
\bottomrule
\end{tabular}

\caption{p-value for ECE}
\end{subtable}
\begin{subtable}[h]{\linewidth}
\small
\centering
\begin{tabular}{lrrrrr}
\toprule
corruption &        0 &        1 &        2 &        3 &        4 \\
model       &          &          &          &          &          \\
\midrule
Dropout     & $2.38\times 10^{-2}$ & $3.21\times 10^{-2}$ & $3.45\times 10^{-2}$ & $3.25\times 10^{-2}$ & $2.94\times 10^{-2}$ \\
Ensemble    & $1.13\times 10^{-2}$ & $1.89\times 10^{-2}$ & $2.44\times 10^{-2}$ & $4.22\times 10^{-2}$ & $5.37\times 10^{-2}$ \\
Independent & $4.50\times 10^{-3}$ & $5.37\times 10^{-3}$ & $5.75\times 10^{-3}$ & $6.13\times 10^{-3}$ & $6.37\times 10^{-3}$ \\
\bottomrule
\end{tabular}

\caption{Sharpness for Accuracy}
\end{subtable}

\begin{subtable}[h]{\linewidth}
\small
\centering
\begin{tabular}{lrrrrr}
\toprule
corruption &        0 &        1 &        2 &        3 &        4 \\
model       &          &          &          &          &          \\
\midrule
Dropout     & $1.38\times 10^{-2}$ & $1.56\times 10^{-2}$ & $1.84\times 10^{-2}$ & $1.70\times 10^{-2}$ & $1.79\times 10^{-2}$ \\
Ensemble    & $5.48\times 10^{-3}$ & $7.71\times 10^{-3}$ & $1.15\times 10^{-2}$ & $1.83\times 10^{-2}$ & $2.54\times 10^{-2}$ \\
Independent & $2.76\times 10^{-3}$ & $3.25\times 10^{-3}$ & $3.48\times 10^{-3}$ & $3.73\times 10^{-3}$ & $3.90\times 10^{-3}$ \\
\bottomrule
\end{tabular}

\caption{Sharpness for ECE}
\end{subtable}

\begin{subfigure}{0.45\linewidth}
\centering
\centerline{\includegraphics[width=\columnwidth]{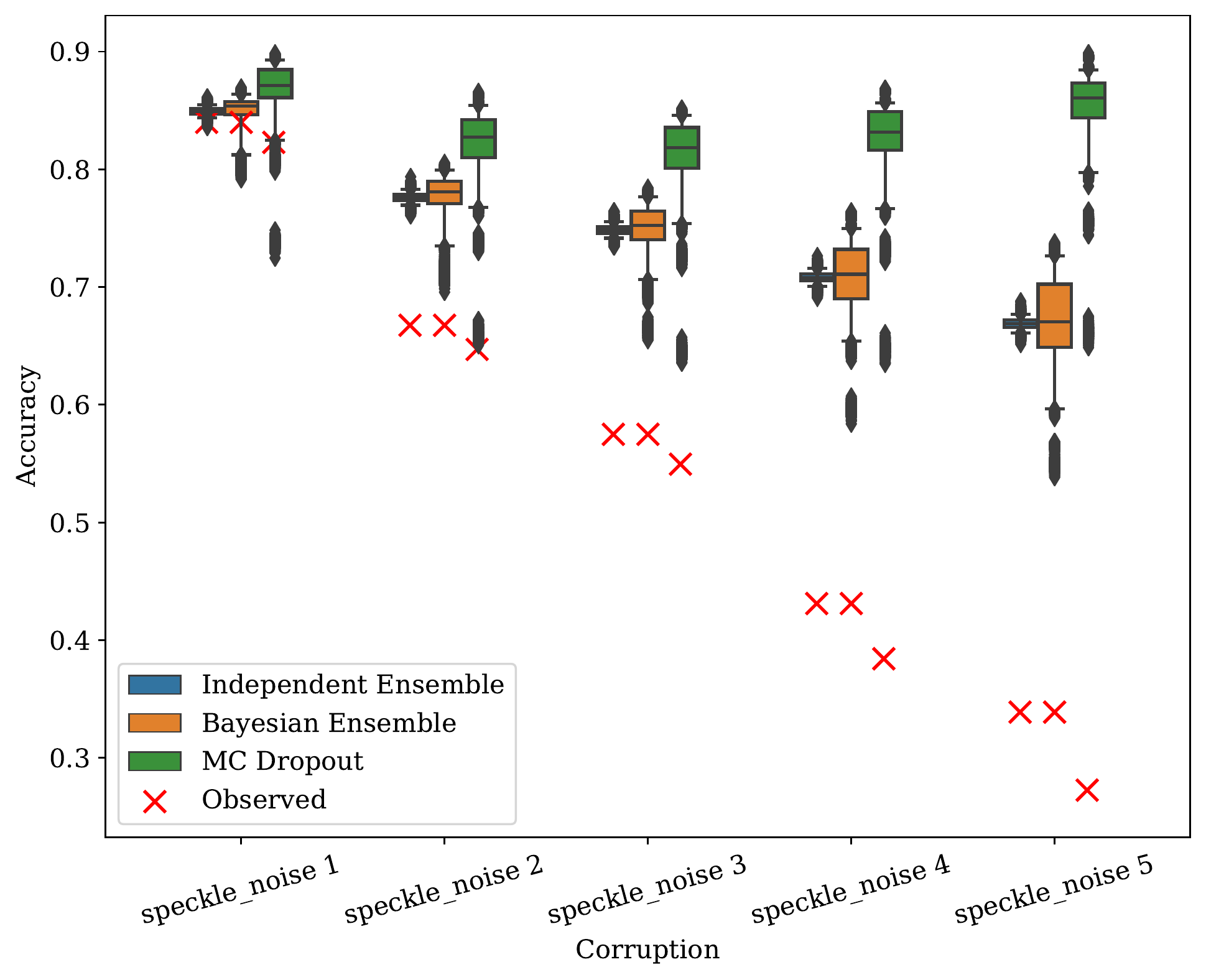}}
\caption{Posterior Check of Accuracy}
\end{subfigure} \hfill
\begin{subfigure}{0.45\linewidth}
\centering
\centerline{\includegraphics[width=\columnwidth]{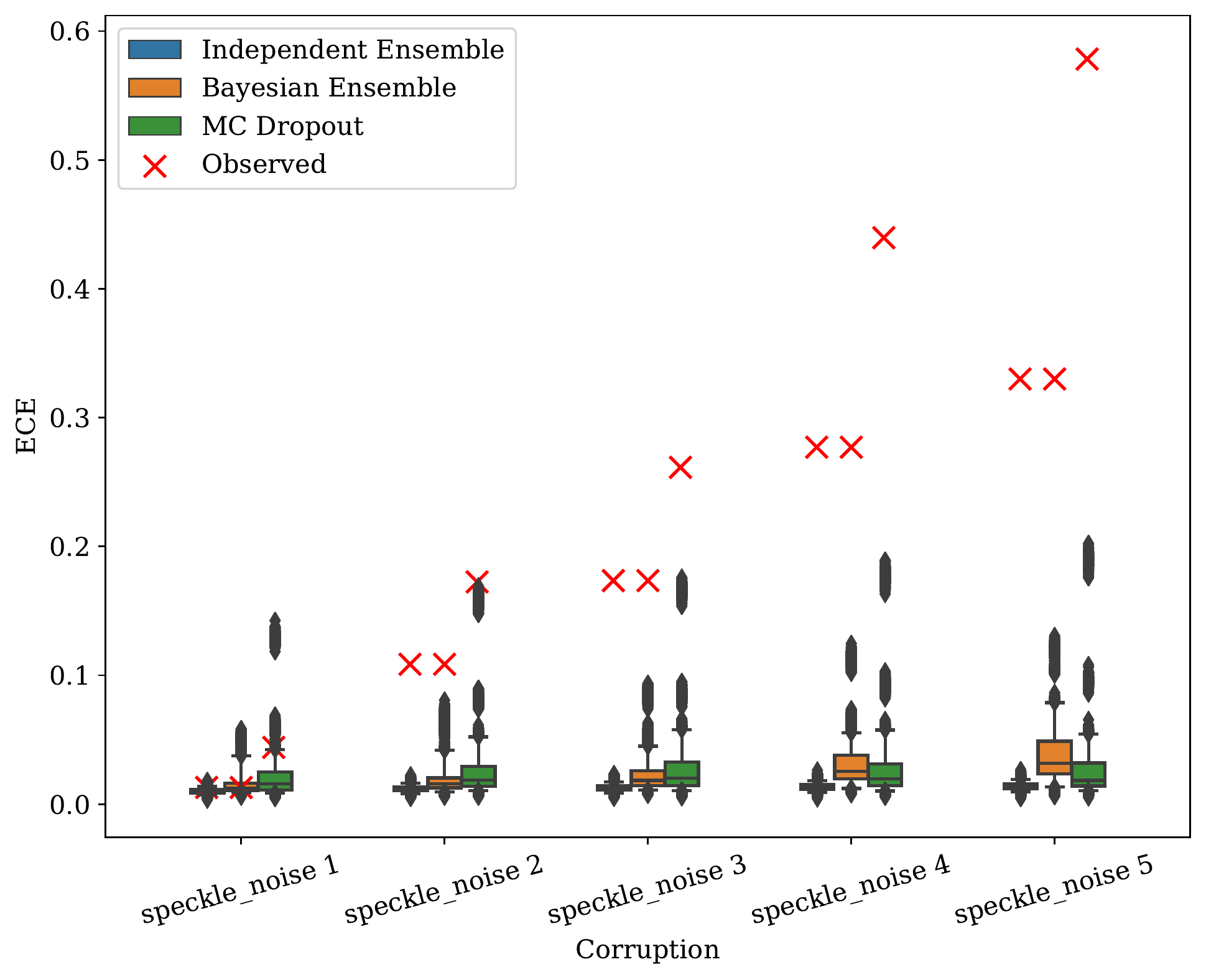}}
\caption{Posterior Check of ECE}
\end{subfigure}
\caption{Posterior predictive checks of recalibrated models trained on CIFAR-10, evaluated on speckle noise data from CIFAR-10-C.}\label{fig:speckle_noise}
\end{minipage}
\end{figure}
    
\begin{figure}
\begin{minipage}{\textwidth}
\begin{subtable}[h]{0.45\textwidth}
\small
\begin{tabular}{lrrrrr}
\toprule
corruption &    0 &    1 &    2 &    3 &    4 \\
model       &      &      &      &      &      \\
\midrule
Dropout     & 0.00 & 0.00 & 0.00 & 0.00 & 0.00 \\
Ensemble    & 0.00 & 0.00 & 0.00 & 0.00 & 0.00 \\
Independent & 0.00 & 0.00 & 0.00 & 0.00 & 0.00 \\
\bottomrule
\end{tabular}

\caption{p-value for Accuracy}
\end{subtable}\hfill
\begin{subtable}[h]{0.45\textwidth}
\small
\begin{tabular}{lrrrrr}
\toprule
corruption &    0 &    1 &    2 &    3 &    4 \\
model       &      &      &      &      &      \\
\midrule
Dropout     & 0.98 & 0.98 & 0.98 & 0.98 & 0.99 \\
Ensemble    & 1.00 & 1.00 & 1.00 & 1.00 & 1.00 \\
Independent & 1.00 & 1.00 & 1.00 & 1.00 & 1.00 \\
\bottomrule
\end{tabular}

\caption{p-value for ECE}
\end{subtable}
\begin{subtable}[h]{\linewidth}
\small
\centering
\begin{tabular}{lrrrrr}
\toprule
corruption &        0 &        1 &        2 &        3 &        4 \\
model       &          &          &          &          &          \\
\midrule
Dropout     & $4.63\times 10^{-2}$ & $4.36\times 10^{-2}$ & $4.74\times 10^{-2}$ & $4.20\times 10^{-2}$ & $4.96\times 10^{-2}$ \\
Ensemble    & $4.50\times 10^{-2}$ & $4.20\times 10^{-2}$ & $3.39\times 10^{-2}$ & $4.24\times 10^{-2}$ & $4.77\times 10^{-2}$ \\
Independent & $6.50\times 10^{-3}$ & $6.50\times 10^{-3}$ & $6.25\times 10^{-3}$ & $6.75\times 10^{-3}$ & $6.62\times 10^{-3}$ \\
\bottomrule
\end{tabular}

\caption{Sharpness for Accuracy}
\end{subtable}

\begin{subtable}[h]{\linewidth}
\small
\centering
\begin{tabular}{lrrrrr}
\toprule
corruption &        0 &        1 &        2 &        3 &        4 \\
model       &          &          &          &          &          \\
\midrule
Dropout     & $2.30\times 10^{-2}$ & $2.34\times 10^{-2}$ & $2.32\times 10^{-2}$ & $2.47\times 10^{-2}$ & $2.36\times 10^{-2}$ \\
Ensemble    & $1.38\times 10^{-2}$ & $1.56\times 10^{-2}$ & $1.28\times 10^{-2}$ & $1.52\times 10^{-2}$ & $1.78\times 10^{-2}$ \\
Independent & $3.94\times 10^{-3}$ & $3.91\times 10^{-3}$ & $3.72\times 10^{-3}$ & $4.08\times 10^{-3}$ & $3.96\times 10^{-3}$ \\
\bottomrule
\end{tabular}

\caption{Sharpness for ECE}
\end{subtable}

\begin{subfigure}{0.45\linewidth}
\centering
\centerline{\includegraphics[width=\columnwidth]{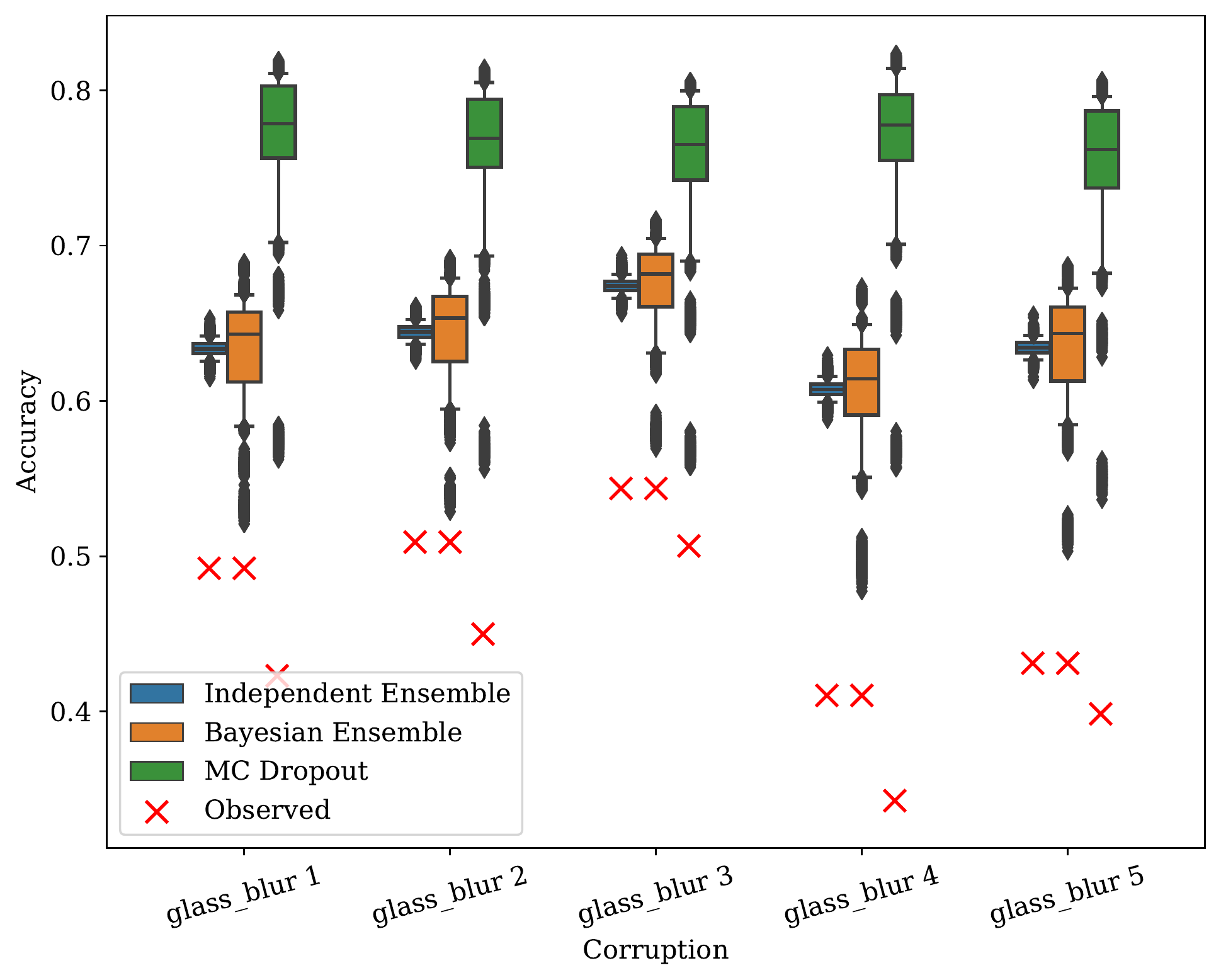}}
\caption{Posterior Check of Accuracy}
\end{subfigure} \hfill
\begin{subfigure}{0.45\linewidth}
\centering
\centerline{\includegraphics[width=\columnwidth]{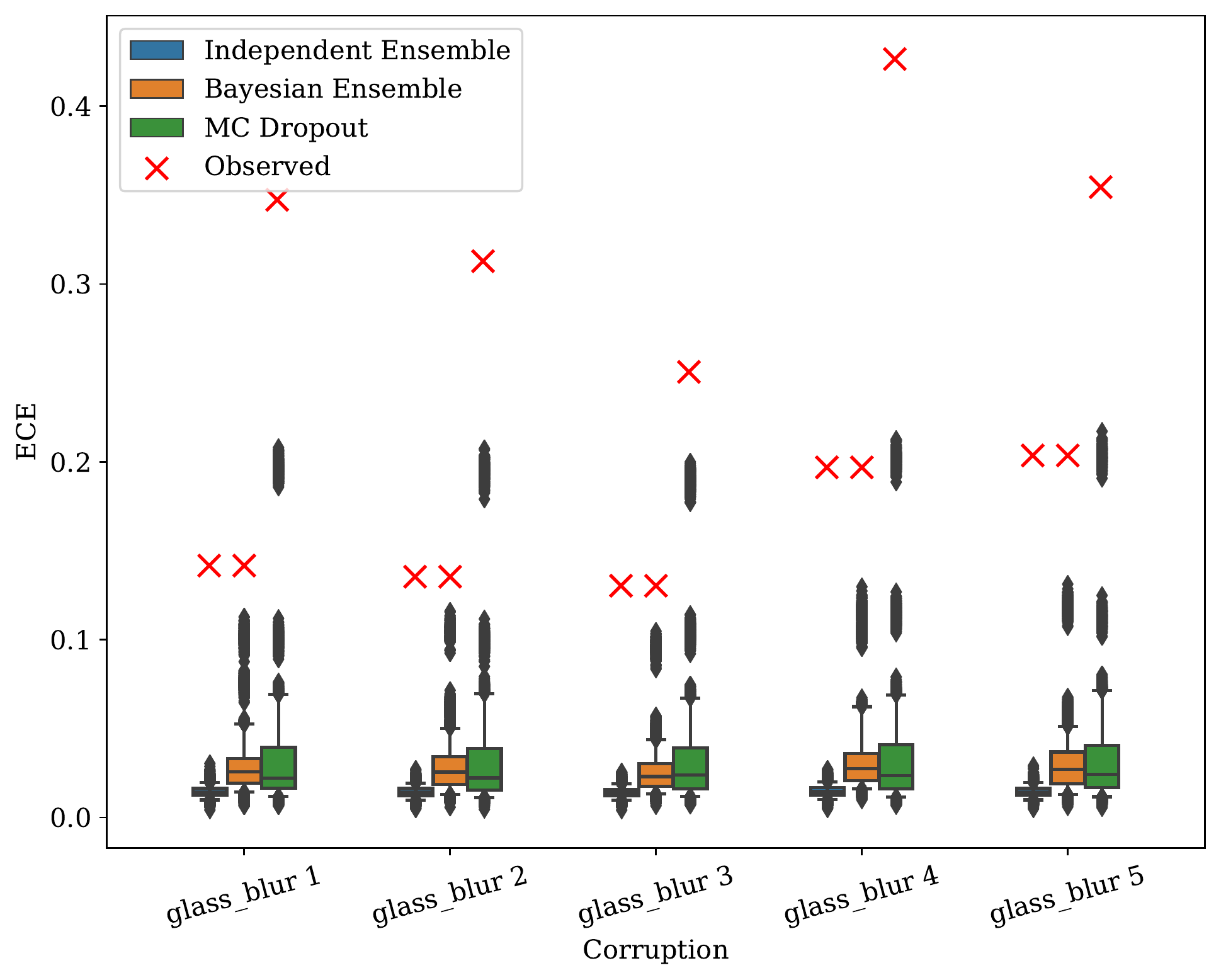}}
\caption{Posterior Check of ECE}
\end{subfigure}
\caption{Posterior predictive checks of recalibrated models trained on CIFAR-10, evaluated on glass blur data from CIFAR-10-C.}\label{fig:glass_blur}
\end{minipage}
\end{figure}
    
\begin{figure}
\begin{minipage}{\textwidth}
\begin{subtable}[h]{0.45\textwidth}
\small
\begin{tabular}{lrrrrr}
\toprule
corruption &    0 &    1 &    2 &    3 &    4 \\
model       &      &      &      &      &      \\
\midrule
Dropout     & 0.72 & 0.17 & 0.04 & 0.06 & 0.02 \\
Ensemble    & 0.84 & 0.16 & 0.02 & 0.03 & 0.00 \\
Independent & 1.00 & 0.00 & 0.00 & 0.00 & 0.00 \\
\bottomrule
\end{tabular}

\caption{p-value for Accuracy}
\end{subtable}\hfill
\begin{subtable}[h]{0.45\textwidth}
\small
\begin{tabular}{lrrrrr}
\toprule
corruption &    0 &    1 &    2 &    3 &    4 \\
model       &      &      &      &      &      \\
\midrule
Dropout     & 0.26 & 0.44 & 0.92 & 0.92 & 0.96 \\
Ensemble    & 0.46 & 0.70 & 0.97 & 0.97 & 1.00 \\
Independent & 0.92 & 1.00 & 1.00 & 1.00 & 1.00 \\
\bottomrule
\end{tabular}

\caption{p-value for ECE}
\end{subtable}
\begin{subtable}[h]{\linewidth}
\small
\centering
\begin{tabular}{lrrrrr}
\toprule
corruption &        0 &        1 &        2 &        3 &        4 \\
model       &          &          &          &          &          \\
\midrule
Dropout     & $3.00\times 10^{-2}$ & $3.15\times 10^{-2}$ & $4.54\times 10^{-2}$ & $4.53\times 10^{-2}$ & $5.47\times 10^{-2}$ \\
Ensemble    & $1.06\times 10^{-2}$ & $1.56\times 10^{-2}$ & $2.14\times 10^{-2}$ & $2.11\times 10^{-2}$ & $2.58\times 10^{-2}$ \\
Independent & $4.13\times 10^{-3}$ & $4.87\times 10^{-3}$ & $5.50\times 10^{-3}$ & $5.25\times 10^{-3}$ & $5.88\times 10^{-3}$ \\
\bottomrule
\end{tabular}

\caption{Sharpness for Accuracy}
\end{subtable}

\begin{subtable}[h]{\linewidth}
\small
\centering
\begin{tabular}{lrrrrr}
\toprule
corruption &        0 &        1 &        2 &        3 &        4 \\
model       &          &          &          &          &          \\
\midrule
Dropout     & $1.56\times 10^{-2}$ & $2.10\times 10^{-2}$ & $2.41\times 10^{-2}$ & $2.44\times 10^{-2}$ & $2.84\times 10^{-2}$ \\
Ensemble    & $4.75\times 10^{-3}$ & $7.73\times 10^{-3}$ & $1.09\times 10^{-2}$ & $1.09\times 10^{-2}$ & $1.23\times 10^{-2}$ \\
Independent & $2.54\times 10^{-3}$ & $2.92\times 10^{-3}$ & $3.32\times 10^{-3}$ & $3.34\times 10^{-3}$ & $3.57\times 10^{-3}$ \\
\bottomrule
\end{tabular}

\caption{Sharpness for ECE}
\end{subtable}

\begin{subfigure}{0.45\linewidth}
\centering
\centerline{\includegraphics[width=\columnwidth]{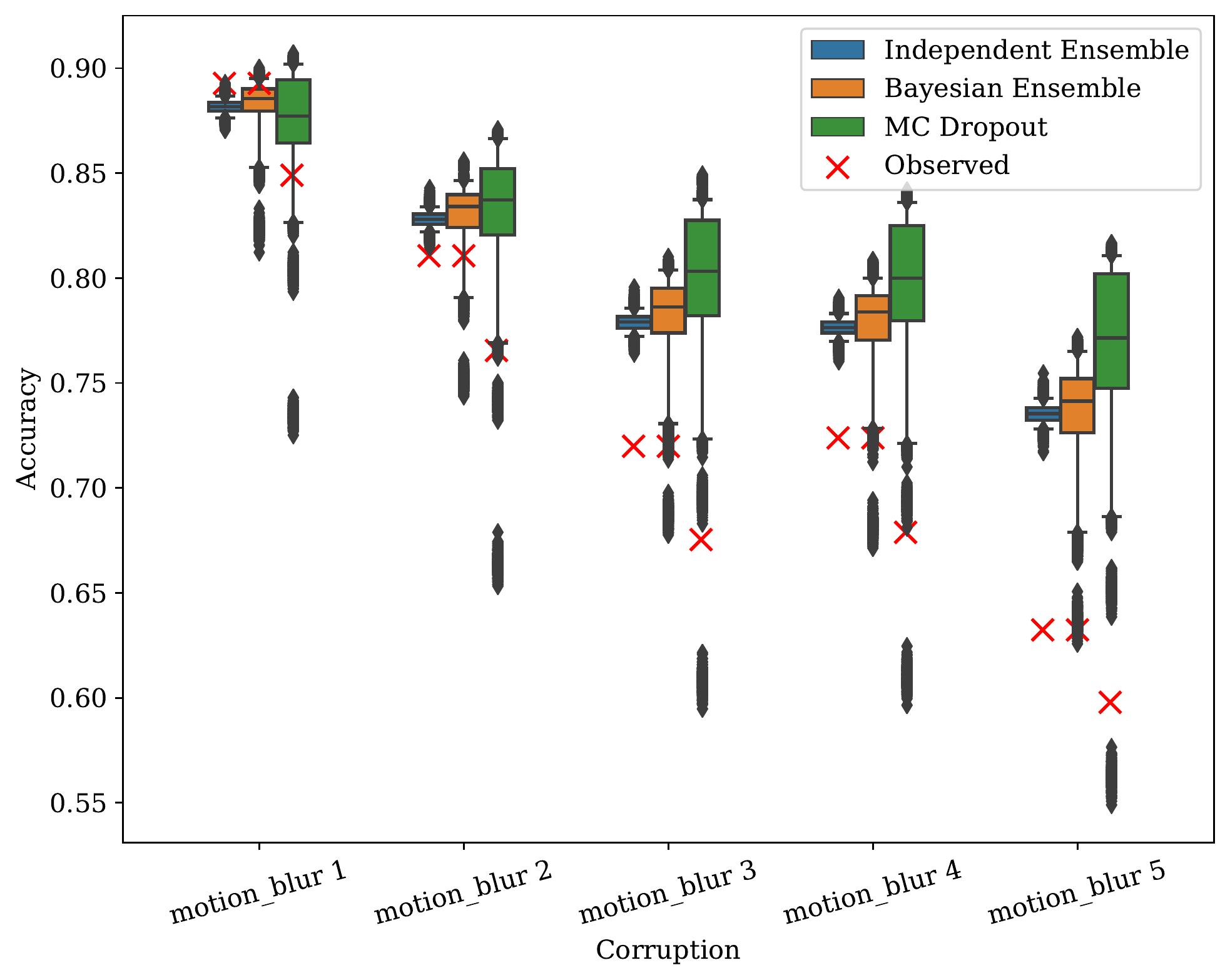}}
\caption{Posterior Check of Accuracy}
\end{subfigure} \hfill
\begin{subfigure}{0.45\linewidth}
\centering
\centerline{\includegraphics[width=\columnwidth]{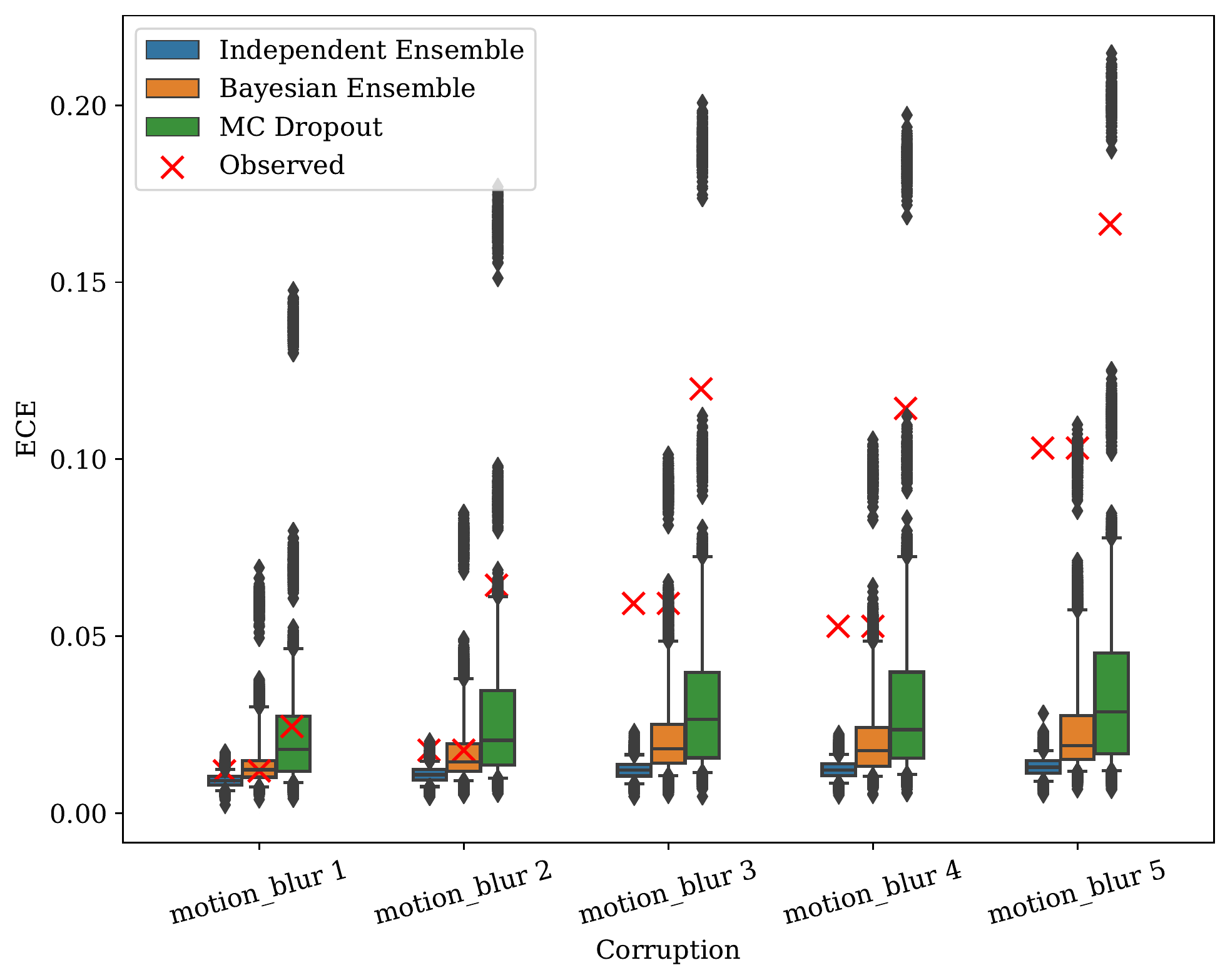}}
\caption{Posterior Check of ECE}
\end{subfigure}
\caption{Posterior predictive checks of recalibrated models trained on CIFAR-10, evaluated on motion blur data from CIFAR-10-C.}\label{fig:motion_blur}
\end{minipage}
\end{figure}
    
\begin{figure}
\begin{minipage}{\textwidth}
\begin{subtable}[h]{0.45\textwidth}
\small
\begin{tabular}{lrrrrr}
\toprule
corruption &    0 &    1 &    2 &    3 &    4 \\
model       &      &      &      &      &      \\
\midrule
Dropout     & 0.47 & 0.12 & 0.03 & 0.02 & 0.00 \\
Ensemble    & 0.63 & 0.17 & 0.06 & 0.02 & 0.00 \\
Independent & 0.98 & 0.00 & 0.00 & 0.00 & 0.00 \\
\bottomrule
\end{tabular}

\caption{p-value for Accuracy}
\end{subtable}\hfill
\begin{subtable}[h]{0.45\textwidth}
\small
\begin{tabular}{lrrrrr}
\toprule
corruption &    0 &    1 &    2 &    3 &    4 \\
model       &      &      &      &      &      \\
\midrule
Dropout     & 0.28 & 0.34 & 0.92 & 0.96 & 0.98 \\
Ensemble    & 0.50 & 0.50 & 0.94 & 0.98 & 1.00 \\
Independent & 0.89 & 0.94 & 1.00 & 1.00 & 1.00 \\
\bottomrule
\end{tabular}

\caption{p-value for ECE}
\end{subtable}
\begin{subtable}[h]{\linewidth}
\small
\centering
\begin{tabular}{lrrrrr}
\toprule
corruption &        0 &        1 &        2 &        3 &        4 \\
model       &          &          &          &          &          \\
\midrule
Dropout     & $2.31\times 10^{-2}$ & $2.91\times 10^{-2}$ & $3.75\times 10^{-2}$ & $3.96\times 10^{-2}$ & $4.14\times 10^{-2}$ \\
Ensemble    & $8.75\times 10^{-3}$ & $1.35\times 10^{-2}$ & $1.70\times 10^{-2}$ & $2.04\times 10^{-2}$ & $3.37\times 10^{-2}$ \\
Independent & $4.00\times 10^{-3}$ & $4.62\times 10^{-3}$ & $5.25\times 10^{-3}$ & $5.50\times 10^{-3}$ & $6.13\times 10^{-3}$ \\
\bottomrule
\end{tabular}

\caption{Sharpness for Accuracy}
\end{subtable}

\begin{subtable}[h]{\linewidth}
\small
\centering
\begin{tabular}{lrrrrr}
\toprule
corruption &        0 &        1 &        2 &        3 &        4 \\
model       &          &          &          &          &          \\
\midrule
Dropout     & $1.09\times 10^{-2}$ & $1.38\times 10^{-2}$ & $1.92\times 10^{-2}$ & $2.00\times 10^{-2}$ & $2.17\times 10^{-2}$ \\
Ensemble    & $4.52\times 10^{-3}$ & $4.46\times 10^{-3}$ & $7.91\times 10^{-3}$ & $9.53\times 10^{-3}$ & $1.35\times 10^{-2}$ \\
Independent & $2.42\times 10^{-3}$ & $2.77\times 10^{-3}$ & $3.26\times 10^{-3}$ & $3.30\times 10^{-3}$ & $3.69\times 10^{-3}$ \\
\bottomrule
\end{tabular}

\caption{Sharpness for ECE}
\end{subtable}

\begin{subfigure}{0.45\linewidth}
\centering
\centerline{\includegraphics[width=\columnwidth]{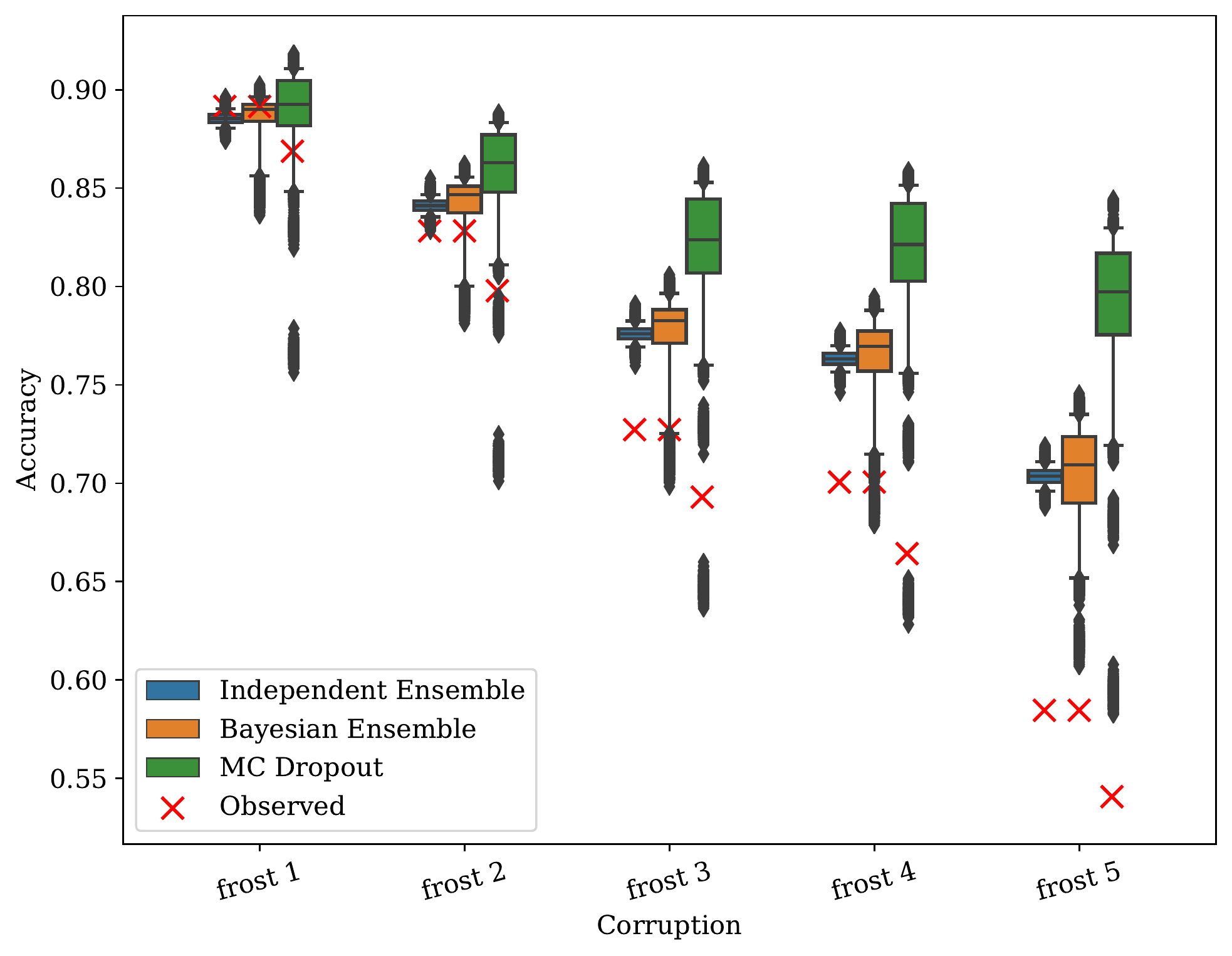}}
\caption{Posterior Check of Accuracy}
\end{subfigure} \hfill
\begin{subfigure}{0.45\linewidth}
\centering
\centerline{\includegraphics[width=\columnwidth]{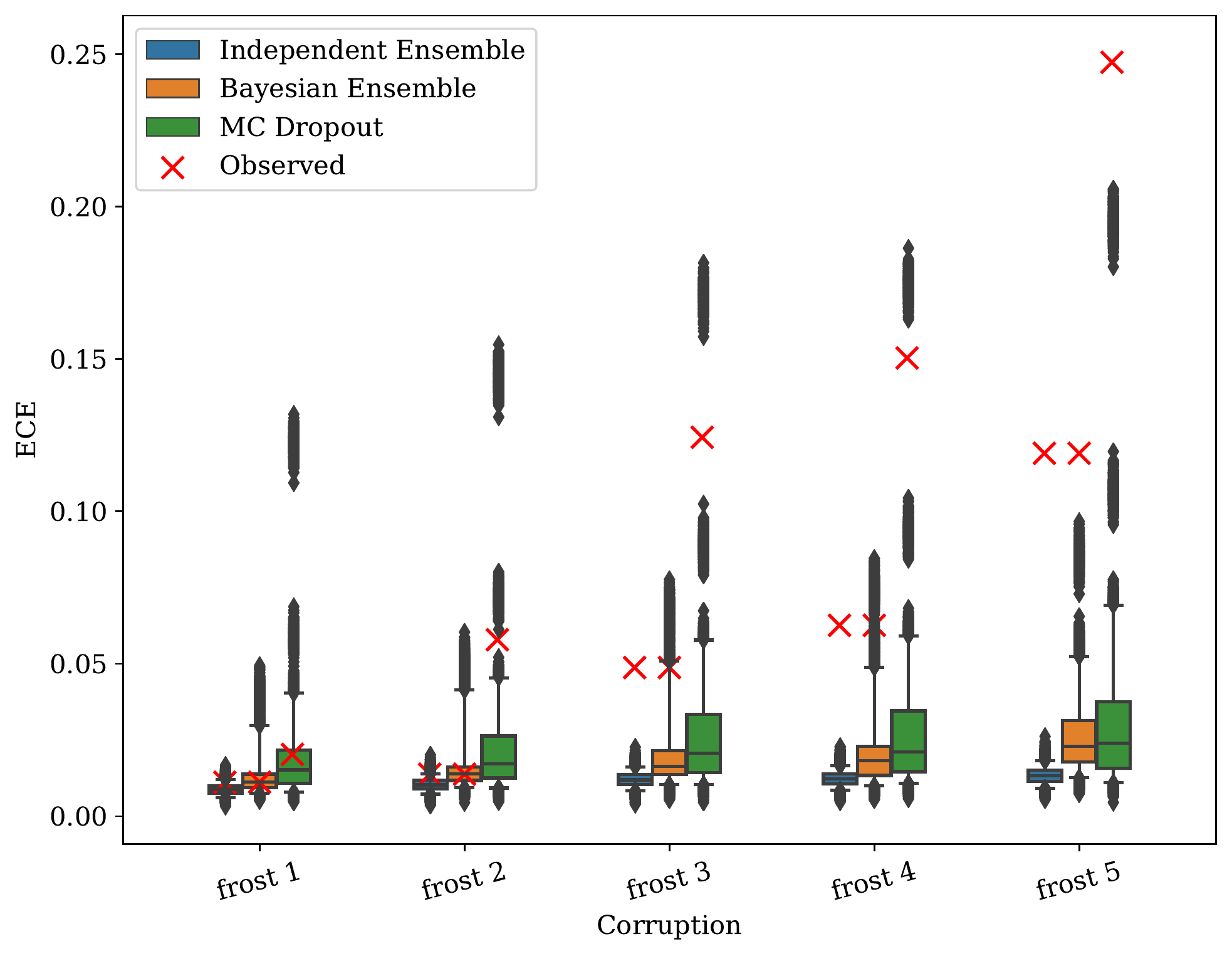}}
\caption{Posterior Check of ECE}
\end{subfigure}
\caption{Posterior predictive checks of recalibrated models trained on CIFAR-10, evaluated on frost data from CIFAR-10-C.}\label{fig:frost}
\end{minipage}
\end{figure}
    
\begin{figure}
\begin{minipage}{\textwidth}
\begin{subtable}[h]{0.45\textwidth}
\small
\begin{tabular}{lrrrrr}
\toprule
corruption &    0 &    1 &    2 &    3 &    4 \\
model       &      &      &      &      &      \\
\midrule
Dropout     & 0.89 & 0.48 & 0.09 & 0.02 & 0.00 \\
Ensemble    & 0.92 & 0.31 & 0.10 & 0.00 & 0.00 \\
Independent & 1.00 & 0.27 & 0.00 & 0.00 & 0.00 \\
\bottomrule
\end{tabular}

\caption{p-value for Accuracy}
\end{subtable}\hfill
\begin{subtable}[h]{0.45\textwidth}
\small
\begin{tabular}{lrrrrr}
\toprule
corruption &    0 &    1 &    2 &    3 &    4 \\
model       &      &      &      &      &      \\
\midrule
Dropout     & 0.18 & 0.17 & 0.80 & 0.96 & 1.00 \\
Ensemble    & 0.52 & 0.26 & 0.89 & 1.00 & 1.00 \\
Independent & 0.81 & 0.78 & 1.00 & 1.00 & 1.00 \\
\bottomrule
\end{tabular}

\caption{p-value for ECE}
\end{subtable}
\begin{subtable}[h]{\linewidth}
\small
\centering
\begin{tabular}{lrrrrr}
\toprule
corruption &        0 &        1 &        2 &        3 &        4 \\
model       &          &          &          &          &          \\
\midrule
Dropout     & $1.84\times 10^{-2}$ & $2.89\times 10^{-2}$ & $3.76\times 10^{-2}$ & $4.70\times 10^{-2}$ & $5.83\times 10^{-2}$ \\
Ensemble    & $5.38\times 10^{-3}$ & $1.74\times 10^{-2}$ & $2.75\times 10^{-2}$ & $4.50\times 10^{-2}$ & $7.09\times 10^{-2}$ \\
Independent & $3.25\times 10^{-3}$ & $4.25\times 10^{-3}$ & $5.00\times 10^{-3}$ & $5.75\times 10^{-3}$ & $6.50\times 10^{-3}$ \\
\bottomrule
\end{tabular}

\caption{Sharpness for Accuracy}
\end{subtable}

\begin{subtable}[h]{\linewidth}
\small
\centering
\begin{tabular}{lrrrrr}
\toprule
corruption &        0 &        1 &        2 &        3 &        4 \\
model       &          &          &          &          &          \\
\midrule
Dropout     & $9.72\times 10^{-3}$ & $1.61\times 10^{-2}$ & $2.10\times 10^{-2}$ & $2.72\times 10^{-2}$ & $3.72\times 10^{-2}$ \\
Ensemble    & $3.10\times 10^{-3}$ & $6.14\times 10^{-3}$ & $1.36\times 10^{-2}$ & $1.78\times 10^{-2}$ & $3.04\times 10^{-2}$ \\
Independent & $2.00\times 10^{-3}$ & $2.59\times 10^{-3}$ & $3.11\times 10^{-3}$ & $3.51\times 10^{-3}$ & $3.84\times 10^{-3}$ \\
\bottomrule
\end{tabular}

\caption{Sharpness for ECE}
\end{subtable}

\begin{subfigure}{0.45\linewidth}
\centering
\centerline{\includegraphics[width=\columnwidth]{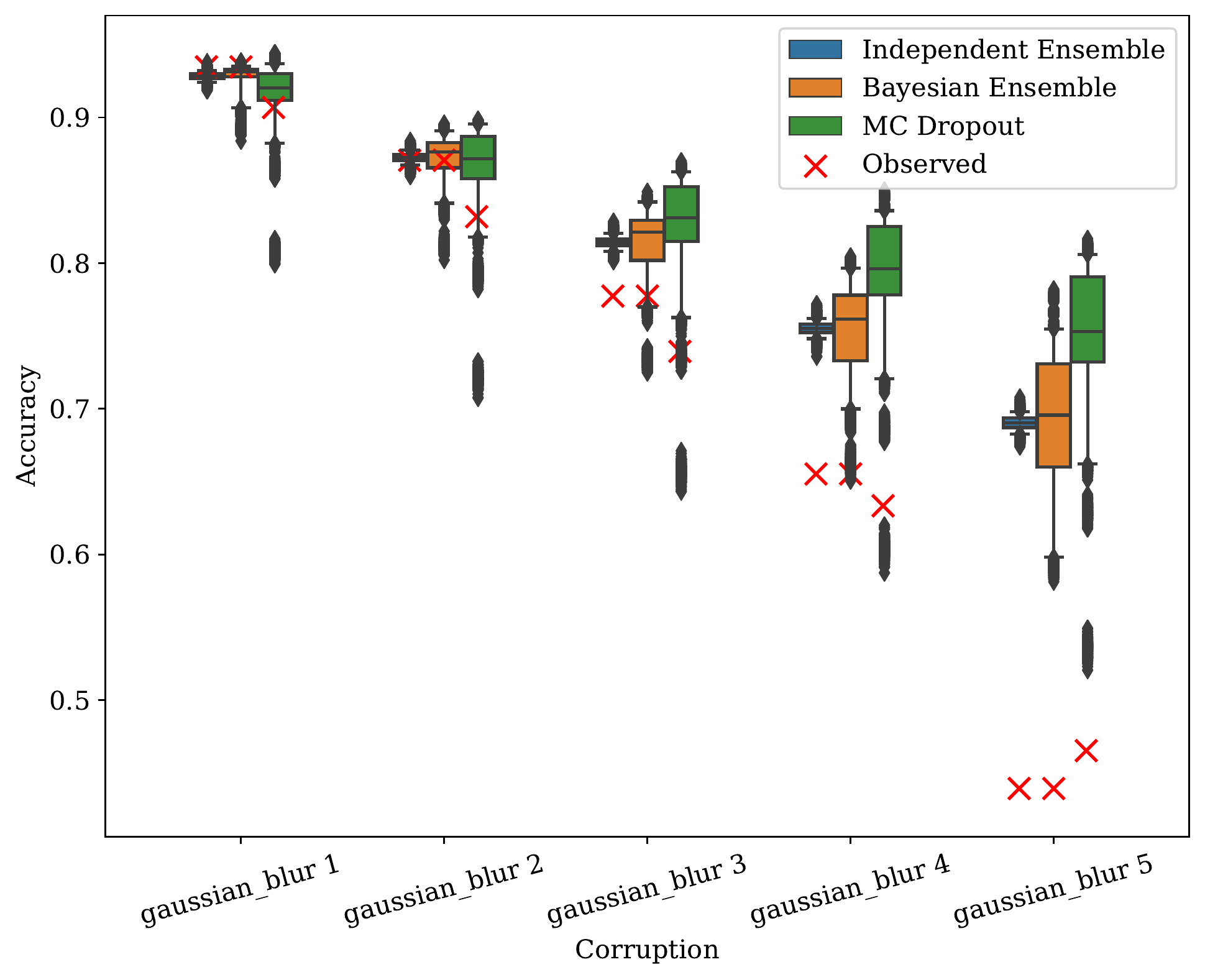}}
\caption{Posterior Check of Accuracy}
\end{subfigure} \hfill
\begin{subfigure}{0.45\linewidth}
\centering
\centerline{\includegraphics[width=\columnwidth]{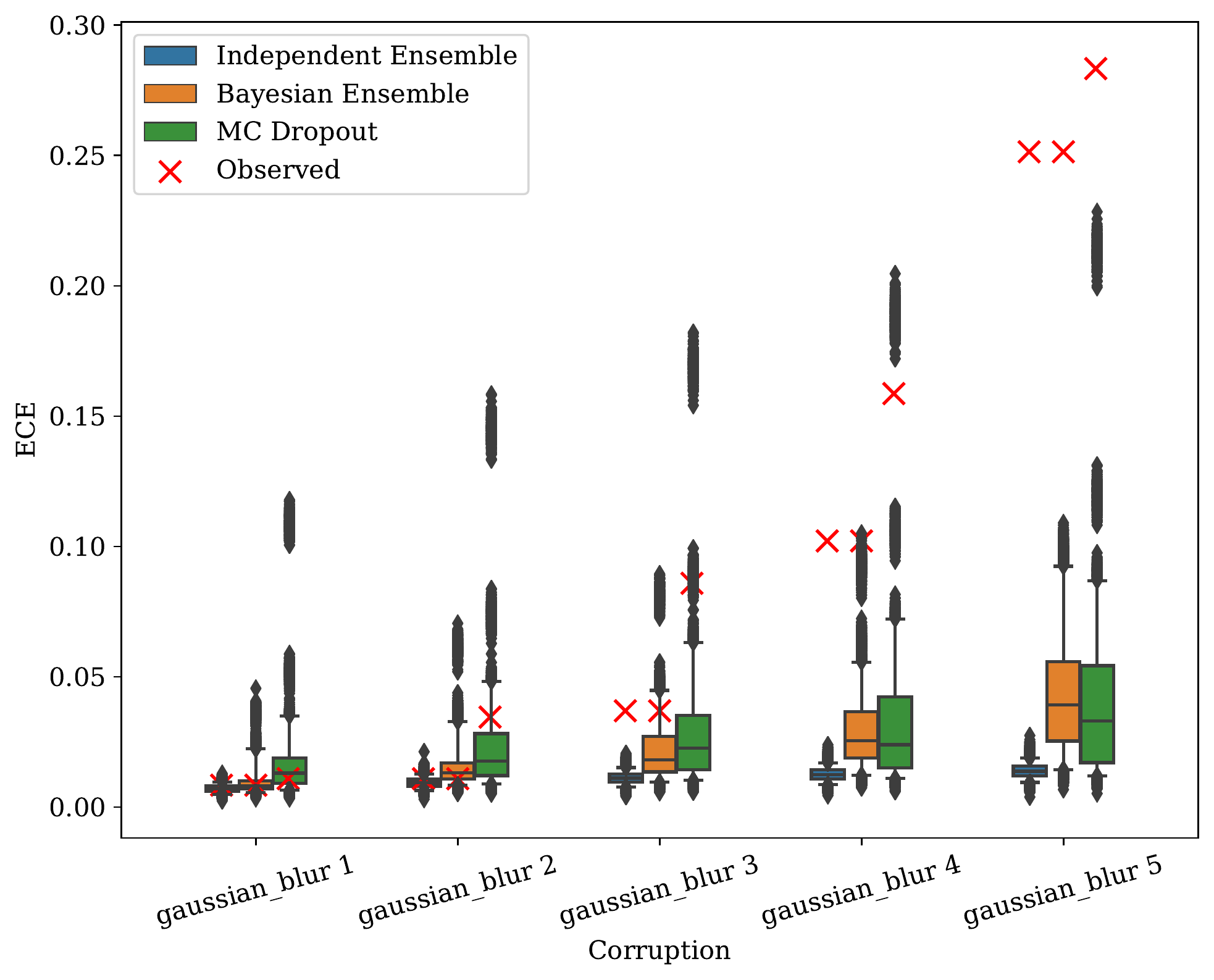}}
\caption{Posterior Check of ECE}
\end{subfigure}
\caption{Posterior predictive checks of recalibrated models trained on CIFAR-10, evaluated on gaussian blur data from CIFAR-10-C.}\label{fig:gaussian_blur}
\end{minipage}
\end{figure}
    
\begin{figure}
\begin{minipage}{\textwidth}
\begin{subtable}[h]{0.45\textwidth}
\small
\begin{tabular}{lrrrrr}
\toprule
corruption &    0 &    1 &    2 &    3 &    4 \\
model       &      &      &      &      &      \\
\midrule
Dropout     & 0.89 & 1.00 & 0.92 & 0.64 & 0.04 \\
Ensemble    & 0.79 & 1.00 & 0.70 & 0.22 & 0.02 \\
Independent & 0.98 & 1.00 & 1.00 & 0.18 & 0.00 \\
\bottomrule
\end{tabular}

\caption{p-value for Accuracy}
\end{subtable}\hfill
\begin{subtable}[h]{0.45\textwidth}
\small
\begin{tabular}{lrrrrr}
\toprule
corruption &    0 &    1 &    2 &    3 &    4 \\
model       &      &      &      &      &      \\
\midrule
Dropout     & 0.21 & 0.38 & 0.06 & 0.01 & 0.96 \\
Ensemble    & 0.61 & 0.78 & 0.20 & 0.06 & 0.98 \\
Independent & 0.90 & 1.00 & 0.66 & 0.23 & 1.00 \\
\bottomrule
\end{tabular}

\caption{p-value for ECE}
\end{subtable}
\begin{subtable}[h]{\linewidth}
\small
\centering
\begin{tabular}{lrrrrr}
\toprule
corruption &        0 &        1 &        2 &        3 &        4 \\
model       &          &          &          &          &          \\
\midrule
Dropout     & $1.90\times 10^{-2}$ & $2.31\times 10^{-2}$ & $2.87\times 10^{-2}$ & $3.71\times 10^{-2}$ & $5.43\times 10^{-2}$ \\
Ensemble    & $5.12\times 10^{-3}$ & $7.25\times 10^{-3}$ & $1.06\times 10^{-2}$ & $1.12\times 10^{-2}$ & $1.99\times 10^{-2}$ \\
Independent & $3.25\times 10^{-3}$ & $3.50\times 10^{-3}$ & $3.87\times 10^{-3}$ & $4.38\times 10^{-3}$ & $5.37\times 10^{-3}$ \\
\bottomrule
\end{tabular}

\caption{Sharpness for Accuracy}
\end{subtable}

\begin{subtable}[h]{\linewidth}
\small
\centering
\begin{tabular}{lrrrrr}
\toprule
corruption &        0 &        1 &        2 &        3 &        4 \\
model       &          &          &          &          &          \\
\midrule
Dropout     & $9.53\times 10^{-3}$ & $1.20\times 10^{-2}$ & $1.41\times 10^{-2}$ & $1.77\times 10^{-2}$ & $3.06\times 10^{-2}$ \\
Ensemble    & $3.10\times 10^{-3}$ & $3.72\times 10^{-3}$ & $3.81\times 10^{-3}$ & $5.11\times 10^{-3}$ & $9.63\times 10^{-3}$ \\
Independent & $1.99\times 10^{-3}$ & $2.10\times 10^{-3}$ & $2.32\times 10^{-3}$ & $2.61\times 10^{-3}$ & $3.37\times 10^{-3}$ \\
\bottomrule
\end{tabular}

\caption{Sharpness for ECE}
\end{subtable}

\begin{subfigure}{0.45\linewidth}
\centering
\centerline{\includegraphics[width=\columnwidth]{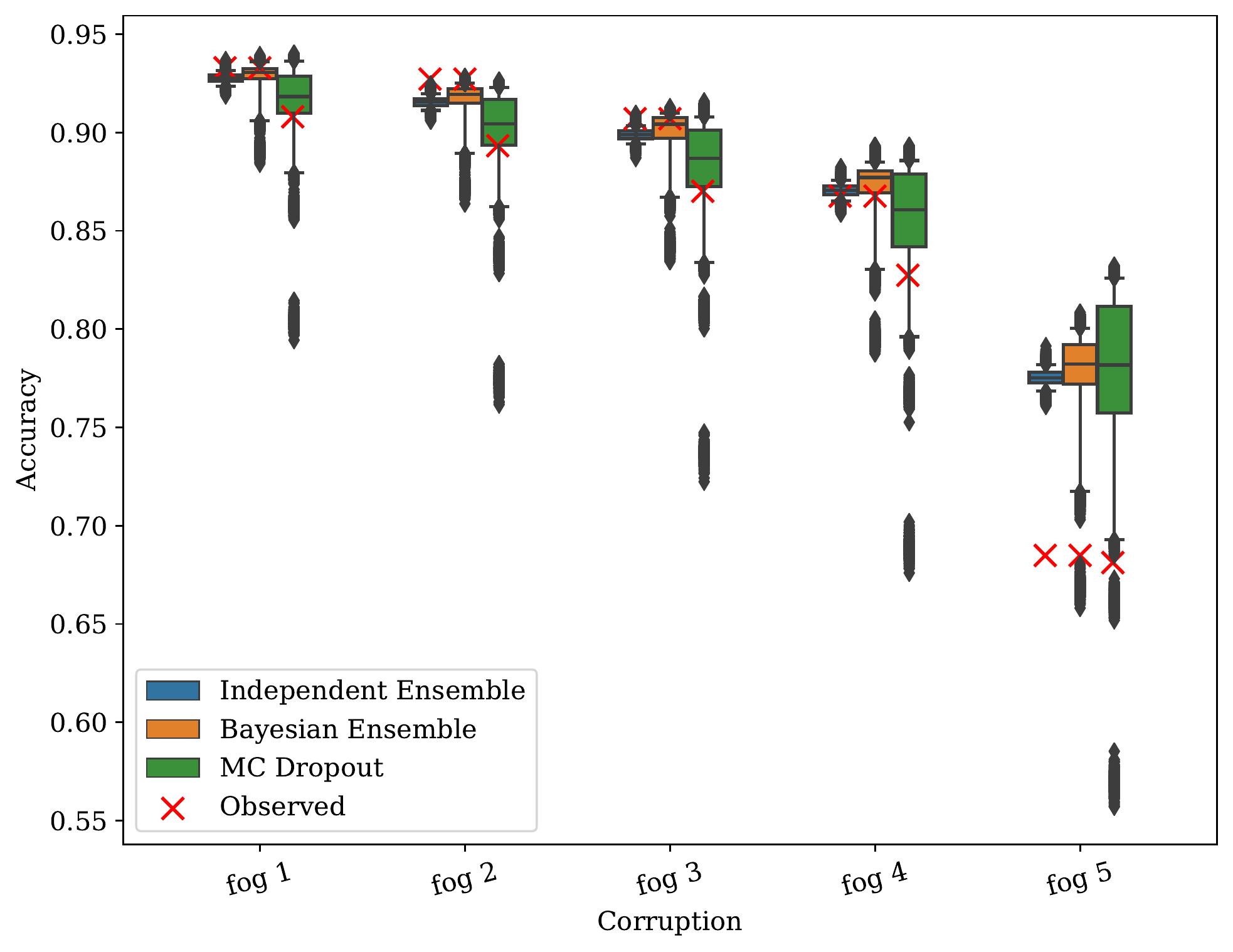}}
\caption{Posterior Check of Accuracy}
\end{subfigure} \hfill
\begin{subfigure}{0.45\linewidth}
\centering
\centerline{\includegraphics[width=\columnwidth]{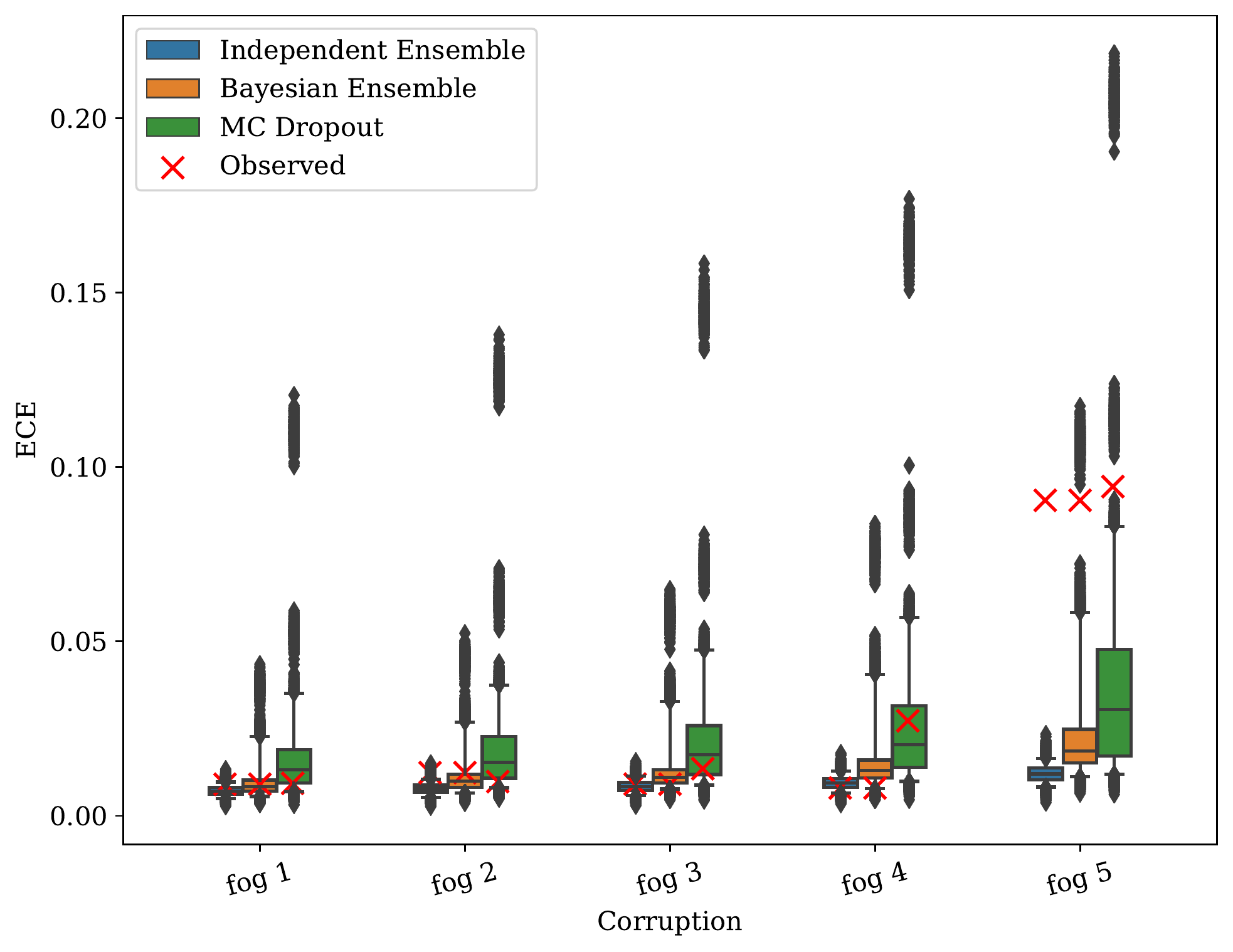}}
\caption{Posterior Check of ECE}
\end{subfigure}
\caption{Posterior predictive checks of recalibrated models trained on CIFAR-10, evaluated on fog data from CIFAR-10-C.}\label{fig:fog}
\end{minipage}
\end{figure}
    
\begin{figure}
\begin{minipage}{\textwidth}
\begin{subtable}[h]{0.45\textwidth}
\small
\begin{tabular}{lrrrrr}
\toprule
corruption &    0 &    1 &    2 &    3 &    4 \\
model       &      &      &      &      &      \\
\midrule
Dropout     & 0.02 & 0.00 & 0.00 & 0.00 & 0.00 \\
Ensemble    & 0.02 & 0.00 & 0.00 & 0.00 & 0.00 \\
Independent & 0.00 & 0.00 & 0.00 & 0.00 & 0.00 \\
\bottomrule
\end{tabular}

\caption{p-value for Accuracy}
\end{subtable}\hfill
\begin{subtable}[h]{0.45\textwidth}
\small
\begin{tabular}{lrrrrr}
\toprule
corruption &    0 &    1 &    2 &    3 &    4 \\
model       &      &      &      &      &      \\
\midrule
Dropout     & 0.96 & 1.00 & 1.00 & 1.00 & 1.00 \\
Ensemble    & 0.98 & 1.00 & 1.00 & 1.00 & 1.00 \\
Independent & 1.00 & 1.00 & 1.00 & 1.00 & 1.00 \\
\bottomrule
\end{tabular}

\caption{p-value for ECE}
\end{subtable}
\begin{subtable}[h]{\linewidth}
\small
\centering
\begin{tabular}{lrrrrr}
\toprule
corruption &        0 &        1 &        2 &        3 &        4 \\
model       &          &          &          &          &          \\
\midrule
Dropout     & $2.78\times 10^{-2}$ & $3.61\times 10^{-2}$ & $3.01\times 10^{-2}$ & $2.45\times 10^{-2}$ & $2.20\times 10^{-2}$ \\
Ensemble    & $1.70\times 10^{-2}$ & $3.16\times 10^{-2}$ & $6.54\times 10^{-2}$ & $8.10\times 10^{-2}$ & $1.04\times 10^{-1}$ \\
Independent & $5.25\times 10^{-3}$ & $5.75\times 10^{-3}$ & $6.13\times 10^{-3}$ & $6.00\times 10^{-3}$ & $6.13\times 10^{-3}$ \\
\bottomrule
\end{tabular}

\caption{Sharpness for Accuracy}
\end{subtable}

\begin{subtable}[h]{\linewidth}
\small
\centering
\begin{tabular}{lrrrrr}
\toprule
corruption &        0 &        1 &        2 &        3 &        4 \\
model       &          &          &          &          &          \\
\midrule
Dropout     & $1.59\times 10^{-2}$ & $1.75\times 10^{-2}$ & $1.83\times 10^{-2}$ & $1.46\times 10^{-2}$ & $1.36\times 10^{-2}$ \\
Ensemble    & $8.57\times 10^{-3}$ & $1.66\times 10^{-2}$ & $2.66\times 10^{-2}$ & $3.37\times 10^{-2}$ & $3.99\times 10^{-2}$ \\
Independent & $3.13\times 10^{-3}$ & $3.54\times 10^{-3}$ & $3.64\times 10^{-3}$ & $3.70\times 10^{-3}$ & $3.76\times 10^{-3}$ \\
\bottomrule
\end{tabular}

\caption{Sharpness for ECE}
\end{subtable}

\begin{subfigure}{0.45\linewidth}
\centering
\centerline{\includegraphics[width=\columnwidth]{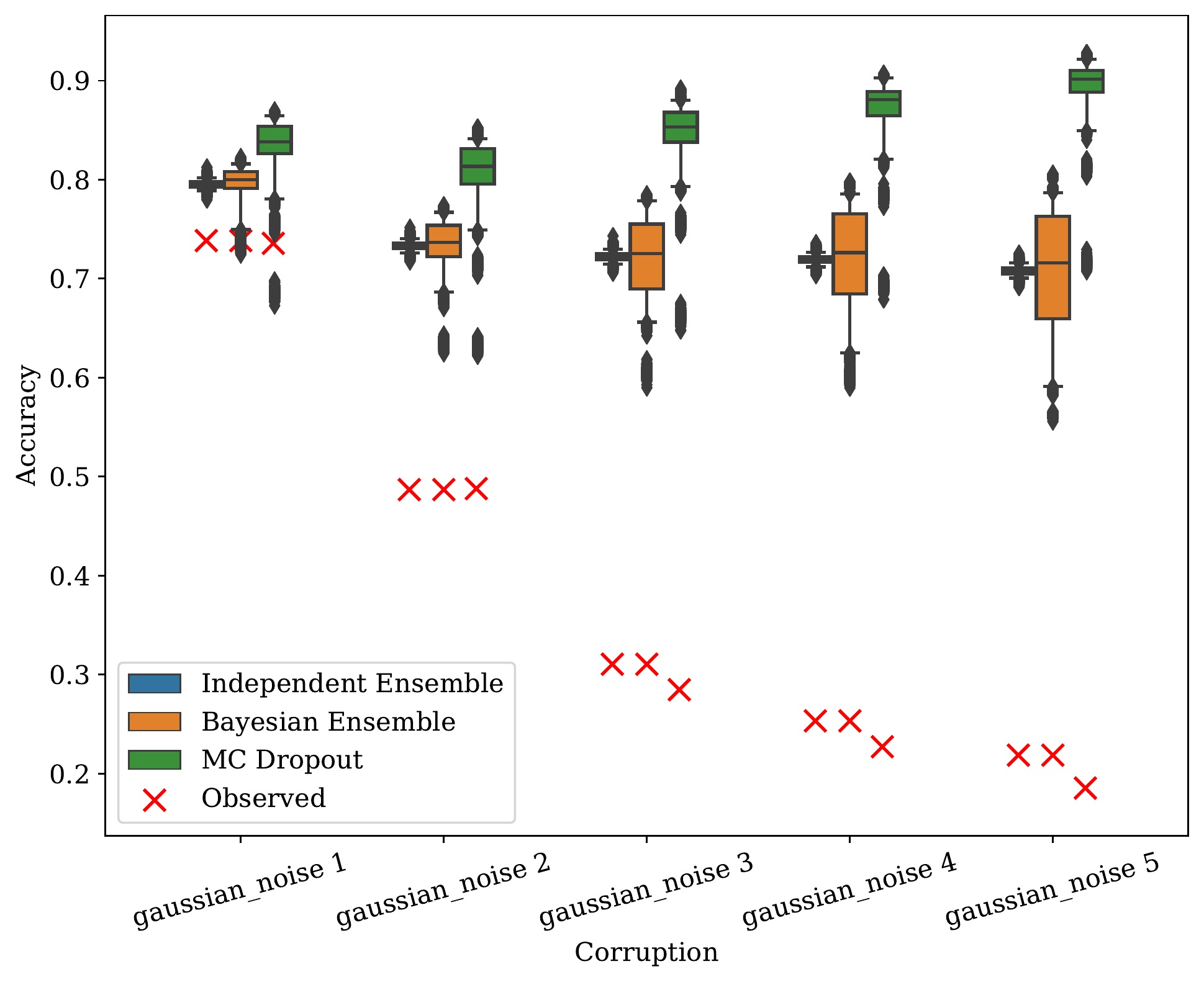}}
\caption{Posterior Check of Accuracy}
\end{subfigure} \hfill
\begin{subfigure}{0.45\linewidth}
\centering
\centerline{\includegraphics[width=\columnwidth]{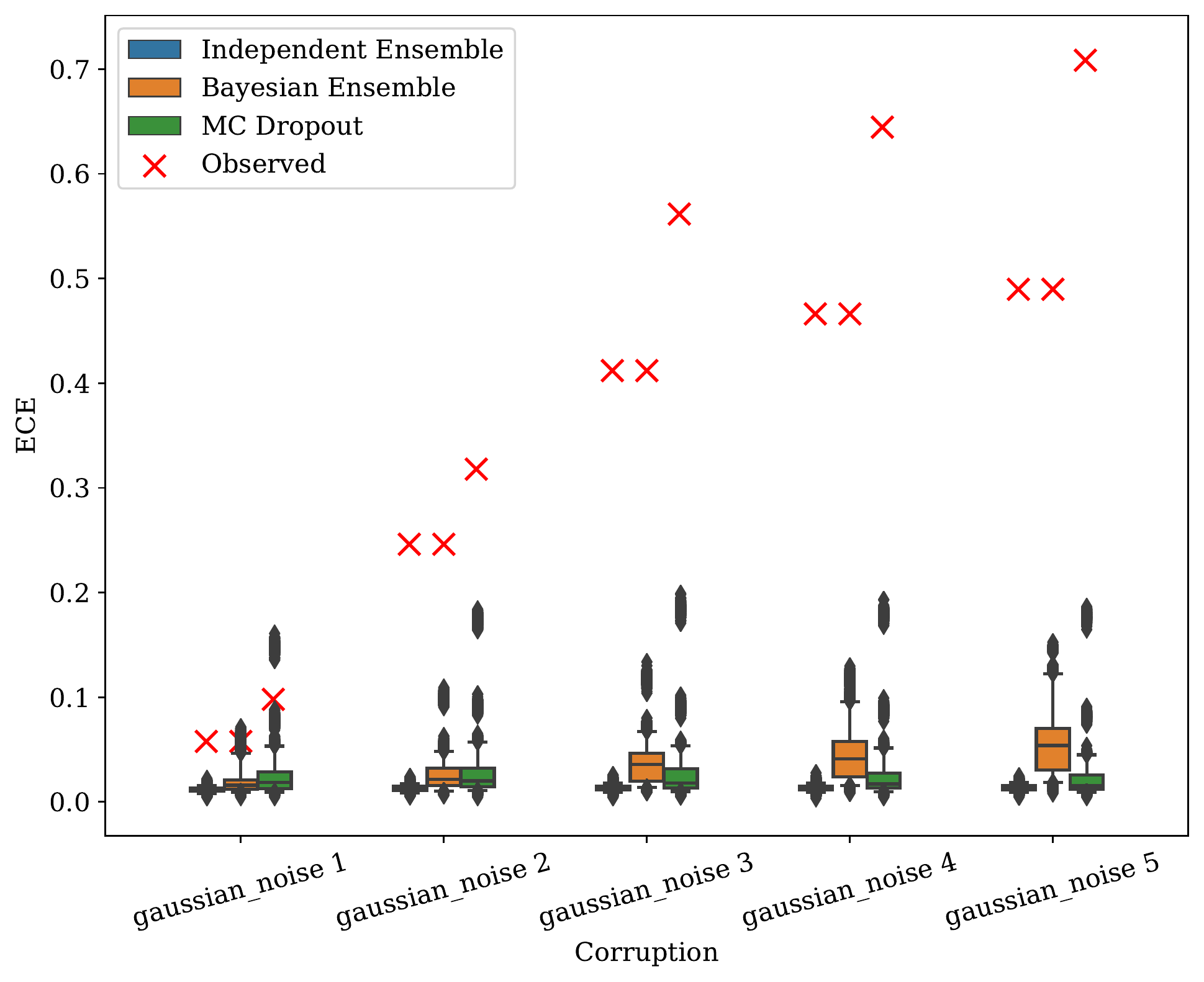}}
\caption{Posterior Check of ECE}
\end{subfigure}
\caption{Posterior predictive checks of recalibrated models trained on CIFAR-10, evaluated on gaussian noise data from CIFAR-10-C.}\label{fig:gaussian_noise}
\end{minipage}
\end{figure}
    
\begin{figure}
\begin{minipage}{\textwidth}
\begin{subtable}[h]{0.45\textwidth}
\small
\begin{tabular}{lrrrrr}
\toprule
corruption &    0 &    1 &    2 &    3 &    4 \\
model       &      &      &      &      &      \\
\midrule
Dropout     & 0.86 & 0.83 & 0.83 & 0.80 & 0.68 \\
Ensemble    & 0.89 & 0.75 & 0.96 & 0.96 & 0.70 \\
Independent & 0.99 & 0.97 & 1.00 & 1.00 & 0.98 \\
\bottomrule
\end{tabular}

\caption{p-value for Accuracy}
\end{subtable}\hfill
\begin{subtable}[h]{0.45\textwidth}
\small
\begin{tabular}{lrrrrr}
\toprule
corruption &    0 &    1 &    2 &    3 &    4 \\
model       &      &      &      &      &      \\
\midrule
Dropout     & 0.27 & 0.27 & 0.18 & 0.10 & 0.26 \\
Ensemble    & 0.71 & 0.72 & 0.57 & 0.33 & 0.59 \\
Independent & 0.94 & 0.93 & 0.85 & 0.65 & 0.93 \\
\bottomrule
\end{tabular}

\caption{p-value for ECE}
\end{subtable}
\begin{subtable}[h]{\linewidth}
\small
\centering
\begin{tabular}{lrrrrr}
\toprule
corruption &        0 &        1 &        2 &        3 &        4 \\
model       &          &          &          &          &          \\
\midrule
Dropout     & $1.81\times 10^{-2}$ & $1.71\times 10^{-2}$ & $2.07\times 10^{-2}$ & $2.24\times 10^{-2}$ & $2.29\times 10^{-2}$ \\
Ensemble    & $5.12\times 10^{-3}$ & $4.87\times 10^{-3}$ & $4.25\times 10^{-3}$ & $5.62\times 10^{-3}$ & $8.25\times 10^{-3}$ \\
Independent & $3.25\times 10^{-3}$ & $3.37\times 10^{-3}$ & $3.37\times 10^{-3}$ & $3.37\times 10^{-3}$ & $4.00\times 10^{-3}$ \\
\bottomrule
\end{tabular}

\caption{Sharpness for Accuracy}
\end{subtable}

\begin{subtable}[h]{\linewidth}
\small
\centering
\begin{tabular}{lrrrrr}
\toprule
corruption &        0 &        1 &        2 &        3 &        4 \\
model       &          &          &          &          &          \\
\midrule
Dropout     & $9.04\times 10^{-3}$ & $8.74\times 10^{-3}$ & $1.00\times 10^{-2}$ & $1.02\times 10^{-2}$ & $1.07\times 10^{-2}$ \\
Ensemble    & $2.63\times 10^{-3}$ & $2.53\times 10^{-3}$ & $2.46\times 10^{-3}$ & $2.88\times 10^{-3}$ & $3.90\times 10^{-3}$ \\
Independent & $1.98\times 10^{-3}$ & $1.97\times 10^{-3}$ & $2.06\times 10^{-3}$ & $2.15\times 10^{-3}$ & $2.36\times 10^{-3}$ \\
\bottomrule
\end{tabular}

\caption{Sharpness for ECE}
\end{subtable}

\begin{subfigure}{0.45\linewidth}
\centering
\centerline{\includegraphics[width=\columnwidth]{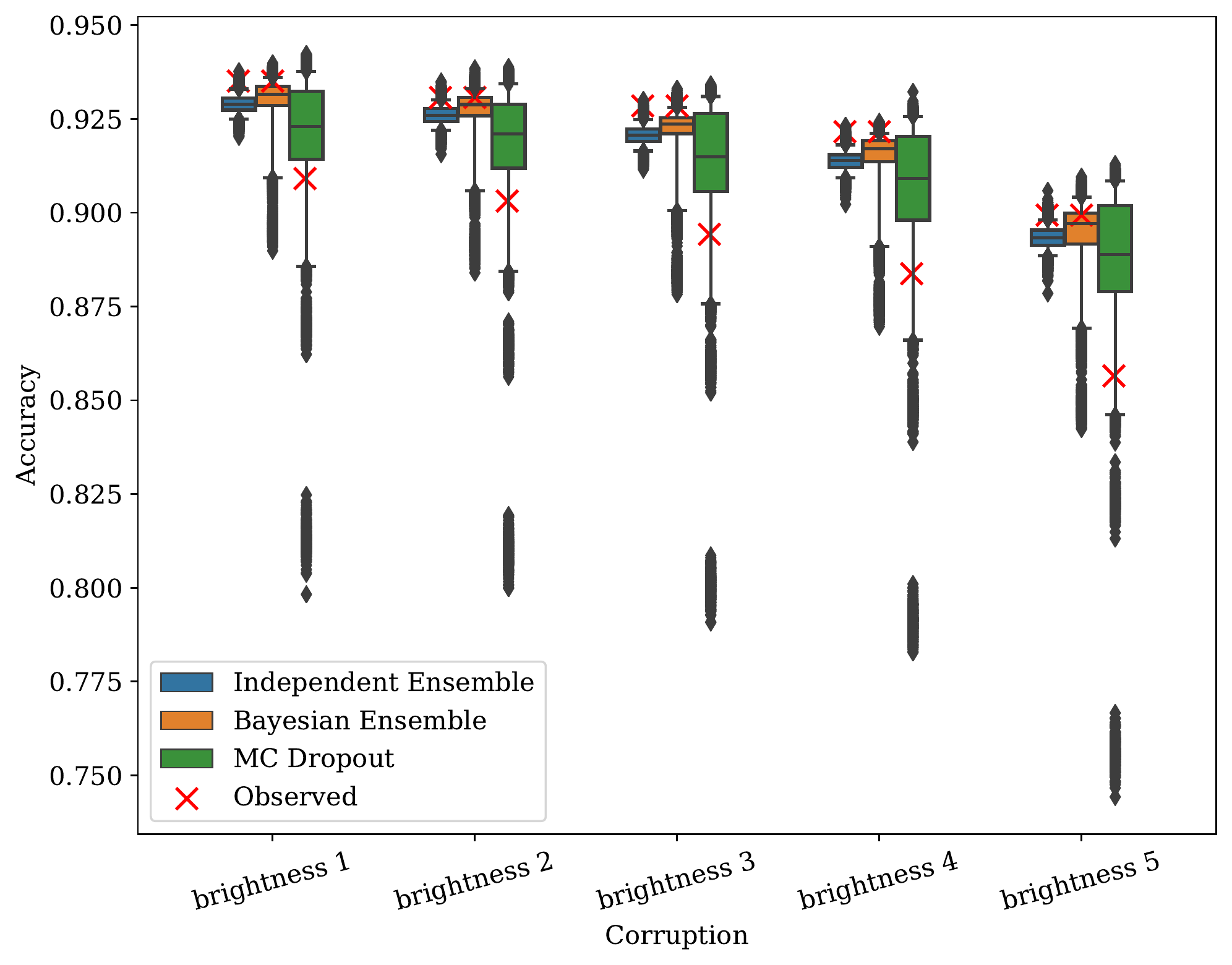}}
\caption{Posterior Check of Accuracy}
\end{subfigure} \hfill
\begin{subfigure}{0.45\linewidth}
\centering
\centerline{\includegraphics[width=\columnwidth]{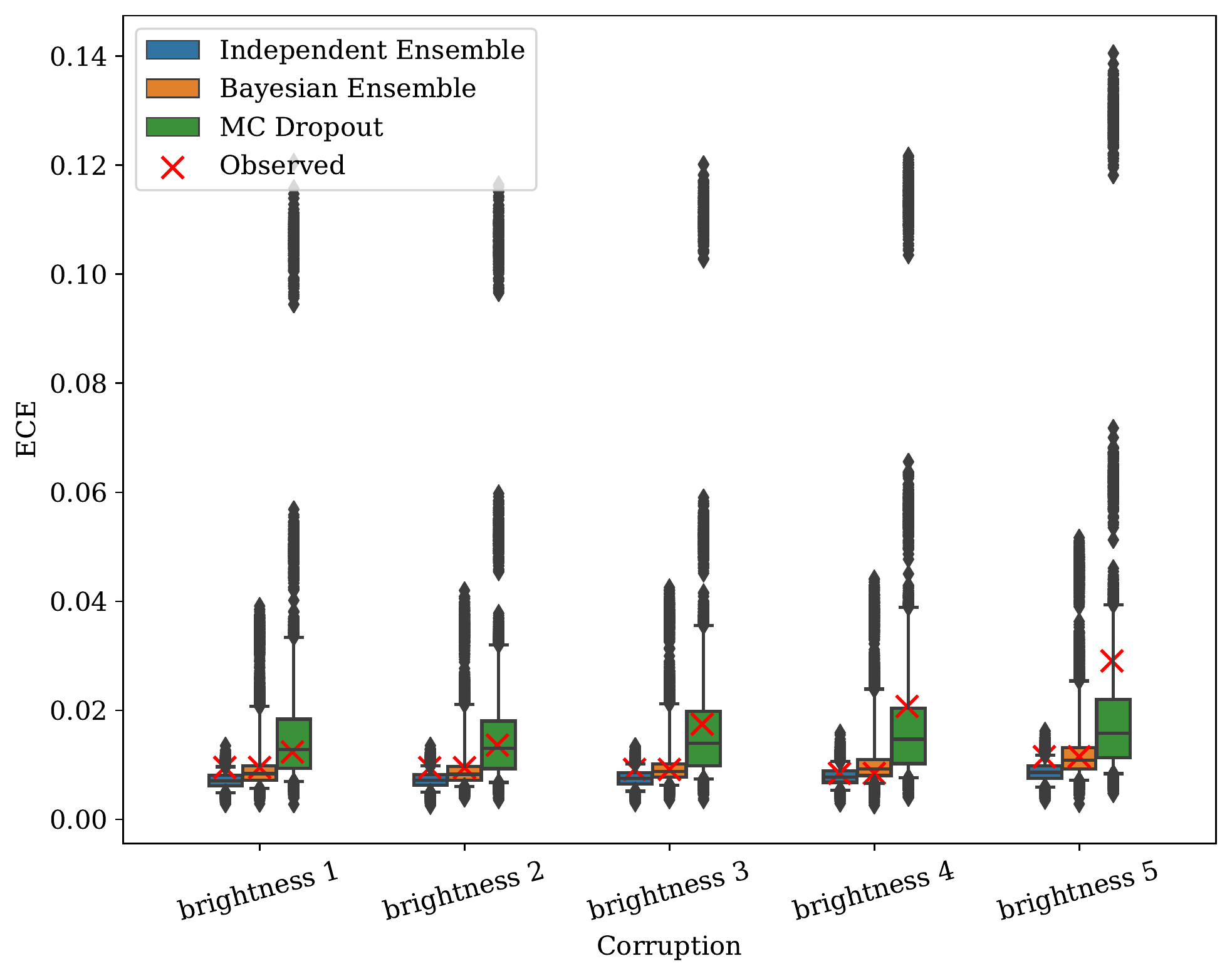}}
\caption{Posterior Check of ECE}
\end{subfigure}
\caption{Posterior predictive checks of recalibrated models trained on CIFAR-10, evaluated on brightness data from CIFAR-10-C.}\label{fig:brightness}
\end{minipage}
\end{figure}
    
\begin{figure}
\begin{minipage}{\textwidth}
\begin{subtable}[h]{0.45\textwidth}
\small
\begin{tabular}{lrrrrr}
\toprule
corruption &    0 &    1 &    2 &    3 &    4 \\
model       &      &      &      &      &      \\
\midrule
Dropout     & 0.75 & 0.78 & 0.44 & 0.19 & 0.11 \\
Ensemble    & 0.91 & 0.89 & 0.31 & 0.20 & 0.13 \\
Independent & 1.00 & 1.00 & 0.43 & 0.01 & 0.00 \\
\bottomrule
\end{tabular}

\caption{p-value for Accuracy}
\end{subtable}\hfill
\begin{subtable}[h]{0.45\textwidth}
\small
\begin{tabular}{lrrrrr}
\toprule
corruption &    0 &    1 &    2 &    3 &    4 \\
model       &      &      &      &      &      \\
\midrule
Dropout     & 0.35 & 0.36 & 0.00 & 0.12 & 0.48 \\
Ensemble    & 0.68 & 0.70 & 0.01 & 0.14 & 0.82 \\
Independent & 0.98 & 0.98 & 0.05 & 0.63 & 1.00 \\
\bottomrule
\end{tabular}

\caption{p-value for ECE}
\end{subtable}
\begin{subtable}[h]{\linewidth}
\small
\centering
\begin{tabular}{lrrrrr}
\toprule
corruption &        0 &        1 &        2 &        3 &        4 \\
model       &          &          &          &          &          \\
\midrule
Dropout     & $2.75\times 10^{-2}$ & $2.69\times 10^{-2}$ & $3.12\times 10^{-2}$ & $3.70\times 10^{-2}$ & $4.25\times 10^{-2}$ \\
Ensemble    & $9.25\times 10^{-3}$ & $1.12\times 10^{-2}$ & $1.44\times 10^{-2}$ & $1.61\times 10^{-2}$ & $1.24\times 10^{-2}$ \\
Independent & $4.16\times 10^{-3}$ & $4.13\times 10^{-3}$ & $4.62\times 10^{-3}$ & $5.25\times 10^{-3}$ & $5.50\times 10^{-3}$ \\
\bottomrule
\end{tabular}

\caption{Sharpness for Accuracy}
\end{subtable}

\begin{subtable}[h]{\linewidth}
\small
\centering
\begin{tabular}{lrrrrr}
\toprule
corruption &        0 &        1 &        2 &        3 &        4 \\
model       &          &          &          &          &          \\
\midrule
Dropout     & $1.44\times 10^{-2}$ & $1.46\times 10^{-2}$ & $1.71\times 10^{-2}$ & $2.27\times 10^{-2}$ & $2.12\times 10^{-2}$ \\
Ensemble    & $4.94\times 10^{-3}$ & $4.31\times 10^{-3}$ & $6.83\times 10^{-3}$ & $7.62\times 10^{-3}$ & $7.31\times 10^{-3}$ \\
Independent & $2.53\times 10^{-3}$ & $2.45\times 10^{-3}$ & $2.75\times 10^{-3}$ & $3.14\times 10^{-3}$ & $3.32\times 10^{-3}$ \\
\bottomrule
\end{tabular}

\caption{Sharpness for ECE}
\end{subtable}

\begin{subfigure}{0.45\linewidth}
\centering
\centerline{\includegraphics[width=\columnwidth]{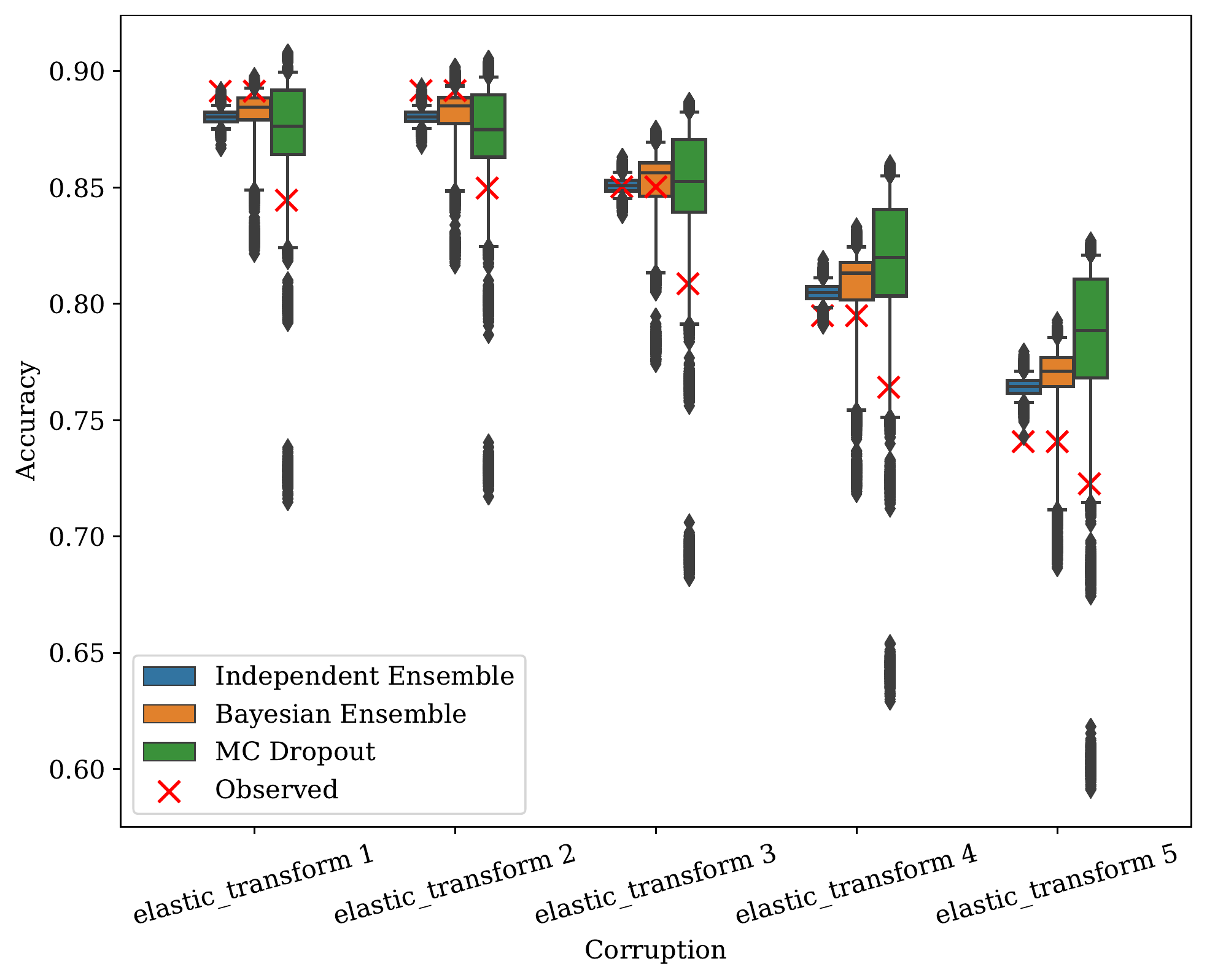}}
\caption{Posterior Check of Accuracy}
\end{subfigure} \hfill
\begin{subfigure}{0.45\linewidth}
\centering
\centerline{\includegraphics[width=\columnwidth]{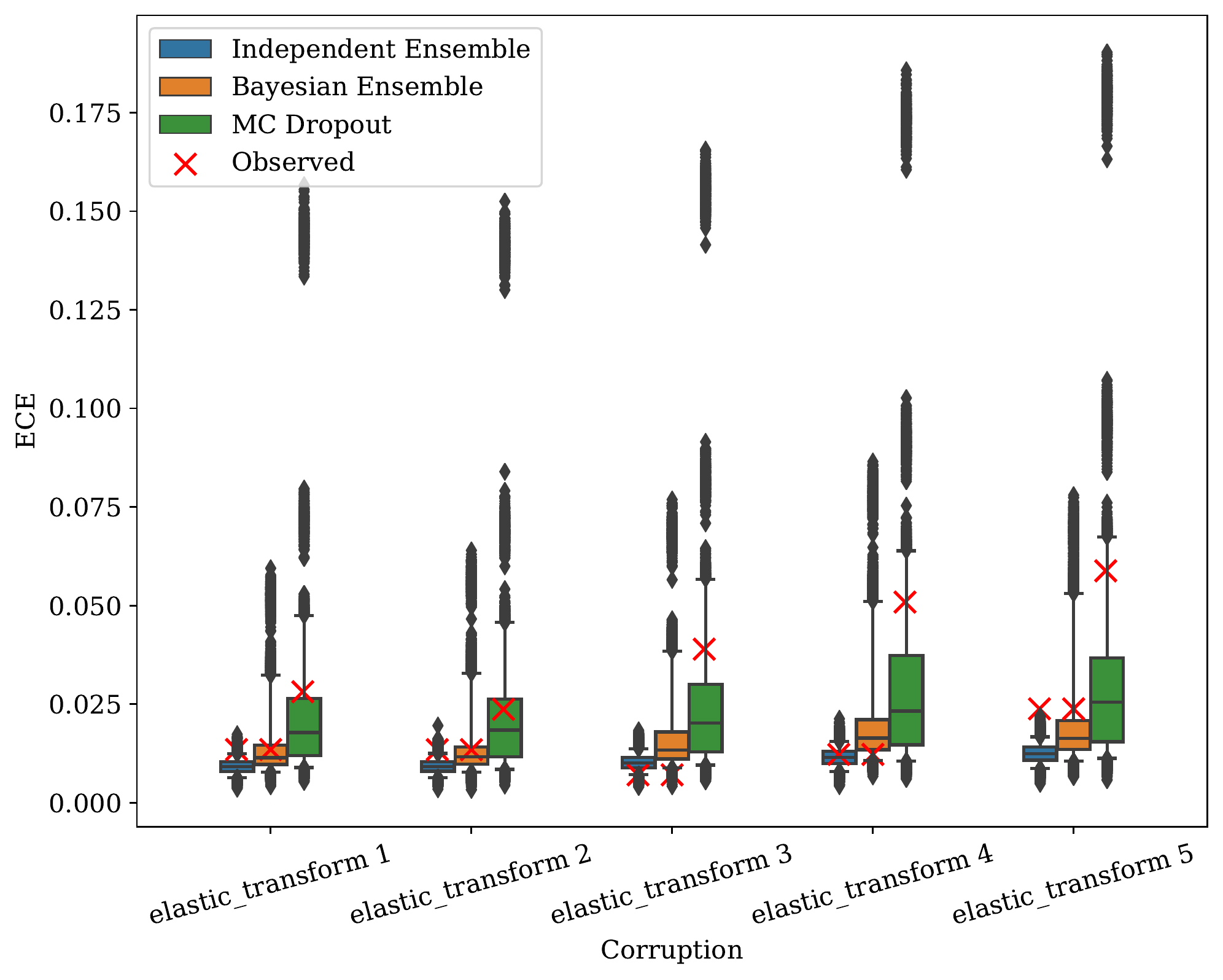}}
\caption{Posterior Check of ECE}
\end{subfigure}
\caption{Posterior predictive checks of recalibrated models trained on CIFAR-10, evaluated on elastic transform data from CIFAR-10-C.}\label{fig:elastic_transform}
\end{minipage}
\end{figure}
    
\begin{figure}
\begin{minipage}{\textwidth}
\begin{subtable}[h]{0.45\textwidth}
\small
\begin{tabular}{lrrrrr}
\toprule
corruption &    0 &    1 &    2 &    3 &    4 \\
model       &      &      &      &      &      \\
\midrule
Dropout     & 0.36 & 0.05 & 0.13 & 0.10 & 0.04 \\
Ensemble    & 0.23 & 0.08 & 0.11 & 0.08 & 0.02 \\
Independent & 0.20 & 0.00 & 0.00 & 0.00 & 0.00 \\
\bottomrule
\end{tabular}

\caption{p-value for Accuracy}
\end{subtable}\hfill
\begin{subtable}[h]{0.45\textwidth}
\small
\begin{tabular}{lrrrrr}
\toprule
corruption &    0 &    1 &    2 &    3 &    4 \\
model       &      &      &      &      &      \\
\midrule
Dropout     & 0.52 & 0.71 & 0.63 & 0.78 & 0.93 \\
Ensemble    & 0.90 & 0.92 & 0.89 & 0.91 & 0.98 \\
Independent & 1.00 & 1.00 & 1.00 & 1.00 & 1.00 \\
\bottomrule
\end{tabular}

\caption{p-value for ECE}
\end{subtable}
\begin{subtable}[h]{\linewidth}
\small
\centering
\begin{tabular}{lrrrrr}
\toprule
corruption &        0 &        1 &        2 &        3 &        4 \\
model       &          &          &          &          &          \\
\midrule
Dropout     & $2.40\times 10^{-2}$ & $3.37\times 10^{-2}$ & $3.07\times 10^{-2}$ & $3.21\times 10^{-2}$ & $3.68\times 10^{-2}$ \\
Ensemble    & $8.13\times 10^{-3}$ & $1.59\times 10^{-2}$ & $1.00\times 10^{-2}$ & $1.11\times 10^{-2}$ & $1.52\times 10^{-2}$ \\
Independent & $4.13\times 10^{-3}$ & $5.13\times 10^{-3}$ & $4.87\times 10^{-3}$ & $5.00\times 10^{-3}$ & $5.37\times 10^{-3}$ \\
\bottomrule
\end{tabular}

\caption{Sharpness for Accuracy}
\end{subtable}

\begin{subtable}[h]{\linewidth}
\small
\centering
\begin{tabular}{lrrrrr}
\toprule
corruption &        0 &        1 &        2 &        3 &        4 \\
model       &          &          &          &          &          \\
\midrule
Dropout     & $1.21\times 10^{-2}$ & $1.68\times 10^{-2}$ & $1.57\times 10^{-2}$ & $1.59\times 10^{-2}$ & $1.84\times 10^{-2}$ \\
Ensemble    & $3.69\times 10^{-3}$ & $6.89\times 10^{-3}$ & $5.28\times 10^{-3}$ & $5.29\times 10^{-3}$ & $6.15\times 10^{-3}$ \\
Independent & $2.50\times 10^{-3}$ & $3.13\times 10^{-3}$ & $2.86\times 10^{-3}$ & $2.94\times 10^{-3}$ & $3.22\times 10^{-3}$ \\
\bottomrule
\end{tabular}

\caption{Sharpness for ECE}
\end{subtable}

\begin{subfigure}{0.45\linewidth}
\centering
\centerline{\includegraphics[width=\columnwidth]{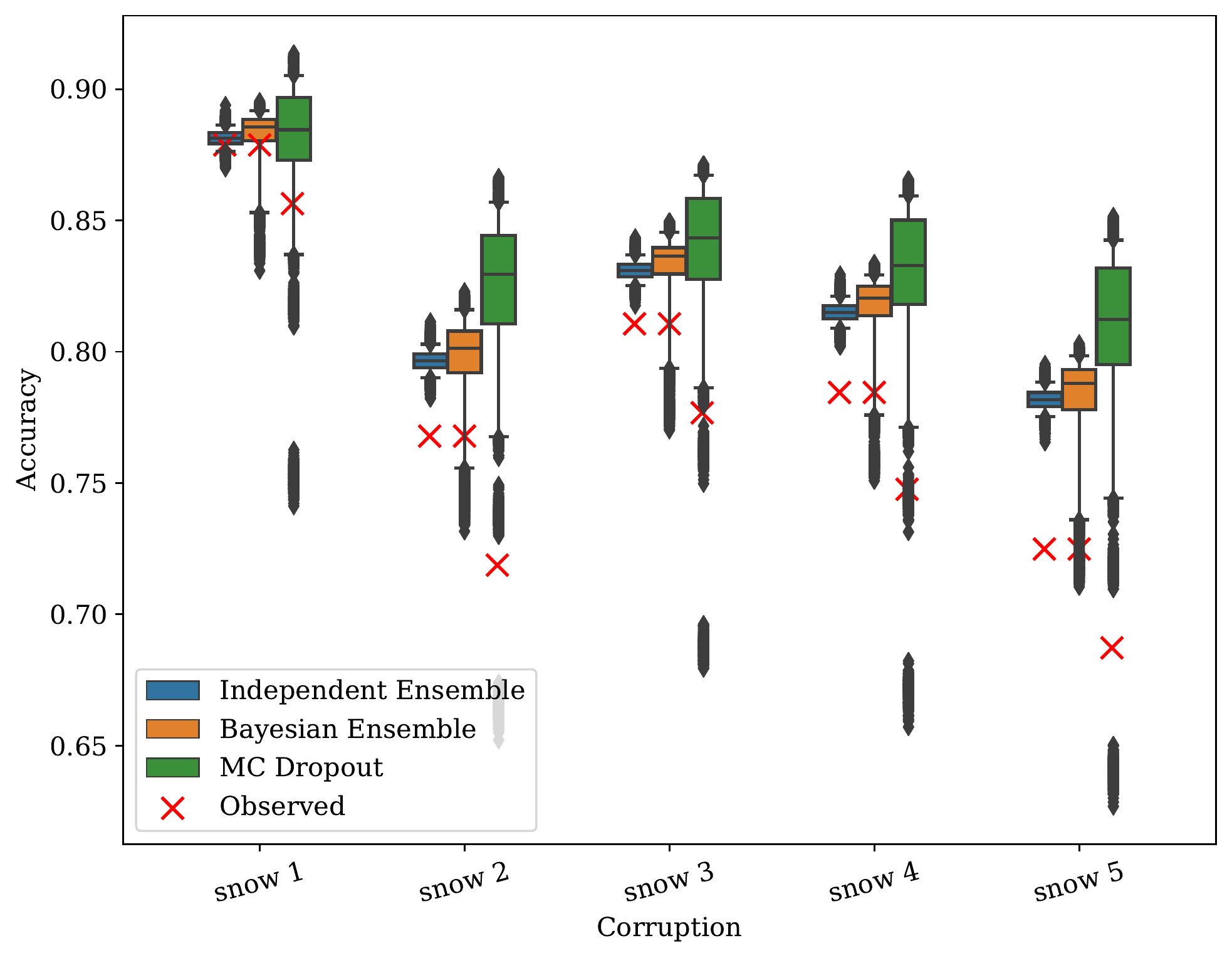}}
\caption{Posterior Check of Accuracy}
\end{subfigure} \hfill
\begin{subfigure}{0.45\linewidth}
\centering
\centerline{\includegraphics[width=\columnwidth]{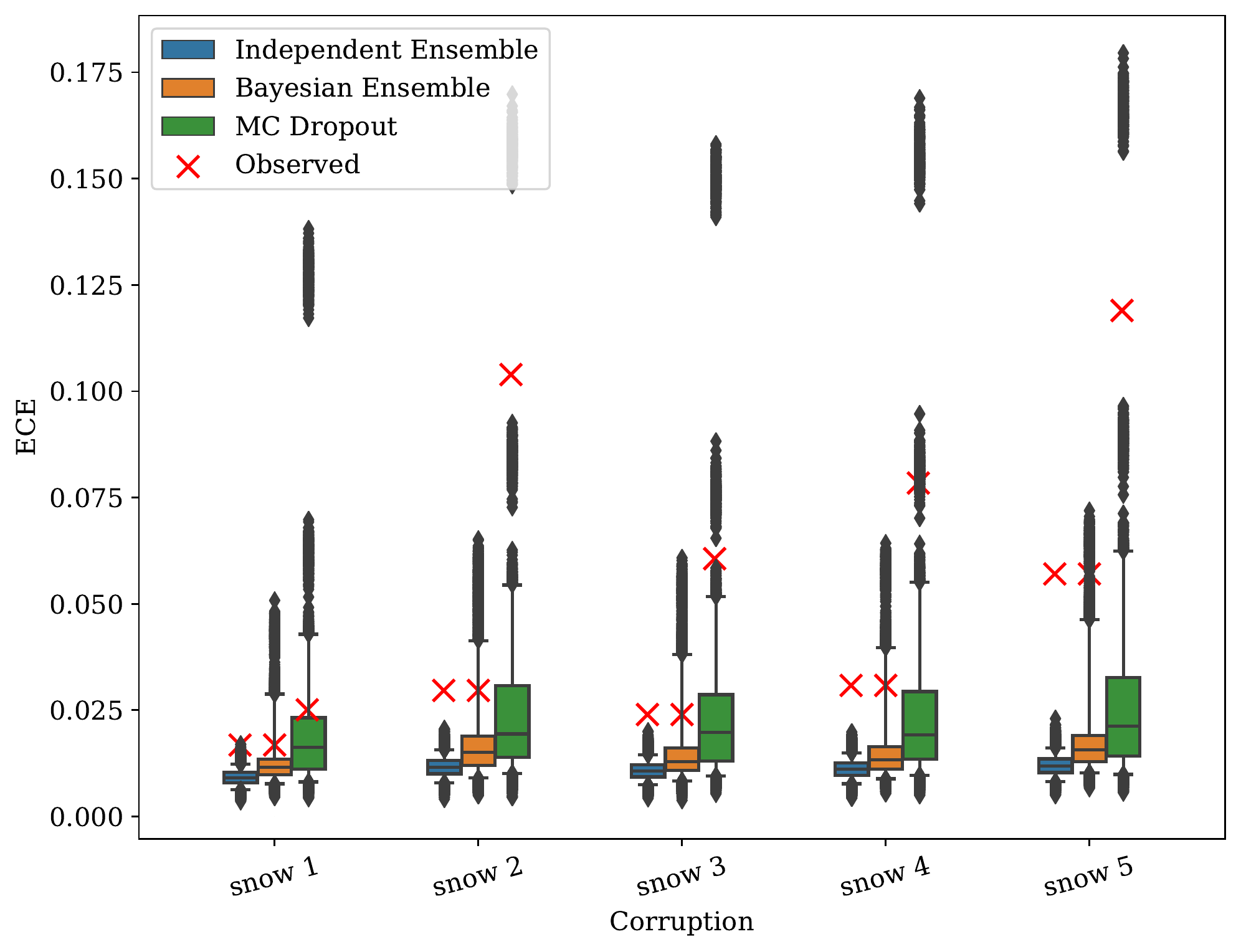}}
\caption{Posterior Check of ECE}
\end{subfigure}
\caption{Posterior predictive checks of recalibrated models trained on CIFAR-10, evaluated on snow data from CIFAR-10-C.}\label{fig:snow}
\end{minipage}
\end{figure}

%% file: main.bbl
\begin{thebibliography}{}

\bibitem[\protect\citeauthoryear{Amini, Schwarting, Soleimany, and Rus}{Amini
  et~al.}{2020}]{amini2020deep}
Amini, A., W.~Schwarting, A.~Soleimany, and D.~Rus (2020).
\newblock Deep evidential uncertainty.

\bibitem[\protect\citeauthoryear{Ashukha, Lyzhov, Molchanov, and
  Vetrov}{Ashukha et~al.}{2020}]{Ashukha2020Pitfalls}
Ashukha, A., A.~Lyzhov, D.~Molchanov, and D.~Vetrov (2020).
\newblock Pitfalls of in-domain uncertainty estimation and ensembling in deep
  learning.
\newblock In {\em International Conference on Learning Representations}.

\bibitem[\protect\citeauthoryear{Badue, Guidolini, Carneiro, Azevedo, Cardoso,
  Forechi, Jesus, Berriel, Paixao, Mutz, et~al.}{Badue
  et~al.}{2021}]{badue2021self}
Badue, C., R.~Guidolini, R.~V. Carneiro, P.~Azevedo, V.~B. Cardoso, A.~Forechi,
  L.~Jesus, R.~Berriel, T.~M. Paixao, F.~Mutz, et~al. (2021).
\newblock Self-driving cars: A survey.
\newblock {\em Expert Systems with Applications\/}~{\em 165}, 113816.

\bibitem[\protect\citeauthoryear{Blundell, Cornebise, Kavukcuoglu, and
  Wierstra}{Blundell et~al.}{2015}]{pmlr-v37-blundell15}
Blundell, C., J.~Cornebise, K.~Kavukcuoglu, and D.~Wierstra (2015, 07--09 Jul).
\newblock Weight uncertainty in neural network.
\newblock In F.~Bach and D.~Blei (Eds.), {\em Proceedings of the 32nd
  International Conference on Machine Learning}, Volume~37 of {\em Proceedings
  of Machine Learning Research}, Lille, France, pp.\  1613--1622. PMLR.

\bibitem[\protect\citeauthoryear{{Bojarski}, {Del Testa}, {Dworakowski},
  {Firner}, {Flepp}, {Goyal}, {Jackel}, {Monfort}, {Muller}, {Zhang}, {Zhang},
  {Zhao}, and {Zieba}}{{Bojarski} et~al.}{2016}]{2016arXiv160407316B}
{Bojarski}, M., D.~{Del Testa}, D.~{Dworakowski}, B.~{Firner}, B.~{Flepp},
  P.~{Goyal}, L.~D. {Jackel}, M.~{Monfort}, U.~{Muller}, J.~{Zhang},
  X.~{Zhang}, J.~{Zhao}, and K.~{Zieba} (2016, April).
\newblock {End to End Learning for Self-Driving Cars}.
\newblock {\em arXiv e-prints\/}, arXiv:1604.07316.

\bibitem[\protect\citeauthoryear{{Brier}}{{Brier}}{1950}]{BrierScore}
{Brier}, G.~W. (1950, January).
\newblock {Verification of Forecasts Expressed in Terms of Probability}.
\newblock {\em Monthly Weather Review\/}~{\em 78\/}(1), 1.

\bibitem[\protect\citeauthoryear{Chen, Dou, Yu, Qin, and Heng}{Chen
  et~al.}{2018}]{chen2018voxresnet}
Chen, H., Q.~Dou, L.~Yu, J.~Qin, and P.-A. Heng (2018).
\newblock Voxresnet: Deep voxelwise residual networks for brain segmentation
  from 3d mr images.
\newblock {\em NeuroImage\/}~{\em 170}, 446--455.

\bibitem[\protect\citeauthoryear{Chen, Behrmann, Duvenaud, and Jacobsen}{Chen
  et~al.}{2019}]{Chen2019ResidualFlows}
Chen, R. T.~Q., J.~Behrmann, D.~K. Duvenaud, and J.-H. Jacobsen (2019).
\newblock Residual flows for invertible generative modeling.
\newblock In H.~Wallach, H.~Larochelle, A.~Beygelzimer, F.~d\' Alch\'{e}-Buc,
  E.~Fox, and R.~Garnett (Eds.), {\em Advances in Neural Information Processing
  Systems 32}, pp.\  9916--9926. Curran Associates, Inc.

\bibitem[\protect\citeauthoryear{{Clevert}, {Unterthiner}, and
  {Hochreiter}}{{Clevert} et~al.}{2015}]{ELU2015}
{Clevert}, D.-A., T.~{Unterthiner}, and S.~{Hochreiter} (2015).
\newblock {Fast and Accurate Deep Network Learning by Exponential Linear Units
  (ELUs)}.
\newblock In {\em International Conference on Learning Representations}.

\bibitem[\protect\citeauthoryear{Der~Kiureghian and Ditlevsen}{Der~Kiureghian
  and Ditlevsen}{2009}]{UncertaintyTaxonomy}
Der~Kiureghian, A. and O.~Ditlevsen (2009).
\newblock Aleatory or epistemic? does it matter?
\newblock {\em Structural safety\/}~{\em 31\/}(2), 105--112.

\bibitem[\protect\citeauthoryear{Ding, Liu, Xiong, and Shi}{Ding
  et~al.}{2020}]{ding2020revisiting}
Ding, Y., J.~Liu, J.~Xiong, and Y.~Shi (2020).
\newblock Revisiting the evaluation of uncertainty estimation and its
  application to explore model complexity-uncertainty trade-off.
\newblock In {\em Proceedings of the IEEE/CVF Conference on Computer Vision and
  Pattern Recognition Workshops}, pp.\  4--5.

\bibitem[\protect\citeauthoryear{Dua and Graff}{Dua and
  Graff}{2017}]{UCIDatasets}
Dua, D. and C.~Graff (2017).
\newblock {UCI} machine learning repository.

\bibitem[\protect\citeauthoryear{Gal and Ghahramani}{Gal and
  Ghahramani}{2016}]{MCDropout}
Gal, Y. and Z.~Ghahramani (2016, 20--22 Jun).
\newblock Dropout as a bayesian approximation: Representing model uncertainty
  in deep learning.
\newblock In M.~F. Balcan and K.~Q. Weinberger (Eds.), {\em Proceedings of The
  33rd International Conference on Machine Learning}, Volume~48 of {\em
  Proceedings of Machine Learning Research}, New York, New York, USA, pp.\
  1050--1059. PMLR.

\bibitem[\protect\citeauthoryear{Gelman, Carlin, Stern, Dunson, Vehtari, and
  Rubin}{Gelman et~al.}{2013}]{gelman2013bayesian}
Gelman, A., J.~B. Carlin, H.~S. Stern, D.~B. Dunson, A.~Vehtari, and D.~B.
  Rubin (2013).
\newblock {\em Bayesian data analysis}.
\newblock CRC press.

\bibitem[\protect\citeauthoryear{Graves}{Graves}{2011}]{PracticalVI2011}
Graves, A. (2011).
\newblock Practical variational inference for neural networks.
\newblock In J.~Shawe-Taylor, R.~Zemel, P.~Bartlett, F.~Pereira, and K.~Q.
  Weinberger (Eds.), {\em Advances in Neural Information Processing Systems},
  Volume~24. Curran Associates, Inc.

\bibitem[\protect\citeauthoryear{Guo, Pleiss, Sun, and Weinberger}{Guo
  et~al.}{2017}]{GuoCalibration}
Guo, C., G.~Pleiss, Y.~Sun, and K.~Q. Weinberger (2017, 06--11 Aug).
\newblock On calibration of modern neural networks.
\newblock In D.~Precup and Y.~W. Teh (Eds.), {\em Proceedings of the 34th
  International Conference on Machine Learning}, Volume~70 of {\em Proceedings
  of Machine Learning Research}, International Convention Centre, Sydney,
  Australia, pp.\  1321--1330. PMLR.

\bibitem[\protect\citeauthoryear{Gupta, Rahimi, Ajanthan, Mensink,
  Sminchisescu, and Hartley}{Gupta et~al.}{2021}]{gupta2021calibration}
Gupta, K., A.~Rahimi, T.~Ajanthan, T.~Mensink, C.~Sminchisescu, and R.~Hartley
  (2021).
\newblock Calibration of neural networks using splines.
\newblock In {\em International Conference on Learning Representations}.

\bibitem[\protect\citeauthoryear{Havaei, Davy, Warde-Farley, Biard, Courville,
  Bengio, Pal, Jodoin, and Larochelle}{Havaei et~al.}{2017}]{havaei2017brain}
Havaei, M., A.~Davy, D.~Warde-Farley, A.~Biard, A.~Courville, Y.~Bengio,
  C.~Pal, P.-M. Jodoin, and H.~Larochelle (2017).
\newblock Brain tumor segmentation with deep neural networks.
\newblock {\em Medical image analysis\/}~{\em 35}, 18--31.

\bibitem[\protect\citeauthoryear{Hendrycks and Dietterich}{Hendrycks and
  Dietterich}{2019}]{hendrycks2019robustness}
Hendrycks, D. and T.~Dietterich (2019).
\newblock Benchmarking neural network robustness to common corruptions and
  perturbations.
\newblock In {\em International Conference on Learning Representations}.

\bibitem[\protect\citeauthoryear{Hernandez-Lobato and Adams}{Hernandez-Lobato
  and Adams}{2015}]{ProbBackProp}
Hernandez-Lobato, J.~M. and R.~Adams (2015, 07--09 Jul).
\newblock Probabilistic backpropagation for scalable learning of bayesian
  neural networks.
\newblock In F.~Bach and D.~Blei (Eds.), {\em Proceedings of the 32nd
  International Conference on Machine Learning}, Volume~37 of {\em Proceedings
  of Machine Learning Research}, Lille, France, pp.\  1861--1869. PMLR.

\bibitem[\protect\citeauthoryear{Huang and Chen}{Huang and
  Chen}{2020}]{huang2020autonomous}
Huang, Y. and Y.~Chen (2020).
\newblock Autonomous driving with deep learning: A survey of state-of-art
  technologies.
\newblock {\em arXiv preprint arXiv:2006.06091\/}.

\bibitem[\protect\citeauthoryear{{Kingma} and {Ba}}{{Kingma} and
  {Ba}}{2015}]{Adam2015}
{Kingma}, D.~P. and J.~{Ba} (2015).
\newblock Adam: A method for stochastic optimization.
\newblock In {\em International Conference on Machine Learning}.

\bibitem[\protect\citeauthoryear{Kingma, Salimans, and Welling}{Kingma
  et~al.}{2015}]{NIPS2015_bc731692}
Kingma, D.~P., T.~Salimans, and M.~Welling (2015).
\newblock Variational dropout and the local reparameterization trick.
\newblock In C.~Cortes, N.~Lawrence, D.~Lee, M.~Sugiyama, and R.~Garnett
  (Eds.), {\em Advances in Neural Information Processing Systems}, Volume~28.
  Curran Associates, Inc.

\bibitem[\protect\citeauthoryear{Krizhevsky and Hinton}{Krizhevsky and
  Hinton}{2009}]{Krizhevsky09CIFAR10}
Krizhevsky, A. and G.~Hinton (2009).
\newblock Learning multiple layers of features from tiny images.
\newblock {\em Master's thesis, Department of Computer Science, University of
  Toronto\/}.

\bibitem[\protect\citeauthoryear{Kuleshov, Fenner, and Ermon}{Kuleshov
  et~al.}{2018}]{kuleshov2018calibration}
Kuleshov, V., N.~Fenner, and S.~Ermon (2018, 10--15 Jul).
\newblock Accurate uncertainties for deep learning using calibrated regression.
\newblock In J.~Dy and A.~Krause (Eds.), {\em Proceedings of the 35th
  International Conference on Machine Learning}, Volume~80 of {\em Proceedings
  of Machine Learning Research}, Stockholmsmässan, Stockholm Sweden, pp.\
  2796--2804. PMLR.

\bibitem[\protect\citeauthoryear{Lakshminarayanan, Pritzel, and
  Blundell}{Lakshminarayanan et~al.}{2017}]{DeepEnsembles}
Lakshminarayanan, B., A.~Pritzel, and C.~Blundell (2017).
\newblock Simple and scalable predictive uncertainty estimation using deep
  ensembles.
\newblock In I.~Guyon, U.~V. Luxburg, S.~Bengio, H.~Wallach, R.~Fergus,
  S.~Vishwanathan, and R.~Garnett (Eds.), {\em Advances in Neural Information
  Processing Systems 30}, pp.\  6402--6413. Curran Associates, Inc.

\bibitem[\protect\citeauthoryear{Nixon, Dusenberry, Zhang, Jerfel, and
  Tran}{Nixon et~al.}{2019}]{nixon2019measuring}
Nixon, J., M.~W. Dusenberry, L.~Zhang, G.~Jerfel, and D.~Tran (2019).
\newblock Measuring calibration in deep learning.
\newblock In {\em CVPR Workshops}, Volume~2.

\bibitem[\protect\citeauthoryear{Ovadia, Fertig, Ren, Nado, Sculley, Nowozin,
  Dillon, Lakshminarayanan, and Snoek}{Ovadia et~al.}{2019}]{Ovadia2019}
Ovadia, Y., E.~Fertig, J.~Ren, Z.~Nado, D.~Sculley, S.~Nowozin, J.~Dillon,
  B.~Lakshminarayanan, and J.~Snoek (2019).
\newblock Can you trust your model\'s uncertainty? evaluating predictive
  uncertainty under dataset shift.
\newblock In H.~Wallach, H.~Larochelle, A.~Beygelzimer, F.~d\'Alch\'{e} Buc,
  E.~Fox, and R.~Garnett (Eds.), {\em Advances in Neural Information Processing
  Systems}, Volume~32. Curran Associates, Inc.

\bibitem[\protect\citeauthoryear{Papadopoulos, Edwards, and
  Murray}{Papadopoulos et~al.}{2000}]{papadopoulos2000}
Papadopoulos, G., P.~Edwards, and A.~Murray (2000, 01).
\newblock Confidence estimation methods for neural networks: a practical
  comparison.
\newblock pp.\  75--80.

\bibitem[\protect\citeauthoryear{Papamakarios, Nalisnick, Rezende, Mohamed, and
  Lakshminarayanan}{Papamakarios et~al.}{2021}]{papamakarios2021normalizing}
Papamakarios, G., E.~Nalisnick, D.~J. Rezende, S.~Mohamed, and
  B.~Lakshminarayanan (2021).
\newblock Normalizing flows for probabilistic modeling and inference.

\bibitem[\protect\citeauthoryear{Pearce, Brintrup, Zaki, and Neely}{Pearce
  et~al.}{2018}]{QualityDrivenLoss2018}
Pearce, T., A.~Brintrup, M.~Zaki, and A.~Neely (2018, 10--15 Jul).
\newblock High-quality prediction intervals for deep learning: A
  distribution-free, ensembled approach.
\newblock Volume~80 of {\em Proceedings of Machine Learning Research},
  Stockholmsmässan, Stockholm Sweden, pp.\  4075--4084. PMLR.

\bibitem[\protect\citeauthoryear{Rezende, Mohamed, and Wierstra}{Rezende
  et~al.}{2014}]{rezende2014stochastic}
Rezende, D.~J., S.~Mohamed, and D.~Wierstra (2014).
\newblock Stochastic backpropagation and approximate inference in deep
  generative models.
\newblock In {\em International Conference on Machine Learning}, pp.\
  1278--1286.

\bibitem[\protect\citeauthoryear{Tagasovska and Lopez-Paz}{Tagasovska and
  Lopez-Paz}{2019}]{SQR}
Tagasovska, N. and D.~Lopez-Paz (2019).
\newblock Single-model uncertainties for deep learning.
\newblock In {\em Advances in Neural Information Processing Systems 32}, pp.\
  6417--6428. Curran Associates, Inc.

\bibitem[\protect\citeauthoryear{Tipping}{Tipping}{2001}]{BayesianRidgeRegression}
Tipping, M.~E. (2001).
\newblock Sparse bayesian learning and the relevance vector machine.
\newblock {\em Journal of machine learning research\/}~{\em 1\/}(Jun),
  211--244.

\bibitem[\protect\citeauthoryear{Wachinger, Reuter, and Klein}{Wachinger
  et~al.}{2018}]{wachinger2018deepnat}
Wachinger, C., M.~Reuter, and T.~Klein (2018).
\newblock Deepnat: Deep convolutional neural network for segmenting
  neuroanatomy.
\newblock {\em NeuroImage\/}~{\em 170}, 434--445.

\bibitem[\protect\citeauthoryear{Welling and Teh}{Welling and
  Teh}{2011}]{Welling2011}
Welling, M. and Y.~W. Teh (2011).
\newblock Bayesian learning via stochastic gradient langevin dynamics.
\newblock In {\em Proceedings of the 28th International Conference on
  International Conference on Machine Learning}, ICML'11, Madison, WI, USA,
  pp.\  681–688. Omnipress.

\bibitem[\protect\citeauthoryear{Zhang, Li, Zhang, Chen, and Wilson}{Zhang
  et~al.}{2020}]{zhang2020csgmcmc}
Zhang, R., C.~Li, J.~Zhang, C.~Chen, and A.~G. Wilson (2020).
\newblock Cyclical stochastic gradient mcmc for bayesian deep learning.
\newblock {\em International Conference on Learning Representations\/}.

\bibitem[\protect\citeauthoryear{Zhou, Greenspan, Davatzikos, Duncan,
  Van~Ginneken, Madabhushi, Prince, Rueckert, and Summers}{Zhou
  et~al.}{2021}]{zhou20021medimaging}
Zhou, S.~K., H.~Greenspan, C.~Davatzikos, J.~S. Duncan, B.~Van~Ginneken,
  A.~Madabhushi, J.~L. Prince, D.~Rueckert, and R.~M. Summers (2021).
\newblock A review of deep learning in medical imaging: Imaging traits,
  technology trends, case studies with progress highlights, and future
  promises.
\newblock {\em Proceedings of the IEEE\/}~{\em 109\/}(5), 820--838.

\end{thebibliography}
